\documentclass{article}
\usepackage[utf8]{inputenc} 
\usepackage[T1]{fontenc}    
\usepackage{hyperref}       
\usepackage{url}            
\usepackage{booktabs}       
\usepackage{amsfonts}       
\usepackage{nicefrac}       
\usepackage{microtype}      
\usepackage{xcolor}         
\usepackage{amsmath}
\usepackage{amsthm}
\usepackage{subcaption}
\usepackage{enumitem}
\usepackage{bbm}
\usepackage{amssymb}
\usepackage[capitalise]{cleveref}
\usepackage{appendix}
\usepackage{comment}

\usepackage{amsmath}
\usepackage{amssymb}
\usepackage{mathtools}
\usepackage{amsthm}

\usepackage{tikz}
\tikzset{
	circ/.style = {circle, draw, inner sep=0.04cm,node contents={}}
}
\usetikzlibrary{shapes,decorations,arrows,calc,arrows.meta,fit,positioning}
\tikzset{
    -Latex,auto,node distance = 0.7 cm and 0.7 cm, semithick,
    state/.style ={ellipse, draw, minimum width = 0.2 cm},
    point/.style = {circle, draw, inner sep=0.04cm,fill,node contents={}},
    bidirected/.style={Latex-Latex,dashed},
    el/.style = {inner sep=2pt, align=left, sloped}
}
\usetikzlibrary{positioning}
\usepackage{graphicx}

\theoremstyle{plain}
\newtheorem{theorem}{Theorem}[section]
\newtheorem{proposition}[theorem]{Proposition}
\newtheorem{lemma}[theorem]{Lemma}

\theoremstyle{definition}
\newtheorem{definition}[theorem]{Definition}

\theoremstyle{remark}

\newcommand{\indep }{\perp\!\!\!\perp}
\newcommand{\dep }{\not\!\perp\!\!\!\perp }

\newcommand{\mX}{\mathbf{X}}
\newcommand{\mx}{\mathbf{x}}
\newcommand{\mY}{\mathbf{Y}}
\newcommand{\my}{\mathbf{y}}
\newcommand{\mZ}{\mathbf{Z}}
\newcommand{\mz}{\mathbf{z}}
\newcommand{\mI}{\mathbf{I}}
\newcommand{\mJ}{\mathbf{J}}
\newcommand{\mK}{\mathbf{K}}
\newcommand{\mk}{\mathbf{k}}
\newcommand{\mW}{\mathbf{W}}
\newcommand{\mw}{\mathbf{w}}
\newcommand{\mV}{\mathbf{V}}
\newcommand{\mv}{\mathbf{v}}
\newcommand{\mE}{\mathbf{E}}
\newcommand{\mL}{\mathbf{L}}
\newcommand{\mR}{\mathbf{R}}
\newcommand{\mS}{\mathbf{S}}
\newcommand{\ms}{\mathbf{s}}

\newcommand{\mcC}{\mathcal{C}}
\newcommand{\mcD}{\mathcal{D}}
\newcommand{\mcF}{\mathcal{F}}
\newcommand{\mcG}{\mathcal{G}}
\newcommand{\mcI}{\mathcal{I}}
\newcommand{\mcL}{\mathcal{L}}
\newcommand{\mcP}{\mathcal{P}}
\newcommand{\mcE}{\mathcal{E}}
\newcommand{\mcM}{\mathcal{M}}

\newcommand{\Aum}{\mathrm{Aug}}
\newcommand{\Aug}{\mathrm{Aug}}
\newcommand{\Twin}{\mathrm{Twin}}
\newcommand{\MAG}{\mathrm{MAG}}
\newcommand{\crightarrow}{o\!\!\rightarrow}

\newcommand{\cleftrightarrow}{o$\mbox{---}$o }

\newcommand{\srightarrow}{*\!\!\rightarrow}
\newcommand{\sleftarrow}{\leftarrow\!\!\!\!*}
\newcommand{\starstar}{*\!\!\mbox{---}\!\!*}

\newcommand{\circlecircle}{o\mbox{---}o}
\newcommand{\circlestar}{o\mbox{---}\!*}

\usepackage[margin=1in]{geometry}
\usepackage{authblk}

\usepackage[utf8]{inputenc} 
\usepackage[T1]{fontenc}    
\usepackage{hyperref}       
\usepackage{url}            
\usepackage{booktabs}       
\usepackage{amsfonts}       
\usepackage{nicefrac}       
\usepackage{microtype}      

\usepackage[noend]{algorithmic}
\RequirePackage{algorithm}

\usepackage{xargs}                      
\usepackage{xcolor}

\usepackage{tikz}
\usetikzlibrary{positioning, calc, shapes.geometric, shapes, shapes.multipart, arrows.meta, arrows, decorations.markings, external, trees}

\usepackage{scalefnt}
\usetikzlibrary{shapes.misc}
\usetikzlibrary{positioning, calc, shapes.geometric, shapes, shapes.multipart, arrows.meta, arrows, decorations.markings, external, trees, fit}
\tikzset{
	-Latex,auto,node distance =1 cm and 1 cm,semithick,
	state/.style ={ellipse, draw, minimum width = 0.7 cm},
	point/.style = {circle, draw, inner sep=0.04cm,fill,node contents={}},
	nnv/.style={
		rectangle, draw,thick,minimum width=0.7cm,minimum height=1.5cm
	},
        nnh/.style={
            rectangle, draw,thick,minimum width=1.5cm,minimum height=1.0cm
          },
	outer/.style={draw=gray,dashed,thick, inner sep=3pt
	},
	XOR/.style={draw,circle,append after command={
			[shorten >=\pgflinewidth, shorten <=\pgflinewidth,]
			(\tikzlastnode.north) edge (\tikzlastnode.south)
			(\tikzlastnode.east) edge (\tikzlastnode.west)
		}
	},
	bidirected/.style={Latex-Latex,dashed},
	el/.style = {inner sep=2pt, align=left, sloped},
	cross/.style={cross out, draw=black, minimum size=2*(#1-\pgflinewidth), inner sep=0pt, outer sep=0pt},
	cross/.default={1pt}
}

\usepackage{natbib}
\usepackage{microtype}

\title{Characterization and Learning of Causal Graphs from \\Hard Interventions}

\author{%
  Zihan Zhou$^{*1}$, Muhammad Qasim Elahi$^{*2}$, Murat Kocaoglu$^1$\\
    Department of Computer Science, Johns Hopkins University$^1$\\
    School of Electrical and Computer Engineering, Purdue University$^2$\\

  \texttt{zzhou150@jhu.edu, elahi0@purdue.edu, 
  mkocaoglu@jhu.edu} \\
}

\usepackage{fancyhdr}
\begin{document}
\flushbottom
\addtocontents{toc}{\protect\setcounter{tocdepth}{-1}}
\def\thefootnote{*}\footnotetext{Equal Contribution}\def\thefootnote{\arabic{footnote}}
\maketitle
\begin{abstract}
    \label{sec:abstract}
 A fundamental challenge in the empirical sciences involves uncovering causal structure through observation and experimentation. Causal discovery entails linking the conditional independence (CI) invariances in observational data to their corresponding graphical constraints via d-separation. In this paper, we consider a general setting where we have access to data from multiple experimental distributions resulting from hard interventions, as well as potentially from an observational distribution. By comparing different interventional distributions, we propose a set of graphical constraints that are fundamentally linked to Pearl's do-calculus within the framework of hard interventions. These graphical constraints associate each graphical structure with a set of interventional distributions that are consistent with the rules of do-calculus. We characterize the interventional equivalence class of causal graphs with latent variables and introduce a graphical representation that can be used to determine whether two causal graphs are interventionally equivalent, i.e., whether they are associated with the same family of hard interventional distributions, where the elements of the family are indistinguishable using the invariances from do-calculus. We also propose a learning algorithm to integrate multiple datasets from hard interventions, introducing new orientation rules. The learning objective is a tuple of augmented graphs which entails a set of causal graphs. We also prove the soundness of the proposed algorithm.

\end{abstract}

\section{Introduction}
Understanding the behavior of complex systems through their causal relationships is a fundamental problem in science. Researchers collect data and perform experiments to analyze how specific phenomena arise or to investigate the structure and function of underlying systems, whether social, biological, or economic \cite{hunermund2023causal,petersen2024causal,sanchez2022causal}. Causal discovery focuses on identifying causal relationships from both observational and interventional data \cite{pearl1995causal,spirtes2001causation,peters2017elements}. A widely used approach for causal discovery models the underlying system as a causal graph, represented by a directed acyclic graph (DAG), where nodes denote random variables and directed edges between nodes $(A \rightarrow B)$ signify causal relations \cite{pearl2009causality,spirtes2001causation}. 

Causal discovery involves deriving constraints from data to infer the underlying causal graph. However, in practice, these constraints are rarely enough to identify the exact causal graph. Instead, they typically define a set of graphs consistent with the data, collectively referred to as an equivalence class (EC) \cite{ali2012towards,meek2013causal}. 
Conditional independence (CI) relations are the primary markers of the underlying causal structure and are used to define an equivalence class. These fundamental probabilistic invariances have been widely explored within the framework of graphical models \cite{pearl1995causal,peters2017elements}. Conditional independencies (CIs) are a powerful tool and serve as the foundation for many causal discovery algorithms, e.g., PC and FCI~\cite{spirtes2001causation}. 

When only observational data is available, the Markov equivalence class (MEC) comprises all causal graphs that exhibit the same set of conditional independences (CIs) among the measured variables, as defined by the d-separation criterion \cite{verma1992algorithm}. The availability of interventional (i.e., experimental) data allows us to reduce the size of the equivalence class, potentially facilitating the recovery of the true causal graph \cite{hauser2012characterization,kocaoglu2017experimental}. 
Hard and soft interventions are two methods for manipulating variables in causal systems. A hard intervention directly sets a variable to a fixed value, removing natural dependencies, while a soft intervention alters the mechanism through which the parents of a variable influence the target variable. While soft interventions are more common in biology as, e.g., in gene knockout experiments~\cite{meinshausen2016methods} since we do not have precise control of mechanisms, in computer systems hard interventions are feasible and have been used for learning causal relations, e.g., in microservice architectures~\cite{wang2023fault}. 

Although a hard intervention can be seen as a special case of a soft one, it can be more informative in the presence of latent variables in many cases. For instance, consider the causal graph \( \mathcal{D}_1 = \{X \rightarrow Z \rightarrow Y, Z \leftrightarrow Y\} \). A hard intervention on \( Z \) breaks the inducing path\footnote{Inducing paths are paths between non-adjacent variables that cannot be blocked by conditioning on any subset of observed variables. They only exist in the presence of unobserved confounders. } $\langle X,Z,Y \rangle$, which implies that after a hard intervention on \( Z \), the variables \( X \) and \( Y \) are no longer dependent. In contrast, in the case of a soft intervention, the incoming edges to \( Z \) are not removed, so a soft intervention will not break the inducing path. It is also worth noting that hard interventions may change d-separation statements non-locally, as seen here between $X,Y$ after $do(z)$, which the existing representations of interventional Markov equivalence classes cannot encode. Now consider another graph \( \mathcal{D}_2 = \{X \rightarrow Z \rightarrow Y, Z \leftrightarrow Y, X \rightarrow Y\} \), which is the same as \( \mathcal{D}_1 \) with the additional edge \( X \rightarrow Y \). With access to a hard intervention on \( Z \), we can distinguish these graphs; however, with a soft intervention, one cannot differentiate between them. This example demonstrates that hard interventions can be more informative in the presence of latent variables and narrow down the search space more effectively compared to soft interventions. A fundamental question is how can we extract as much causal knowledge as possible from a collection of hard interventional datasets. To the best of our knowledge, this problem has been open before this work.  

Motivated by this, our paper considers a general setting where multiple experimental distributions resulting from hard interventions are available alongside (optionally) observational distributions. 
Prior work has focused on characterizing the $\mathcal{I}$-Markov equivalence class through distributional invariances both within and across a set of observational and interventional distributions \cite{hauser2012characterization,yang2018characterizing}. The closest work to ours is \cite{kocaoglu2019characterization}. However, they deal with soft interventions, whereas we consider a setting where experimental data comes from hard interventions. 
This can lead to more invariances that can be inferred from the experimental data compared to the soft intervention case, which can potentially further reduce the size of  $\mathcal{I}$-Markov equivalence class. However, the existing work cannot utilize these additional invariances.

We propose using \textit{do-constraints} with hard interventions, a concept that emerges from comparing observational and experimental distributions, extending Pearl's do-calculus to uncover new structural insights in causal graphs~\cite{kocaoglu2019characterization,jaber2020causal}.
They emerge as the converse of the causal calculus developed by  \cite{pearl1995causal}. These constraints, distinct from traditional conditional independence (CI) relations, are derived by contrasting distributions such as $P(y|x)$ and $P(y|\mathrm{do}(x))$ through a \textit{do-see test}. When differences are detected, they reveal open backdoor paths in the graph, aiding structure learning. We leverage 
\textit{F-nodes}, introduced by Pearl, to explicitly encode intervention effects in augmented graphs \cite{pearl1993aspects}. These nodes make the effects of interventions visible within the graph, enabling the application of do-calculus tests and capturing key structural knowledge such as the existence of a backdoor path from $X$ to $Y$ when $F_X$ is not d-separated from $Y$ given $X$. Such augmented representations, widely used in inference and identification, highlight the utility of do-constraints alongside CI relations for learning causal structures \cite{yang2018characterizing,kocaoglu2019characterization,mooij2020joint}.

We say that a set of interventional distributions satisfies the $\mcI$-Markov property with respect to a graph if these distributions adhere to the invariance constraints imposed by the causal calculus rules of that graph. We first extend the causal calculus rules to operate between arbitrary sets of hard interventions. We say that two causal graphs, $\mathcal{D}_1$ and $\mathcal{D}_2$, are $\mathcal{I}$-Markov equivalent if the set of distributions that are $\mathcal{I}$-Markov to both $\mathcal{D}_1$ and $\mathcal{D}_2$ is the same. Using the augmented graph, we identify a graphical condition that is both necessary and sufficient for two Causal Bayesian Networks (CBNs) with latents to be $\mathcal{I}$-Markov equivalent under the framework of hard interventions. Finally, we propose a sound algorithm for causal discovery from 
hard-interventional datasets. 
Our main contributions can be summarized as follows:

\begin{itemize}
    \item We propose a characterization of $\mathcal{I}$-Markov equivalence between two causal graphs with latent variables for a given intervention set $\mathcal{I}$, based on a generalization of do-calculus under  hard interventions.
    \item We provide a graphical characterization of $\mathcal{I}$-Markov equivalence for causal graphs with latent variables under the framework of hard interventions.
    \item We introduce a learning algorithm for inferring the graphical structure using a combination of different interventional data, while utilizing the corresponding new constraints. This procedure includes a new set of orientation rules, and we formally prove its soundness.
\end{itemize}

\textbf{Outline of the paper:} In Section~\ref{sec: related works}, we briefly cover the necessary background knowledge and summarize the related works. In Section~\ref{sec: Do constraints}, we derive a set of do-constraints to combine the observational and interventional distributions. In Section~\ref{sec: IMEC}, we characterize the graph conditions for $\mcI$-Markov equivalence class. In Section~\ref{sec: intermediate}, we construct a more compact graphical structure to capture the characterization. In Section~\ref{sec: Learning algo}, we propose a learning algorithm that recovers the causal graph from given interventional distributions. In Section~\ref{sec: Exp}, we show some experimental results to compare the size of interventional Markov equivalence classes under hard interventions and soft interventions. In Section~\ref{sec: conclusion}, we conclude with a discussion of limitations and future extensions. Finally, we discuss the potential broader impact of our work. The related proofs and extra information are included in the Appendix.

\label{sec: intro}
\section{Background and Related Works}
\label{sec: related works}
In this section, we briefly describe related background knowledge and notations in this paper. Throughout this paper, we use upper case letters to denote variables, lower case letters to denote realizations, and bold letters for sets.

\textbf{Causal Bayesian Network (CBN):} Given a set of variables $\mV$, $P(\mv)$ represents the joint distribution for $\mV = \mv$. A hard intervention $do(\mX = \mathbf{x})$ refers to setting a subset $\mX \subseteq \mV$ to constants $\mx$. It breaks the causal relationship between the intervened variables and their parents. The interventional distribution is  $P_\mathbf{x}(\mv)$. Let $\mcP$ denote the tuple of all interventional distributions for all $\mX \subseteq \mV$. Then, a directed acyclic graph (DAG) $\mcD = (\mV, \mE)$ is said to be a causal Bayesian network compatible with $\mcP$ if and only if, for all $\mX \subseteq \mV, P_\mx(\mv) = \prod_{ \{i|V_i \notin \mX\} }P(v_i|pa_i)$, for all $\mv$ consistent with $\mx$, and where $\mathbf{pa}_i$ is the set of parents of $V_i$ in $\mcD$. $\mcD$ is said to be causal if it satisfies this condition. $V[\mcD]$ and $E[\mcD]$ denote the set of all nodes and all edges of graph $\mcD$ respectively. Causal graphs entail specific conditional independence (CI) relationships among observable variables via d-separation statements.  The d-separation serves as a criterion to determine whether a set of variables $\mX$ is independent of another set $\mY$, given $\mZ$.

If a causal graph has latent variables, it is denoted as $\mcD = (\mV \cup \mL, \mE)$ where $\mV$ represents observable variables, $\mL$ represents latent variables, and $\mE$ denotes the edges. If a latent variable $L\in \mL$ is a common cause of two observable variables, we use a curved bidirected edge between the two children variables. Such a causal graph is called Acyclic Directed Mixed Graph (ADMG). However, unlike causal graphs with sufficiency, the observed distribution is obtained by marginalizing $\mL$ out as the Markovian condition does not hold in this case.
\begin{equation}
    P(\mv) = \sum_{\mL} \prod _{ \{ i|T_i \in \mL \cup \mV  \}} P(t_i | pa_i)
\end{equation}
Two causal graphs are called Markov equivalent if they encode the same set of CI statements over $\mV$.

\textbf{Ancestral graphs:} 
Ancestral graph is a graphical representation widely used for a class of Markov equivalent causal graphs with latent variables. In an ADMG, $X$ is an ancestor of $Y$ if there is a directed path from $X$ to $Y$\footnote{We follow the convention that a node is an ancestor of itself.}. $X$ is a spouse of $Y$ if $X \leftrightarrow Y$ is present. An inducing path relative to $\mZ$ is a path on which every non-endpoint vertex $T\notin \mZ$ is a collider on the path (the two adjacent nodes are into it) and every collider is an ancestor of one of the endpoints. An ADMG is ancestral if it does not contain any almost directed cycle. It is maximal if there is no inducing path (relative to the empty set) between any pair of non-adjacent vertices. It is called a Maximal Ancestral Graph (MAG) if it is both maximal and ancestral~\citep{richardson2002ancestral}. In~\citet{zhang2008causal}, the authors show how to uniquely construct a MAG $\mcM_{\mcD}$ for a causal graph with latents $\mcD = (\mV \cup \mL, \mE)$, such that all the (conditional) independence statements and ancestral relationships over $\mV$ are preserved. Such CI statements are called m-separation statements in ADMGs.

In a graph $\mcD$, a triple $\langle X, Y, Z\rangle$ is unshielded if $X,Y$ are adjacent and $Y, Z$ are adjacent while $X, Z$ are not adjacent. If both edges are into $Y$, then it is an unshielded collider. A path between $X$ and $Y$, $p = \langle X, ..., W, Z, Y\rangle$, is a discriminating path for $Z$ if (1) $p$ includes at least three edges; (2) $Z$ is a non-endpoint node on $p$, and is adjacent to $Y$ on $p$; and (3) $X$ is not adjacent to $Y$, and every node between $X$ and $Z$ is a collider on $p$ and is a parent of $Y$. Two MAGs are Markov equivalent if and only if (1) they have the same skeleton; (2) they have the same unshielded colliders; and (3) if a path $p$ is a discriminating path for $Z$ in both MAGs, then Z is a collider on the path in one graph if and only if it is a collider on the path in the other. A partial ancestral graph (PAG) represents a Markov equivalence class of MAGs. It can be learned from CI statements over the observable variables under faithfulness. When observational data is provided, FCI algorithm is a commonly used algorithm used to recover the PAG and is proved to be sound and complete in~\citet{zhang2008completeness}.

\textbf{Related works:} There are many works in the literature~\citep{chickering2002optimal, hyttinen2013experiment, eberhardt2007causation, shanmugam2015learning, kocaoglu2017experimental} related to learning the causal structure from a combination of observational and interventional data. Under the assumption of sufficiency, ~\citet{hauser2012characterization, hauser2014two} introduced the Markov equivalence characterization. \citet{yang2018characterizing} further showed that the same characterization can be used for both hard and soft interventions. More works aimed at the cases where latents are present in the graph. If only observational data is available, \citet{zhang2008completeness} showed the property and proposed the sound and complete FCI algorithm to learn a PAG. \citet{spirtes2001causation, colombo2012learning, spirtes1991probability, colombo2014order, ghassami2018budgeted, kocaoglu2017experimental} proposed FCI-variant algorithms under different settings. \citet{kocaoglu2019characterization} introduced a characterization for $\mcI$-Markov equivalence class for soft interventions using augmented graphs with $F$ node and proposed an FCI-variant algorithm to learn it. Following this, \cite{jaber2020causal} characterized the $\Psi$-MEC for unknown soft interventions and proposed a learning algorithm. \citet{li2023causal} introduced the S-Markov property and learning algorithm when data from multiple domains are provided. 

\textbf{Notations:} For disjoint sets of variables $\mX, \mY, \mZ$, a CI statement \textit{'$\mX$ is independent of $\mY$ conditioning on $\mZ$'} is represented by $\mX \indep \mY | \mZ$. Similarly, in a causal graph $\mcD$, the d-separation statement \textit{'$\mX$ is independent of $\mY$ conditioning on $\mZ$ in graph $\mcD$'} is denoted as $(\mX \indep \mY | \mZ)_{\mcD}$. A set of interventions is $\mcI \subseteq 2^{\mV}$, where $2^{\mV}$ is the power set of $\mV$. For two interventions $\mI, \mJ \in \mcI$, the symmetric difference is $\mI \Delta \mJ := (\mI \setminus \mJ ) \cup (\mJ \setminus \mI)$. $\mcD_{\overline{X}}$/$\mcD_{\underline{X}}$ is the graph obtained by removing all the edges into/out of $\mX$ from $\mcD$. For $\mcD_{\overline{\mX, \mY(\mZ)}}$, $\mY(\mZ)$ is the subset of $\mY$ that are not ancestors of $\mZ$ in the graph $\mcD_{\overline{\mX}}$. In a PAG, a circle mark in an edge $X\crightarrow Y$ can be either an arrowtail or an arrowhead which is not determined. A star mark in an edge $X\srightarrow Y$ is used as a wildcard which can be a circle, arrowhead, or arrowtail. We assume that there is no selection bias. 
\section{Combining Observational and Experimental Distributions under Do-Calculus}
\label{sec: Do constraints}
One of the most renowned contributions to causal inference is the development of do-calculus (also known as causal calculus)~\cite{pearl1995causal, pearl2009causality}. Do-calculus is a set of three inference rules that enable the transformation of distributions associated with a causal graph. It leverages the graphical structure to determine when and how interventions can be adjusted or "removed" from expressions. In the context of hard interventions, the theorem is stated as follows\footnote{Here we put condition on $z$ for Rule 2. While this is redundant, we aim to show its clear connection with the corresponding $F$ node d-separations in the augmented graphs (see Section~\ref{sec: IMEC}), which requires conditioning.}:

\begin{theorem}
    \label{thm: do-calculus}
    (Theorem 3 in ~\citet{pearl1995causal}). Let $\mcD = (\mV \cup \mL, \mE)$ be a causal graph. Then the following statements hold for any distribution that is consistent with $\mcD$
    
    \textit{Rule 1 (see-see)}: For any $\mX \subseteq \mV$ and disjoint $\mY, \mZ, \mW \subseteq \mV$
    
    $P_{\mx}(\my|\mw, \mz) = P_{\mx}(\my|\mw)$, if $\mY \indep \mZ|\mW, \mX$ in $\mcD_{\overline{\mX}}$
    
    \textit{Rule 2 (do-see)}: For any disjoint $\mX, \mY, \mZ \subseteq \mV$ and $\mW \subseteq \mV \setminus (\mZ \cup \mY)$

    $P_{\mx, \mz}(\my|\mw, \mz) = P_{\mx}(\my|\mw, \mz)$, if $\mY \indep \mZ|\mW, \mX$ in $\mcD_{\overline{\mX}, \underline{\mZ}}$

    \textit{Rule 3 (do-do)}: For any disjoint $\mX, \mY, \mZ \subseteq \mV$ and $\mW \subseteq \mV \setminus (\mZ \cup \mY)$

    $P_{\mx, \mz}(\my | \mw) = P_{\mx}(\my|\mw)$, if $\mY \indep \mZ | \mW, \mX $ in $\mcD_{\overline{\mX \mZ(W)}}$

    where $\mZ(\mW)\subseteq \mZ$ are non-ancestors of $\mW$ in $\mcD_{ \overline{\mX} }$.
\end{theorem}
Similar to the observations in~\citet{kocaoglu2019characterization} that the converse of the rules can be utilized to derive insights of the graph structures, here we also need a set of statements for hard interventions. With soft intervention, the interventional graph remains the same since no causal relationship is broken by soft interventions. However, hard interventions induce changes to the causal graph, making it potentially more informative in learning the graph structure. The intuition is that with hard interventions, the causal graph becomes sparser and thus more do-invariance statements can be found to constrain the graph. Accordingly, we show the following proposition that characterizes the graph conditions from the invariance of two arbitrary intervention sets. Throughout the paper, for a pair of targets $\mI, \mJ$ and a conditioning set $\mW$, we define the following useful sets: $\mK = \mI \Delta \mJ, \mK_\mI = \mK \setminus \mJ, \mK_\mJ = \mK \setminus \mI, \mW_\mI = \mK_\mI \cap \mW, \mW_\mJ = \mK_\mJ \cap \mW, \mR = \mK \setminus \mW, \mR_\mI = \mR \cap \mK_\mI, \mR_\mJ = \mR \cap \mK_\mJ$. 
\begin{proposition}
    \label{prop: do-constraint}
    (Generalized do-calculus for hard interventions). Let $\mcD = (\mV \cup \mL, \mE)$ be a causal graph with latents. Then, the following holds for any tuple of hard-interventional distributions $(P_{\mI})_{ \mI \in \mcI }$ consistent with $\mcD$, where $\mcI \subseteq 2^{\mV}$.

    \textit{Rule 1 (conditional independence)}: For any $\mI \subseteq \mV$ and disjoint $\mY, \mZ, \mW \subseteq (\mV \setminus \mI)$
    
    $\quad P_{\mI}(\my | \mw, \mz) = P_{\mI}( \my | \mw)$, if \; $\mY\indep \mZ | \mW, \mI$ in $\mcD_{\overline{\mI}}$
    
    \textit{Rule 2 (do-see)}: For any $\mI, \mJ \subseteq \mV$ and disjoint $\mY, \mW \subseteq \mV \setminus \mK$, where $\mK := \mI \Delta \mJ$

    $\quad P_{\mI}(\my |\mw, \mk) = P_{\mI, \mJ}(\my|\mw, \mk) = P_{\mJ}( \my | \mw, \mk)$, if \; $( \mY\indep \mK_\mJ| \mW, \mI )_{ \mcD_{\overline{\mI}, \underline{\mK_\mJ}}} \land ( \mY \indep \mK_\mI|\mW, \mJ )_{ \mcD_{\overline{\mJ}, \underline{\mK_\mI}}}$

    \textit{Rule 3 (do-do)}: For any $\mI, \mJ \subseteq \mV$ and disjoint $\mY, \mW \subseteq \mV \setminus \mK$, where $\mK := \mI \Delta \mJ$

    $\quad P_{\mI}( \my | \mw) = P_{\mI, \mJ}(\my|\mw) = P_{\mJ}( \my | \mw)$, \; if \; $(\mY \indep \mK_\mJ | \mW, \mI)_{ \mcD_{\overline{\mI, \mK_\mJ(\mW)}}} \land ( \mY \indep \mK_\mI|\mW, \mJ)_{\mcD_{\overline{\mJ, \mK_\mI(\mW)}}}$
 
    \textit{Rule 4 (mixed do-see/do-do)}: For any $\mI, \mJ \subseteq \mV$ and disjoint $\mY, \mW \subseteq \mV$, where $\mK := \mI \Delta \mJ$

    $\quad P_\mI(\my | \mw) = P_{\mI, \mJ}(\my|\mw, \mk) = P_\mJ( \my | \mw)$, \; if \; $(\mY \indep \mR_\mJ|\mW, \
    \mI)_{\mcD_{\overline{\mI, \mR_\mJ(\mW)}}} \land ( \mY \indep \mW_\mJ | \mW, \mI)_{\mcD_{\overline{\mI}, \underline{\mW_\mJ}}} \land (\mY \indep \mR_\mI|\mW, \mJ)_{\mcD_{\overline{\mJ, \mR_\mI(\mW)}}} \land (\mY \indep \mW_\mI | \mW, \mJ)_{\mcD_{\overline{\mJ}, \underline{\mW_\mI}}}$
\end{proposition}
Note that Rule 2 and Rule 3 are special cases of Rule 4. We present all three to make the connection
to standard causal calculus rules more explicit. In the following sections, we will show how the generalized rules can be crucial in characterizing and learning the $\mathcal{I}$-Markov Equivalence Class ($\mcI$-MEC).

\section{$\mcI$-Markov Equivalence Class}
\label{sec: IMEC}
In this section, we will characterize the graphical conditions for interventional Markov equivalence class. First of all, we start by introducing the definition of interventional Markov equivalence based on the new do-constraint rules.
\begin{definition}
    \label{def: I-Markov}
    Consider the tuples of absolutely continuous probability distributions $(P_\mI)_{ \mI \in \mcI}$ over a set of variables $\mV$. A tuple $(P_\mI)_{ \mI \in \mcI}$ satisfies the $\mcI$-Markov property with respect to a causal graph $\mcD = (\mV \cup \mL, \mE)$ if the following holds for disjoint $\mY, \mZ, \mW \subseteq \mV$:
    \begin{enumerate}
        \item For $\mI \in \mcI$: $P_\mI( \my | \mw, \mz) = P_\mI( \my | \mw)$ \; if \; $\mY \indep \mZ| \mW, \mI$ in $\mcD_{\overline{\mI}}$
        \item For $\mI, \mJ \in \mcI$: $P_\mI( \my | \mw) = P_\mJ( \my | \mw)$ \; if $(\mY \mathrel{\perp\!\!\!\perp} \mR_J|\mW, \mI)_{\mcD_{\overline{\mI, \mR_\mJ(\mW)}}} \land (\mY \indep \mW_\mJ|\mW, \mI)_{\mcD_{\overline{\mI}, \underline{\mW_\mJ}}} \land \,\,(\mY \indep \mR_\mI| \mW, \mJ)_{\mcD_{ \overline{\mJ, \mR_\mI(\mW)}}} \land (\mY \indep \mW_\mI|\mW, \mJ)_{\mcD_{\overline{\mJ}, \underline{\mW_\mI}}}$
    \end{enumerate}

    The set of all tuples that satisfy the $\mcI$-Markov property with respect to $\mcD$ are denoted by $\mcP_{\mcI}(\mcD, \mV)$.
\end{definition}
The two conditions of $\mathcal{I}$-Markov property correspond to the first Rule in Theorem~\ref{thm: do-calculus} and Rule 4 in Proposition~\ref{prop: do-constraint} respectively. When $\mcI = \emptyset$, i.e. we only have access to observational distribution, this definition aligns with the well-known definition of Markov equivalence. It only implies the first condition on the observational distribution $P(\mV)$. Accordingly, two causal graphs are said to be $\mcI$-Markov equivalent if they induce the same constraints to the interventional distribution tuple which we formalize as follows:
\begin{definition}
    \label{def: I-Markov equivalence}
    Given two causal graphs $\mcD_1 = (\mV \cup \mL_1, \mE_1)$ and $\mcD_2 = (\mV \cup \mL_2, \mE_2)$, and a set of intervention targets $\mcI \subseteq 2^V$, $\mcD_1$ and $\mcD_2$ are $\mathcal{I}$-Markov equivalent if $\mcP_{\mcI}(\mcD_1, \mV) = \mcP_{\mcI}(\mcD_2, \mV)$.
\end{definition}
The challenge of checking the $\mathcal{I}$-Markov property in Definition~\ref{def: I-Markov} involves checking multiple graph conditions in different graph mutilations of $\mcD$.  In order to construct a more compact representation of $\mcD$ that captures all the graph conditions, we introduce the augmented pair graph defined as follows\footnote{Throughout this paper, we use the superscript for (a set of) nodes to denote the interventional domain.}.
\begin{definition}[Augmented Pair Graph]
    \label{def: augmented pair}
     Given a causal graph $\mcD = (\mV \cup \mL, \mE)$ and a set of intervention targets $\mcI \subseteq 2^V$, for a pair of interventions $\mI, \mJ \in \mcI$, $\mK = \mI \Delta \mJ$, the augmented pair graph of $\mcD$, denoted as $\Aug_{(\mathbf{I}, \mathbf{J})}(\mcD)$, is constructed as follows: $ \Aug_{(\mathbf{I}, \mathbf{J})}(\mcD) = (\mV^{(\mathbf{I})}\cup \mV ^{(\mathbf{J})} \cup \{F^{(\mathbf{I}, \mathbf{J})}\}, \mE^{(\mathbf{I})} \cup \mE^{(\mathbf{J})} \cup \mcE) $, where $F^{(\mathbf{I}, \mathbf{J})}$ is an auxiliary node with the superscript representing the pair of intervention targets it refers to, $\mE^{(\mathbf{I})} = E[\mcD_{\overline{I}}], \mE^{(\mathbf{J})} = E[\mcD_{\overline{J}}], \mcE = \{ (F^{(\mathbf{\mI}, \mathbf{\mJ})}, S)\}_{S\in \mK^{(\mathbf{\mI})} \cup \mK^{(\mathbf{\mJ})}}$, with $S$ as a singleton.
\end{definition}
In words, for each pair of interventions $\mI, \mJ$, we create the augmented pair graph by creating two copies of vertices $\mV^{(\mathbf{I})}, \mV^{(\mathbf{J})}$ and adding the edges between the vertices with those in the corresponding interventional graphs $\mcD_{\overline{\mI}}, \mcD_{\overline{\mJ}}$, and then connecting the auxiliary node $F$ to all the nodes in the symmetric difference of $\mI, \mJ$. We will omit the subscript for the graph and superscripts for $F$ node when the pair of interventions is clear from the context. This kind of construction has been proposed and used in the causality literature before~\citep{eberhardt2007interventions, hauser2012characterization, pearl2009causality, dawid2002influence}. The constructed augmented pairs allow us to test the m-separation statements as listed in Definition~\ref{def: I-Markov} without looking into mutilations of the original graph $\mcD$. This is illustrated by the following Proposition.
\begin{proposition}
    \label{prop: F node}
    Given a causal graph $\mcD = (\mV \cup \mL, \mE)$ and a set of intervention targets $\mcI \subseteq 2^V$, for each pair of interventions $\mI, \mJ \in \mcI$, $\mK = \mI \Delta \mJ$, and the corresponding augmented pair graph $ \Aug_{(\mathbf{I}, \mathbf{J})}(\mcD) = (\mV^{(\mathbf{I})}\cup \mV ^{(\mathbf{J})} \cup \{F^{(\mathbf{I}, \mathbf{J})}\}, \mE^{(\mathbf{I})} \cup \mE^{(\mathbf{J})} \cup \mcE), \mcE = \{ (F^{(\mathbf{I}, \mathbf{J})}, S)\}_{S\in K^{(\mathbf{I})} \cup K^{(\mathbf{J})}}$, we have the following equivalence statements:

    For disjoint $\mY, \mZ, \mW \subseteq \mV$:
    \begin{equation}
        (\mY \indep \mZ |\mW, \mI)_{\mcD_{\overline{\mI}}} \iff (\mY \indep \mZ |\mW, \mI, F^{(\mI, \mJ)})_{ \Aug_{(\mathbf{I}, \mathbf{J})}(\mcD)}
    \end{equation}
    
    For disjoint $\mY, \mZ, \mW \subseteq \mV$:
    \begin{equation}
        \begin{split}
        \left
        \{
        \begin{array}{ll}
          (\mY\indep \mR_\mJ|\mW, \mI)_{\mcD_{\overline{\mI, \mR_\mJ(\mW)}}}\\
          (\mY\indep \mW_\mJ|\mW \setminus \mW_\mJ, \mI)_{\mcD_{\overline{\mI}, \underline{\mW_\mJ}}}\\
          (\mY\indep \mR_\mI|\mW, \mJ)_{\mcD_{\overline{\mJ, \mR_\mI(\mW)}}} \\
          (\mY\indep \mW_\mI|\mW \setminus \mW_\mI, \mJ)_{\mcD_{\overline{\mJ}, \underline{\mW_\mI}}}
        \end{array}
        \right. \\
        \iff 
        \left
            \{
        \begin{array}{cc}
            (F^{(\mI, \mJ)} \indep \mY^{(\mathbf{I})} | \mI^{(\mathbf{I})}, \mW^{(\mathbf{I})})_{\Aug_{(\mathbf{I}, \mathbf{J})}(\mcD)} \\
            (F^{(\mI, \mJ)} \indep \mY^{(\mathbf{J})} | \mJ^{(\mathbf{J})}, \mW^{(\mathbf{J})})_{\Aug_{(\mathbf{I}, \mathbf{J})}(\mcD)}
        \end{array}
        \right.
        \end{split}
    \end{equation}
\end{proposition}

\begin{figure*}[ht]
    \centering
    \begin{subfigure}{0.24\columnwidth}
        \centering
        \begin{tikzpicture}
            \node (1) {{$X$}};
            \node (2) [right of = 1,xshift = 0.6 cm] {{$Z$}};
              \node (3) [right of = 2,xshift = 0.6 cm] { {$Y$}};    

        \draw[->] (1) -- (2) ;
        \draw[->] (2) -- (3) ;
        \draw[<->] (2) edge[bend left=50] (3); 
        \end{tikzpicture}
        \caption{Graph $\mcD_1$}
        \label{fig: D_1}
    \end{subfigure}%
   \begin{subfigure}{0.24\columnwidth}
        \centering
        \begin{tikzpicture}
            \node (1) {{$X^{(1)}$}};
            \node (2) [right of = 1,xshift = 0.6 cm] {{$Z^{(1)}$}};
              \node (3) [right of = 2,xshift = 0.6 cm] { {$Y^{(1)}$}};    
              \node (4) [below of = 2, yshift = -0.1 cm] {$F$};
              \node (5) [below of = 4, yshift= -0.1 cm] {$Z^{(2)}$};
              \node (6) [left of = 5, xshift= -0.6 cm] {$X^{(2)}$};
              \node (7) [right of = 5, xshift= 0.6 cm] {$Y^{(2)}$};

        \draw[->] (1) -- (2) ;
        \draw[->] (2) -- (3) ;
        \draw[<->] (2) edge[bend left=50] (3); 
        \draw[->] (4) -- (2) ;
        \draw[->] (4) -- (5) ;
        \draw[->] (5) -- (7) ;
        \end{tikzpicture}
        \caption{$\Aug_{(\emptyset, \{ Z\})}(\mcD_1)$}
        \label{fig: Aug_1}
    \end{subfigure}%
    \quad
      \begin{subfigure}{0.24\columnwidth}
        \centering
        \begin{tikzpicture}
            \node (1) {{$X^{(1)}$}};
            \node (2) [right of = 1,xshift = 0.6 cm] {{$Z^{(1)}$}};
              \node (3) [right of = 2,xshift = 0.6 cm] { {$Y^{(1)}$}};    
              \node (4) [below of = 2, yshift = -0.1 cm] {$F$};
              \node (5) [below of = 4, yshift= -0.1 cm] {$Z^{(2)}$};
              \node (6) [left of = 5, xshift= -0.6 cm] {$X^{(2)}$};
              \node (7) [right of = 5, xshift= 0.6 cm] {$Y^{(2)}$};

        \draw[->] (1) -- (2) ;
        \draw[->] (2) -- (3) ;
        \draw[->] (1) edge[bend left=50] (3); 
        \draw[->] (4) -- (2) ;
        \draw[->] (4) -- (5) ;
        \draw[->] (5) -- (7) ;
        \draw[->] (4) -- (3) ;
        \end{tikzpicture}
        \caption{$\MAG(\Aug_{(\emptyset, \{ Z\})}(\mcD_1))$}
        \label{fig: mag_1}
    \end{subfigure}%
    \begin{subfigure}{0.24\columnwidth}
        \centering
        \begin{tikzpicture}
            \node (1) {{$X^{(1)}$}};
            \node (2) [right of = 1,xshift = 0.6 cm] {{$Z^{(1)}$}};
              \node (3) [right of = 2,xshift = 0.6 cm] { {$Y^{(1)}$}};    
              \node (4) [below of = 2, yshift = -0.1 cm] {$F$};
              \node (5) [below of = 4, yshift= -0.1 cm] {$Z^{(2)}$};
              \node (6) [left of = 5, xshift= -0.6 cm] {$X^{(2)}$};
              \node (7) [right of = 5, xshift= 0.6 cm] {$Y^{(2)}$};

        \draw[->] (1) -- (2) ;
        \draw[->] (2) -- (3) ;
        \draw[->] (1) edge[bend left=50] (3); 
        \draw[->] (4) -- (2) ;
        \draw[->] (4) -- (5) ;
        \draw[->] (5) -- (7) ;
        \draw[->] (4) -- (3) ;
        \draw[->] (4) -- (7) ;
        \end{tikzpicture}
        \caption{$\Twin_{(\emptyset, \{ Z\})}(\mcD_1)$}
        \label{fig: twin_1}
    \end{subfigure}%

     \begin{subfigure}{0.24\columnwidth}
        \centering
        \begin{tikzpicture}
            \node (1) {{$X$}};
            \node (2) [right of = 1,xshift = 0.6 cm] {{$Z$}};
              \node (3) [right of = 2,xshift = 0.6 cm] { {$Y$}};    
        \draw[->] (1) edge[bend left=70] (3) ;
        \draw[->] (1) -- (2) ;
        \draw[->] (2) -- (3) ;
        \draw[<->] (2) edge[bend left=40] (3); 
        \end{tikzpicture}
        \caption{Graph $\mcD_2$}
        \label{fig: D_2}
    \end{subfigure}%
   \begin{subfigure}{0.24\columnwidth}
        \centering
        \begin{tikzpicture}
            \node (1) {{$X^{(1)}$}};
            \node (2) [right of = 1,xshift = 0.6 cm] {{$Z^{(1)}$}};
              \node (3) [right of = 2,xshift = 0.6 cm] { {$Y^{(1)}$}};    
              \node (4) [below of = 2, yshift = -0.1 cm] {$F$};
              \node (5) [below of = 4, yshift= -0.1 cm] {$Z^{(2)}$};
              \node (6) [left of = 5, xshift= -0.6 cm] {$X^{(2)}$};
              \node (7) [right of = 5, xshift= 0.6 cm] {$Y^{(2)}$};
        \draw[->] (6) edge[bend right=30] (7) ;
        \draw[->] (1) edge[bend left=60] (3) ;
        \draw[->] (1) -- (2) ;
        \draw[->] (2) -- (3) ;
        \draw[<->] (2) edge[bend left=40] (3); 
        \draw[->] (4) -- (2) ;
        \draw[->] (4) -- (5) ;
        \draw[->] (5) -- (7) ;
        \end{tikzpicture}
        \caption{$\Aug_{(\emptyset, \{ Z\})}(\mcD_2)$}
        \label{fig: Aug_2}
    \end{subfigure}%
      \begin{subfigure}{0.24\columnwidth}
        \centering
        \begin{tikzpicture}
            \node (1) {{$X^{(1)}$}};
            \node (2) [right of = 1,xshift = 0.6 cm] {{$Z^{(1)}$}};
              \node (3) [right of = 2,xshift = 0.6 cm] { {$Y^{(1)}$}};    
              \node (4) [below of = 2, yshift = -0.1 cm] {$F$};
              \node (5) [below of = 4, yshift= -0.1 cm] {$Z^{(2)}$};
              \node (6) [left of = 5, xshift= -0.6 cm] {$X^{(2)}$};
              \node (7) [right of = 5, xshift= 0.6 cm] {$Y^{(2)}$};
        \draw[->] (6) edge[bend right=30] (7) ;
        \draw[->] (1) -- (2) ;
        \draw[->] (2) -- (3) ;
        \draw[->] (1) edge[bend left=50] (3); 
        \draw[->] (4) -- (2) ;
        \draw[->] (4) -- (5) ;
        \draw[->] (5) -- (7) ;
        \draw[->] (4) -- (3) ;
        \end{tikzpicture}
        \caption{$\MAG(\Aug_{(\emptyset, \{ Z\})}(\mcD_2))$}
        \label{fig: mag_2}
    \end{subfigure}%
    \begin{subfigure}{0.24\columnwidth}
        \centering
        \begin{tikzpicture}
            \node (1) {{$X^{(1)}$}};
            \node (2) [right of = 1,xshift = 0.6 cm] {{$Z^{(1)}$}};
              \node (3) [right of = 2,xshift = 0.6 cm] { {$Y^{(1)}$}};    
              \node (4) [below of = 2, yshift = -0.1 cm] {$F$};
              \node (5) [below of = 4, yshift= -0.1 cm] {$Z^{(2)}$};
              \node (6) [left of = 5, xshift= -0.6 cm] {$X^{(2)}$};
              \node (7) [right of = 5, xshift= 0.6 cm] {$Y^{(2)}$};
        \draw[->] (6) edge[bend right=30] (7) ;
        \draw[->] (1) -- (2) ;
        \draw[->] (2) -- (3) ;
        \draw[->] (1) edge[bend left=50] (3); 
        \draw[->] (4) -- (2) ;
        \draw[->] (4) -- (5) ;
        \draw[->] (5) -- (7) ;
        \draw[->] (4) -- (3) ;
        \draw[->] (4) -- (7) ;
        \end{tikzpicture}
        \caption{$\Twin_{(\emptyset, \{ Z\})}(\mcD_2)$}
        \label{fig: twin_2}
    \end{subfigure}%
    \caption{Illustration of the construction of twin augmented MAGs where $\mcD_1$ and  $\mcD_2$ are not  $\mathcal{I}-$Markov equivalent. (a) and (e) are two causal graphs, $\mcD_1$ and $\mcD_2$ respectively, given intervention targets $\mcI = \{ \mI_1 = \emptyset, \mI_2 = \{Z \} \}$. (b) and (f) are the augmented pair graphs for $\mcD_1$ and $\mcD_2$ respectively. (c) and (g) are the MAG of the augmented pair graphs for $\mcD_1$ and $\mcD_2$ respectively. (d) and (h) are the twin augmented MAGs for $\mcD_1$ and $\mcD_2$ respectively. $F\rightarrow Y^{(1)}$ in $\MAG(\Aug_{(\emptyset, \{ Z\})}(\mcD_1))$ and $\MAG(\Aug_{(\emptyset, \{ Z\})}(\mcD_2))$ because there is an inducing path $\langle F, Z^{(1)}, Y^{(1)} \rangle$ in both augmented pair graphs. In the twin augmented graphs, we further add $F\rightarrow Y^{(2)}$ to make the adjacencies around the $F$ node symmetric.}
    \label{fig: ex of twin}
\end{figure*}
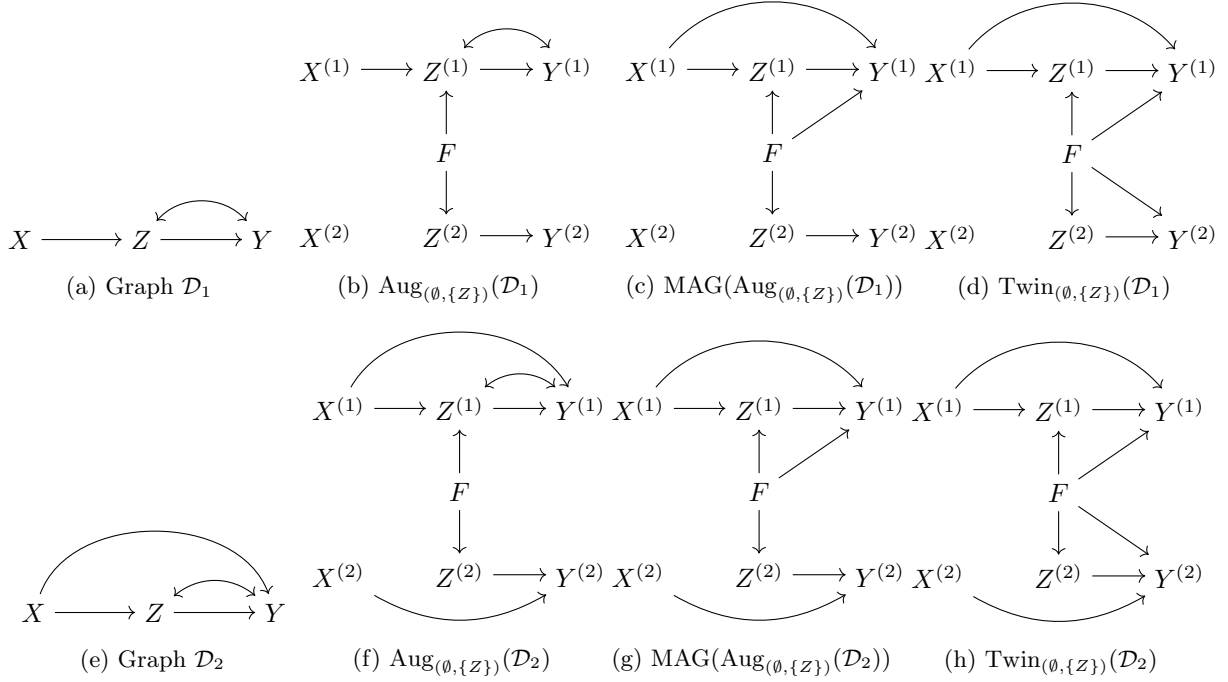

While the augmented pair graphs encode the same CI statements as the original graph, we know that different graphs may entail the same CI statements. To characterize the $\mcI$-Markov equivalence, we utilize the structure of Maximal Ancestral Graphs (MAGs). MAGs represent the Markov equivalence class of the original graph, making it possible to compare and analyze equivalence classes without needing the full graph with latent variables. We introduce the following definition to construct a graph structure that captures the $\mcI$-Markov equivalence.
\begin{definition}
    \label{def: twin augmented mag}
    (Twin Augmented MAG). Given a causal graph $\mcD = (\mV \cup \mL, \mE)$ and a set of interventions $\mcI \subseteq 2^V$, for each pair of intervention targets $\mI, \mJ \in \mcI$, $\mK = \mI \Delta \mJ$, and the corresponding augmented pair graph $\Aug_{(\mathbf{I}, \mathbf{J})}(\mcD) = (\mV^{(\mathbf{I})}\cup \mV ^{(\mathbf{J})} \cup \{F^{(\mathbf{I}, \mathbf{J})}\}, \mE^{(\mathbf{I})} \cup \mE^{(\mathbf{J})} \cup \mcE), \mcE = \{ (F^{(\mathbf{I}, \mathbf{J})}, S)\}_{S\in K^{(\mathbf{I})} \cup K^{(\mathbf{J})}} $, construct the MAG\footnote{We use the conventional steps to construct the MAG from an ADMG: For each pair of nodes $X, Y$, if $X$ is $Y$'s ancestor/descendant/spouse and there is an inducing path between them, we orient $X\rightarrow Y/X\leftarrow Y/X\leftrightarrow Y$ between them, otherwise they are not adjacent.} of the augmented pair graph and denote it as $\MAG(\Aug_{(\mathbf{I}, \mathbf{J})}(\mcD))$. The twin augmented MAG, denoted as $\Twin_{(\mathbf{I}, \mathbf{J})}(\mcD)$, is constructed by adding edges $(F, S^{(\mathbf{I})})$ and $(F, S^{(\mathbf{J})})$ to $\MAG(\Aug_{(\mathbf{I}, \mathbf{J})}(\mcD))$ if for the singleton $S\in \mV$,  $S^{(\mathbf{I})}$ or $S^{(\mathbf{J})}$ is adjacent to $F$ in $\MAG(\Aug_{(\mathbf{I}, \mathbf{J})}(\mcD))$.
\end{definition}

\begin{lemma}
    \label{lm: MAG of twin mag}
    Twin augmented MAGs are valid MAGs.
\end{lemma}
\textit{Proof.} The extra cycles contain $F$, while $F$ has only outgoing edges. Thus, the twin augmented MAGs are ancestral. Suppose there is an inducing path $\langle F, X_1, X_2, ..., X_k\rangle$, $F$ and $X_k$ are not adjacent. Then $X_{k-1}$ has a directed edge to $X_k$ in $\MAG(\Aug(\mcD))$ while $\MAG(\Aug(\mcD))$ is a MAG by definition. Thus, a contradiction arises, and the supposition does not hold.  The twin augmented MAGs are maximal \qed

The motivation of adding extra edges to $F$ nodes comes from the fact that $F\indep Y^{(\mI)}$ itself is not testable by comparing the invariances using $P_\mI, P_\mJ$. It requires access to $P_{\mI, \mJ}$ which is not necessarily given. Therefore, when the invariance does not hold, we cannot distinguish if $F$ is non-separable to $Y$ in only one domain or in both domains. Next, we give a graphical characterization of $\mcI$-Markov equivalence between causal graphs using Definition~\ref{def: twin augmented mag}. 

\begin{theorem}
    \label{thm: I-MEC}
    Given two causal graphs $\mcD_1 = (\mV \cup \mL_1, \mE_1)$ and $\mcD_2 = (\mV \cup \mL_2, \mE_2)$, and a set of intervention targets $\mcI \subseteq 2^V$, $\mcD_1$ and $\mcD_2$ are $\mcI$-Markov equivalent with respect to $\mcI$ if and only if for each pair of interventions $\mI, \mJ \in \mcI$, $\mcM_1 = \Twin_{(\mathbf{I}, \mathbf{J})}(\mcD_1), \mcM_2 = \Twin_{(\mathbf{I}, \mathbf{J})}(\mcD_2)$:
    \begin{enumerate}
        \item $\mcM_1$ and $\mcM_2$ have the same skeleton;
        \item $\mcM_1$ and $\mcM_2$ have the same unshielded colliders;
        \item If a path $p$ is a discriminating path for a node $Y$ in both $\mcM_1$ and $\mcM_2$, then $Y$ is a collider on the path if and only if it is a collider on the path in the other.
    \end{enumerate}
\end{theorem}

To illustrate the concepts, we construct the examples in Figure~\ref{fig: ex of twin}. Graph $\mcD_1$ in Figure~\ref{fig: D_1} is the original graph with a confounder between $Z, Y$. Assume that the intervention set is $\mcI = \{ \mI_1 = \emptyset, \mI_2 =  \{Z\} \}$. Here we use the index of the targets in the superscripts of the variables for simplicity. Figure~\ref{fig: Aug_1} shows how we build the augmented pair graph by adding the $F$ node and pointing it to both $Z$s in the two interventional subgraphs. In Figure~\ref{fig: mag_1}, we show the MAG of $\Aug_{(\emptyset, \{ Z\})}(\mcD_1)$. The edge of $(F, Y^{(1)})$ is added because there is an inducing path $\langle F, Z^{(1)}, Y^{(1)} \rangle$ in $\Aug_{(\emptyset, \{ Z\})}(\mcD_1)$, which means that $F$ is not m-separable from $Y^{(1)}$ and $F$ is an ancestor of $Y^{(1)}$. Now that $(F, Y^{(1)})$ is presented in $\MAG(\Aug_{(\emptyset, \{ Z\})}(\mcD_1))$ but not $(F, Y^{(2)})$, according to Definition~\ref{def: twin augmented mag}, we also add the edge of $(F, Y^{(2)})$ to finally construct the twin augmented MAG of $\mcD_1$, which is $\Twin_{(\emptyset, \{ Z\})}(\mcD_1)$.

Next, we repeat the same process for $\mathcal{D}_2$ as illustrated in Figure~\ref{fig: D_2}. Comparing $\Twin_{(\emptyset, \{ Z\})}(\mathcal{D}_1)$ and $\Twin_{(\emptyset, \{ Z\})}(\mathcal{D}_2)$, we see that they do not satisfy the three conditions from Theorem~\ref{thm: I-MEC} since they have different skeletons. Therefore, we conclude that $\mathcal{D}_1$ and $\mathcal{D}_2$ are not $\mathcal{I}$-Markov equivalent with respect to the given intervention set $\mcI$. 
 This example also highlights that hard interventions can be more informative than soft interventions because, under a soft intervention on $Z$, one cannot distinguish the two graphs, and they will be \(\mathcal{I}\)-Markov equivalent using the results from \cite{kocaoglu2019characterization}. However, with a hard intervention, we can distinguish the two graphs $\mathcal{D}_1$ and $\mathcal{D}_2$ from one another.

\section{$\mcI$-Augmented MAG}
\label{sec: intermediate}
We have demonstrated that the characterization of $\mcI$-Markov equivalence between two causal graphs can be effectively captured intervetional distributionsusing the proposed twin augmented MAGs. However, for an intervention target set of size $k$, we will have to inspect all  \(\binom{k}{2}\) such structures. This is undesirable, as there exists only one underlying causal graph, and a more compact graph representation is preferred. 
Additionally, each twin augmented MAG encodes information for only a single pair of distributions, whereas we aim for an objective that encapsulates as much information as possible. 
To address these challenges, we propose a new graphical structure defined as follows:
\begin{definition}[$\mcI$-augmented MAG]
    \label{def: I aug mag}
     Given a causal graph $\mcD = (\mV, \mE)$ and a set of intervention targets $\mcI$, construct all twin augmented MAGs $\Twin_{(\mathbf{I}, \mathbf{J})}(\mcD)$ for all $\mJ \in \mcI \setminus \{\mI\}$. For each $\mI \in \mcI$, the $\mcI$-augmented MAG related to $\mcI$ is defined as $\Aum_\mI(\mcD, \mcI) = (\mV \cup \mcF, E(\mcD_{\overline{\mI}}) \cup \mcE ) $ where $\mcF = \{F^{(\mI, \mJ)}\}_{\mJ \in \mcI \setminus \{\mI\} }, \mcE = \{ (F^{(\mI, \mJ)}, X^{(\mI)} )\}_{ (F^{(\mI, \mJ)}, X^{(\mI)} ) \in E(\Twin_{(\mI, \mJ) }(\mcD)), \mJ \in \mcI \setminus \{ \mI \} }$. In other words, it is the graph union of each twin augmented MAG's induced subgraph on $\mV^{(\mI)} \cup \{F^{(\mI, \mJ)} \}$.  \\
     The $\mcI$-augmented MAG tuple $\mathcal{N}_{\mcI}(\mcD)$ is a tuple of all $\mcI$-augmented MAGs $\mathcal{N}_{\mcI}(\mcD) = (\Aum_\mI(\mcD, \mcI))_{\mI \in \mcI}$.
\end{definition}

\textbf{Remark: } The $\mcI$-augmented MAG $\Aum_\mI(\mcD, \mcI)$ preserves all the m-separation statements in the domain of $do(\mI)$ from the twin augmented MAGs with $\mI$ in the intervention pair, $\Twin_{(\mathbf{I}, \mathbf{J})}(\mcD), \mJ \in \mcI\setminus \{\mI\}$. The structure of $\mV^{(\mI)} \cup F$ within each twin augmented MAG is preserved in the $\mcI$-augmented MAG and is not affected by the other domains.

The constructed \( \mathcal{I} \)-augmented MAG tuple consists of only \( k \) graphs, each of which encapsulates more information on the domain than a twin augmented MAG. Specifically, the set of ADMGs consistent with a twin augmented MAG in one domain is a superset of those consistent with an \( \mcI \)-augmented MAG, as the \( \mcI \)-augmented MAG imposes stricter constraints of separations across other domains on the causal graph. Furthermore, the graphical conditions for two causal graphs to be \(\mcI \)-Markov equivalent, as stated in Theorem~\ref{thm: I-MEC}, remain valid when using the \(\mcI \)-augmented MAG. Hence, the \( \mathcal{I} \)-augmented MAG serves as the unified and compact graphical representation.  

\begin{figure*}[ht]
    \centering
 \begin{subfigure}{0.32\columnwidth}
        \centering
        \begin{tikzpicture}
            \node (1) {{$X^{(1)}$}};
            \node (2) [right of = 1,xshift = 0.6 cm] {{$Z^{(1)}$}};
              \node (3) [right of = 2,xshift = 0.6 cm] { {$Y^{(1)}$}};    
              \node (4) [below of = 2, yshift = -0.1 cm] {$F^{(1,2)}$};
              \node (5) [below of = 4, yshift= -0.1 cm] {$Z^{(2)}$};
              \node (6) [left of = 5, xshift= -0.6 cm] {$X^{(2)}$};
              \node (7) [right of = 5, xshift= 0.6 cm] {$Y^{(2)}$};

        \draw[->] (1) -- (2) ;
        \draw[->] (2) -- (3) ;
        \draw[->] (1) edge[bend left=50] (3); 
        \draw[->] (6) edge[bend right=50] (7); 
        \draw[->] (6) -- (5) ;
        \draw[->] (4) -- (1) ;
        \draw[->] (4) -- (6) ;
        \draw[->] (5) -- (7) ;
        \end{tikzpicture}
        \caption{ $\Twin_{(\emptyset,\{X\})}(\mcD_1)$}
        \label{fig: twin_1 X empty}
    \end{subfigure}%
\begin{subfigure}{0.32\columnwidth}
    \centering
    \begin{tikzpicture}
        \node (1) {{$X^{(1)}$}};
        \node (2) [right of = 1,xshift = 0.6 cm] {{$Z^{(1)}$}};
          \node (3) [right of = 2,xshift = 0.6 cm] { {$Y^{(1)}$}};    
          \node (4) [below of = 2, yshift = -0.1 cm] {$F^{(1,3)}$};
          \node (5) [below of = 4, yshift= -0.1 cm] {$Z^{(3)}$};
          \node (6) [left of = 5, xshift= -0.6 cm] {$X^{(3)}$};
          \node (7) [right of = 5, xshift= 0.6 cm] {$Y^{(3)}$};

    \draw[->] (1) -- (2) ;
    \draw[->] (2) -- (3) ;
    \draw[->] (1) edge[bend left=50] (3); 
    \draw[->] (4) -- (2) ;
    \draw[->] (4) -- (5) ;
    \draw[->] (5) -- (7) ;
    \draw[->] (4) -- (3) ;
    \draw[->] (4) -- (7) ;
    \end{tikzpicture}
    \caption{$\Twin_{(\emptyset,\{Z\})}(\mcD_1)$}
    \label{fig: twin_1 empty Z}
\end{subfigure}%
    \begin{subfigure}{0.32\columnwidth}
    \centering
    \begin{tikzpicture}
        \node (1) {{$X^{(2)}$}};
        \node (2) [right of = 1,xshift = 0.6 cm] {{$Z^{(2)}$}};
          \node (3) [right of = 2,xshift = 0.6 cm] { {$Y^{(2)}$}};    
          \node (4) [below of = 2, yshift = -0.1 cm] {$F^{(2,3)}$};
          \node (5) [below of = 4, yshift= -0.1 cm] {$Z^{(3)}$};
          \node (6) [left of = 5, xshift= -0.6 cm] {$X^{(3)}$};
          \node (7) [right of = 5, xshift= 0.6 cm] {$Y^{(3)}$};

    \draw[->] (1) -- (2) ;
    \draw[->] (2) -- (3) ;
    \draw[->] (1) edge[bend left=50] (3); 
    \draw[->] (4) -- (2) ;
    \draw[->] (4) -- (5) ;
    \draw[->] (5) -- (7) ;
    \draw[->] (4) -- (3) ;
    \draw[->] (4) -- (7) ;
    \draw[->] (4) -- (1) ;
    \draw[->] (4) -- (6) ;
    \end{tikzpicture}
    \caption{$\Twin_{(\{X\},\{Z\})}(\mcD_1)$}
    \label{fig: twin_1 X Z}
\end{subfigure}%

 \begin{subfigure}{0.32\columnwidth}
    \centering
    \begin{tikzpicture}
      \node (4) [below of = 2, yshift = -0.1 cm] {$F^{(1, 2)}$};
      \node (5) [below of = 4, yshift= -0.1 cm] {$Z^{(1)}$};
      \node (6) [left of = 5, xshift= -0.6 cm] {$X^{(1)}$};
      \node (7) [right of = 5, xshift= 0.6 cm] {$Y^{(1)}$};
    \node (12) [below of = 5, yshift = -0.25 cm] {$F^{(1, 3)}$};
    \draw[->] (6) edge[bend right=25] (7); 
    \draw[->] (4) -- (6) ;
    \draw[->] (5) -- (7) ;
    \draw[->] (12) -- (6) ;
    \draw[->] (6) -- (5) ;

    \end{tikzpicture}
    \caption{ $\Aum_{\emptyset}(\mcD_1, \mcI)$}
    \label{fig: Aum D_1 empty}
\end{subfigure}%
 \begin{subfigure}{0.32\columnwidth}
    \centering
    \begin{tikzpicture}
          \node (4) [below of = 2, yshift = -0.1 cm] {$F^{(1, 2)}$};
          \node (5) [below of = 4, yshift= -0.1 cm] {$Z^{(2)}$};
          \node (6) [left of = 5, xshift= -0.6 cm] {$X^{(2)}$};
          \node (7) [right of = 5, xshift= 0.6 cm] {$Y^{(2)}$};
          \node (12) [below of = 5, yshift = -0.25 cm] {$F^{(2, 3)}$};

    \draw[->] (6) edge[bend right=25] (7); 
    \draw[->] (4) -- (6) ;
    \draw[->] (6) -- (5) ;
    \draw[->] (5) -- (7) ;
    \draw[->] (12) -- (5) ;
    \draw[->] (12) -- (7) ;
      \draw[->] (12) -- (6) ;
    \end{tikzpicture}
    \caption{ $\Aum_{\{X\}}(\mcD_1, \mcI)$}
    \label{fig: Aum D_1 X}
\end{subfigure}%
 \begin{subfigure}{0.32\columnwidth}
    \centering
    \begin{tikzpicture}
      \node (4) [below of = 2, yshift = -0.1 cm] {$F^{(1, 3)}$};
      \node (5) [below of = 4, yshift= -0.1 cm] {$Z^{(3)}$};
      \node (6) [left of = 5, xshift= -0.6 cm] {$X^{(3)}$};
      \node (7) [right of = 5, xshift= 0.6 cm] {$Y^{(3)}$};
      \node (12) [below of = 5, yshift = -0.25 cm] {$F^{(2, 3)}$};
      
    \draw[->] (4) -- (5) ;
    \draw[->] (4) -- (7) ;
    \draw[->] (5) -- (7) ;
    \draw[->] (12) -- (6) ;
    \draw[->] (12) -- (5) ;
    \draw[->] (12) -- (7) ;

    \end{tikzpicture}
    \caption{ $\Aum_{\{Z\}}(\mcD_1, \mcI)$}
    \label{fig: Aum D_1 Z}
\end{subfigure}%
    
    \caption{Illustration of the construction of $\mathcal{I-}$augmented MAGs from twin augmented MAGs. Figure~\ref{fig: D_1} is the ground truth graph. The intervention targets are $\mcI = \{ \mI_1 = \emptyset, \mI_2 = \{X\}, \mI_3 = \{Z\} \}$. (a), (b), and (c) are the twin augmented MAGs. (d), (e), and (f) are the $\mcI$-augmented MAGs.}
    \label{fig: I aug mag}
\end{figure*}
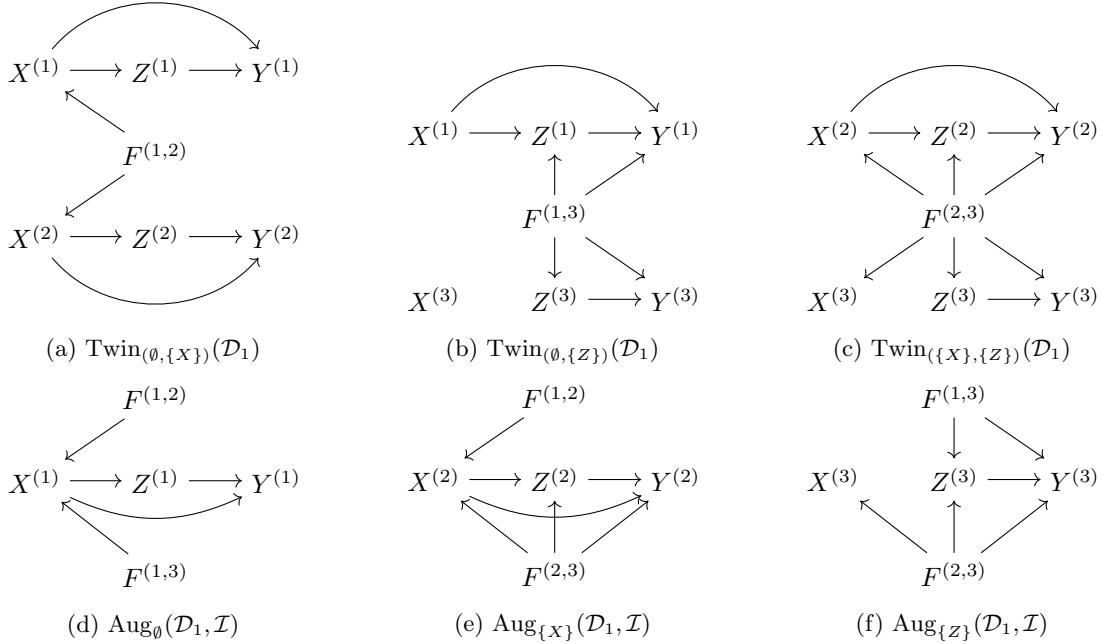

In Figure~\ref{fig: I aug mag}, we demonstrate the construction of \( \mathcal{I} \)-augmented MAGs with respect to three datasets, including observational data and interventions on \( X \) and \( Z \), i.e., \( \mathcal{I} = \{ \mI_1 = \emptyset, \mI_2 = \{X\}, \mI_3 = \{Z\}\} \), for the graph \( \mathcal{D}_1 \) from Figure~\ref{fig: D_1}. For simplicity, we relabel the observational domain and the interventional domains for targets \(X\) and \(Z\) as 1, 2, and 3, respectively, in the $\mcI$-augmented MAGs shown in Figure~\ref{fig: I aug mag}. Figure~\ref{fig: twin_1 X empty}, Figure~\ref{fig: twin_1 empty Z}, and Figure~\ref{fig: twin_1 X Z} are the twin augmented MAGs given $(\emptyset, \{X\}), (\emptyset, \{Z\}),$ and $(\{X\}, \{Z\})$ respectively. Based on the twin augmented MAGs, we construct the $\mcI$-augmented MAGs as shown in Figure~\ref{fig: Aum D_1 empty}, Figure~\ref{fig: Aum D_1 X}, and Figure~\ref{fig: Aum D_1 Z} for the domains $\emptyset, \{X\}$ and $\{Z\}$ respectively. Each $\mcI$-augmented MAG has the domain-specific skeleton in the center with the $F$ nodes around it indicating the invariances from other domains. The $\mcI$-augmented MAGs entail the same information about the causal graph as the twin augmented MAGs, but they have a much more compact representation. We show the equivalence between twin augmented MAGs and $\mcI$-augmented MAGs with the following proposition.

\begin{proposition}
    \label{prop: I aug mag I markov equivalent}
    Given two causal graphs $\mcD_1 = (\mV \cup \mL_1, \mE_1), \mcD_2 = (\mV \cup \mL_2, \mE_2)$ and a set of intervention targets $\mcI \subseteq 2^V$, construct the $\mcI$-augmented MAGs following the steps in Definition~\ref{def: I aug mag} of $\mcD_1, \mcD_2$. $\mcD_1$ and $\mcD_2$ are $\mcI$-Markov equivalent with respect to $\mcI$ if and only if for each $\mI \in \mcI$, $\Aum_\mI(\mcD_1, \mcI)$ and $\Aum_\mI(\mcD_2, \mcI)$ satisfy the 3 conditions in Theorem~\ref{thm: I-MEC}.
\end{proposition}

\section{Learning by Combining Experiments}
\label{sec: Learning algo}
In this section, we develop an algorithm to learn the causal structure from given datasets. Here we do not assume that the given datasets contain the observational data. Like any learning algorithm, a faithfulness assumption is necessary to infer graphical properties from the corresponding distributional constraints. Therefore, we assume that the provided interventional distributions are h-faithful to the causal graph $\mcD$, defined below.
\begin{definition}[h-faithful]
    \label{def: h-faithfulness}
    Consider a causal graph $\mcD = (\mV \cup \mL, \mE)$. A tuple of distributions $(P_\mI)_{\mI \in \mcI} \in \mcP(\mcD, \mV)$ is called h-faithful to graph $\mcD$ if the converse for each of the conditions given in Definition~\ref{def: I-Markov} holds.
\end{definition}

\subsection{Learning Objective}
 Similar to the case when only observational data is available, it is hard to recover the whole graph in general. Therefore, the objective of the algorithm is to learn a graphical representation that demonstrates a set of $\mcI$-Markov equivalent graphs. 
 However, although the MAG of the augmented pair graphs proposed in Definition~\ref{def: twin augmented mag} denotes the ground truth, it is not always a fundamentally learnable structure from the distributions. To witness, let us consider the example in Figure~\ref{fig: mag_1}. The edge $(F, Y^{(2)})$ is not in $\MAG(\Aug_{(\emptyset, \{Z\})}(\mcD_1))$. While we can only learn from the distributions that $P_{obs}(y|\mw) \neq P_Z(y|\mw), \mW \subseteq \{X, Z\}$. The inequality tells us that there is an inducing path from $F$ to $Y^{(1)}$ or $Y^{(2)}$ which we cannot distinguish. Therefore, we proposed the twin augmented MAG to be able to capture the characterization for $\mcI$-Markov equivalent graphs. Based on that, we construct the $\mcI$-augmented MAG which is a more compact graphical structure and we utilize as the learning objective. Accordingly, we define the $\mcI$-augmented graph as follows.
\begin{definition}[$\mcI$-augmented Graph]
    Given a causal graph $\mcD$ and a set of intervention targets $\mcI$, for each $\mI\in \mcI$, let $\mcM = \Aum_\mI(\mcD, \mcI)$ and let $[\mcM]$ be the set of $\mcI$-augmented MAGs corresponding to all the causal graphs that are $\mcI$-Markov equivalent to $\mcD$ given $\mcI$. For any $\mI \in \mcI$, the $\mcI$-augmented graph, denoted as $\mcG_\mI(\mcD, \mcI)$, is a graph such that:
    \begin{enumerate}
        \item $\mcG_\mI(\mcD, \mcI)$ has the same adjacencies as $\mcM$, and any member of $[\mcM]$ does; and
        \item every non-circle mark in $\mcG_\mI(\mcD, \mcI)$ is an invariant mark in $[\mcM]$.
    \end{enumerate}
    \label{def: I aug network}
The $\mcI$-augmented graph tuple $\mcL_{\mcI}(\mcD)$ is a tuple of all $\mcI$-augmented graphs $\mcL_{\mcI}(\mcD) = (\mcG_\mI (\mcD, \mcI) )_{\mI \in \mcI}$. We will omit the graph $\mcD$ or the intervention targets $\mcI$ in the brackets when it is clear from the context for simplicity.
\end{definition}

\subsection{The Learning Algorithm}
\begin{algorithm}[tb]
   \caption{Main Causal Discovery Algorithm}
   \label{alg: I-MEC learning}
\begin{algorithmic}
   \STATE {\bfseries Input:} Intervention targets $\mcI$, interventional distributions $(P_\mI)_{\mI \in \mcI}$, observable variables $\mV$\\
   Initialize $\mcL_\mcI$ as an empty tuple;\\
   \textbf{Phase I:} Initialize with Complete Graphs;\\
   \FOR{$\mI$ {\bfseries in} $\mcI$}
        \STATE Duplicate $\mV$ to create $\mV^{(\mathbf{I})}$;
        \STATE Put a circle edge $(\cleftrightarrow$) between every pair of nodes in $\mV^{(\mI)}$;

        \STATE Run Algorithm~\ref{alg: create F nodes} on $\mcI, \mI$ to get $\mcF_\mI$ as $F$ nodes;

        \FOR{$\mJ$  {\bfseries in} $\mcI\setminus\{ \mI \}$}
            \STATE Put a circle edge $(\cleftrightarrow)$ between every $V\in \mV^{(\mI)}$ and $F^{(\mI, \mJ)} \in \mcF_\mI$;
        \ENDFOR 
        
   \ENDFOR
   \textbf{Phase II:} Learning the Skeleton and Separating Sets;\\
   $SepSet \leftarrow \emptyset$;
   \FOR{$\mI$ {\bfseries in} $\mcI$}
    \STATE $E(\mcG_\mI) \leftarrow \emptyset$;
        \FOR{$\mJ$  {\bfseries in} $\mcI$}
            \STATE Run Algorithm~\ref{alg: find sep set} on $\mI, \mJ, (P_\mI)_{ \mI \in \mcI}, \mV, \mcF_\mI$ to get $\mcE_\mJ, SepSet_\mJ$;
            \STATE $E(\mcG_\mI) \leftarrow E(\mcG_\mI) \cup \mcE_\mJ$;
            \STATE $SepSet \leftarrow SepSet \cup SepSet_\mJ$;
        \ENDFOR 
        \STATE $V(\mcG_\mI) \leftarrow \mV^{(\mathbf{\mI})} \cup \mcF_\mI$;
        \STATE Add $\mcG_\mI$ to $\mcL_\mcI$;
   \ENDFOR
   
   \textbf{Phase III:} Apply Orientation Rules to each $\mcG_\mI, \mI \in \mcI$;\\
    \textbf{Rule 0:} For every unshielded triple $\langle X, Y, Z\rangle$ in $\mcG_\mI, \mI \in \mcI$, orient it as $X \srightarrow Y \sleftarrow Z$ if $Y  \notin SepSet(X, Z)$. \\
   
   First apply Rule 0, then apply 7 FCI rules in \citet{zhang2008completeness} together with the following 4 additional rules to each $\mcG_\mcI$ until none applies.\\

   \textbf{Rule 8:} For any edge adjacent to an $F$ node, orient the edge out of the $F$ node. 

   \textbf{Rule 9:} For any $\mI \in \mcI$, if $X\in \mI$, $X^{(\mathbf{I})}, Y^{(\mathbf{I})}$ are adjacent in $\mcG_\mI$, then orient $X^{(\mathbf{I})}\circlestar Y^{(\mathbf{I})}$ as $X^{(\mathbf{I})}\rightarrow Y^{(\mathbf{I})}$. 
   
   \textbf{Rule 10:} If $X^{(\mathbf{I})}\rightarrow Y^{(\mathbf{I})}$ in $\mcG_\mI$ for some $\mI \in \mcI$, replace the circle mark at $Y^{(\mathbf{J})}$ between $X^{(\mathbf{J})}$ and $Y^{(\mathbf{J})}$ in $\mcG_\mJ$ with an arrowhead for any $\mJ \in \mcI \setminus \{ \mI \}$.

   \textbf{Rule 11:} In $\mcG_\mI$, $\mI, \mJ \in \mcI$, if $\mJ = \mI \cup \{X\}$, $F^{(\mI, \mJ)}$ is adjacent to $Y^{(\mathbf{I})}, Y\notin \mJ$, then orient $X^{(\mathbf{I})} \starstar Y^{(\mathbf{I})}$ as $X^{(\mathbf{I})}\rightarrow Y^{(\mathbf{I})}$. 

   \STATE{\bfseries Output:}  $\mcI$-augmented graph tuple $\mcL_\mcI$
\end{algorithmic}
\end{algorithm}

\begin{algorithm}[tb]
   \caption{Algorithm for Creating $F$ Nodes}
   \label{alg: create F nodes}
\begin{algorithmic}
   \STATE {\bfseries Input:} Intervention set $\mcI$, intervention $\mI$\\
   Initialize $\mathcal{F} \leftarrow \emptyset$;
   \FOR{$\mJ$ {\bfseries in} $\mcI \setminus \{ \mI \}$}
        \STATE $\mathcal{F} \leftarrow \mathcal{F} \cup \{F^{(\mI, \mJ)}\}$
   \ENDFOR
   \STATE{\bfseries Output:}  The set of $F$ nodes $\mathcal{F}$
\end{algorithmic}
\end{algorithm}
We propose Algorithm~\ref{alg: I-MEC learning} to learn the $\mcI$-augmented graph tuple from the given experiments. The algorithm is inspired by the FCI algorithm. To explain the algorithm, we first outline the FCI procedure, which operates in three phases given an independence model over the measured variables~\cite{zhang2006causal}. In phase I, it initializes a complete graph with only circle edges (\cleftrightarrow), then removes edges between pairs of nodes if a separating set between them exists, and records all the separating sets. In Phase II, it identifies unshielded triples $\langle A, B, C\rangle$ and orients the edges into $B$ if $B$ is not in the separating set of $A$ and $C$. In phase III, it applies a set of orientation rules to refine the graph. The orientation rules serve for the MAG properties, found separating sets, and the soundness and completeness of the previous phases, i.e., all correct edges and unshielded colliders have been directed. Notice that if two nodes $X, Y$ are separated given $\mZ$ in the $\mcI$-augmented MAG, they are also separated given $\mZ \cup \mcF$ since $F$ nodes are introduced as source nodes and thus conditioning on them will not activate any extra path.

\begin{algorithm}[ht]
   \caption{Algorithm for Finding Separating Set}
   \label{alg: find sep set}
\begin{algorithmic}
   \STATE {\bfseries Input:} Target $\mI$ and $\mJ$, interventional distributions $(P_\mI)_{\mI \in \mcI}$,  observable variables $\mV$, $F$ nodes $\mcF$\\
   Initialize $\mcE \leftarrow \emptyset$;
   \IF{$\mI = \mJ$}
        \FOR{$X, Y$ \textbf{in} $\mV$}
            \STATE $SepFlag \leftarrow False$;
            \FOR{$\mW \subset \mV$}
                \IF{$P_\mI(y|\mw, x) = P_\mI(y|\mw)$} 
                    \STATE $SepFlag \leftarrow True$
                    \STATE $SepSet(X^{(\mI)}, Y^{(\mI)}) = \mW^{(\mI)} \cup \mcF \cup \mI^{(\mI)}$;
                \ENDIF
            \ENDFOR
            \IF{$SepFlag = False$}
                \STATE $\mcE \leftarrow \mcE \cup (X^{(\mI)}, Y^{(\mI)})$;
            \ENDIF
        \ENDFOR
   \ELSE
   \FOR{$Y$  {\bfseries in} $\mV$}
        \STATE $SepFlag \leftarrow False$;
        \IF{$Y$  {\bfseries in} $\mI \Delta \mJ$}
            \STATE Pass;
        \ELSE
        \FOR{$\mW \subseteq \mV$}
            \IF{$P_\mI(y|\mw) = P_\mJ(y|\mw)$}
                \STATE $SepFlag \leftarrow True$;
                \STATE $SepSet(F, Y^{(\mI)}) \leftarrow \mW^{(\mI)} \cup \mcF \cup \mI^{(\mI)}$;
                \STATE $SepSet(F, Y^{(\mJ)} ) \leftarrow \mW^{(\mJ)} \cup \mcF \cup \mJ^{(\mJ)}$;
            \ENDIF
        \ENDFOR
        \ENDIF
         \IF{$SepFlag = False$}
            \STATE $\mcE \leftarrow \mcE \cup (F^{(\mI, \mJ)}, Y^{(\mI)}) $;
        \ENDIF
   \ENDFOR
   \ENDIF
   \STATE{\bfseries Output:}  Circle edges $\mcE$, separating sets $SepSet$
\end{algorithmic}
\end{algorithm}

Algorithm~\ref{alg: I-MEC learning} follows a similar framework to that of FCI. It learns the $\mcI$-augmented graph tuple $\mcL_\mcI$ by iteratively recovering $\mcG_\mI$ for each $\mI \in \mcI$. In Phase I, it initializes the $\mcI$-augmented graph $\mcG_\mI$ for each $\mI \in \mcI$. It puts a circle edge between each pair of nodes $X, Y \in \mV$. This constructs the domain specific skeleton $G^{(\mI)}$ under target $\mI$. After that, we attach the $F$ nodes to the $\mcI$-augmented graph using Algorithm~\ref{alg: create F nodes} and then put a circle edge between any $F$ node and any $X \in \mV$.
In Phase II, we learn the skeleton for each $\mcI$-augmented graph. $\mcG_\mI$. For each $\mI \in \mcI$, we retrieve the $\mcI$-augmented graph $\mcG_\mI$ with all circle edges.  Algorithm~\ref{alg: find sep set} tests if there is a separating set between any pair of nodes $X, Y$ in $\mcG_\mI$. If both nodes are non-$F$ nodes, this can be tested through checking if $P_\mI(y|\mw, x)$ and $P_\mI(y|\mw)$ are equal. If one of them is an $F$ node, this can be tested by checking the equality between $P_\mI(y|\mw)$ and $P_\mJ(y|\mw)$ for $Y\in \mV, \mW \subseteq \mV$. Two $F$ nodes are separated by default. If there is no such set $\mW$, we preserve the circle edge between $X$ and $Y$. Otherwise, we remove the circle edge from $\mcG_\mI$.
In Phase III, we apply orientation rules to learn more edges in the constructed $\mcI$-augmented graphs. Notice that the skeleton $G^{(\mathbf{I})}$ in $\mcG_\mI$ is a PAG; thus, the FCI rules are still applicable here. We first use Rule 0 to orient all the unshielded triples by checking if the node in the center of the triple is in the separating set of the two end nodes. After that, Algorithm~\ref{alg: I-MEC learning} will repeatedly apply the FCI rules (Rules 5 to 7 are not included here as they are related to selection bias nodes) together with 4 new rules until none apply. 

\textbf{Rule 8 (F-node edges):} For any edge adjacent to an $F$ node, orient the edge out of the $F$ node. 

\textbf{Rule 9 (Intervened nodes):} For any $\mI \in \mcI$, if $X\in \mI$, $X^{(\mathbf{I})}, Y^{(\mathbf{I})}$ are adjacent in $\mcG_\mI$, then orient $X^{(\mathbf{I})}\circlestar Y^{(\mathbf{I})}$ as $X^{(\mathbf{I})}\rightarrow Y^{(\mathbf{I})}$. The intuition of this rule is that since $X$ is intervened, all the non-descendants of $X$ become separable from $X$ in $\mcD_{\overline{\mI}}$. Thus, $Y$ has to be a descendant of $X$ in $\mcD_{\overline{\mI}}$.

\textbf{Rule 10 (Consistency of Skeletons):} If $X^{(\mathbf{I})}\rightarrow Y^{(\mathbf{I})}$ in $\mcG_\mI$ for some $\mI \in \mcI$, replace the circle mark at $Y^{(\mathbf{J})}$ between $X^{(\mathbf{J})}$ and $Y^{(\mathbf{J})}$ in $\mcG_\mJ$ with an arrowhead for any $\mJ \in \mcI \setminus \{ \mI \}$. The intuition of this rule stems from the fact that each skeleton is obtained from the same causal graph $\mcD$. The ancestral relationship between any pair of nodes cannot be reversed by hard interventions.

\textbf{Rule 11 (Inducing Path):} In $\mcG_\mI$, $\mI, \mJ \in \mcI$, if $\mJ = \mI \cup \{X\}$, $F^{(\mI, \mJ)}$ is adjacent to $Y^{(\mathbf{I})}, Y\notin \mJ$, then orient $X^{(\mathbf{I})} \starstar Y^{(\mathbf{I})}$ as $X^{(\mathbf{I})}\rightarrow Y^{(\mathbf{I})}$. 
The intuition is that the $F$ nodes cannot be separated from $Y\notin \mK$, meaning there is an inducing path to $Y^{(\mathbf{I})}$ or $Y^{(\mathbf{J})}$ through $X^{(\mathbf{I})}$ or $X^{(\mathbf{J})}$. If $X$ is intervened, the inducing path cannot go through $X^{(\mJ)}$. 

To illustrate how each step in Algorithm~\ref{alg: I-MEC learning} works, we show an example in Appendix~\ref{app: ex learning}. We establish the soundness of the proposed algorithm.

\begin{theorem}
\label{thm: soundness of algo}
    Consider a set of interventional distributions $(P_\mI)_{\mI \in \mcI}$ that are h-faithful to a causal graph $\mcD = (\mV \cup \mL, \mE)$, where $\mcI$ is a set of intervention targets. Algorithm~\ref{alg: I-MEC learning} is sound, i.e., every adjacency and arrowhead/tail orientation in the returned $\mcI$-augmented graph $\mcG_\mI(\mcD, \mcI)$ for each $\mI \in \mcI$ is common for all $\mcI$-augmented MAGs of $\mcD^{\prime}$, $\mcG_\mI(\mcD^\prime, \mcI)$ for any $\mcD^{\prime}$ which is $\mcI$-Markov equivalent to $\mcD$.
\end{theorem}

\section{Experiments}
\label{sec: Exp}
\subsection{Enumerate All ADMGs}
In this experiment, we aim to demonstrate that the $\mcI$-MEC size under hard interventions is smaller than that under soft interventions in general. Here is how we set up the experiments. Given the number of observable variables $n$, we iterate through all possible ADMGs structures with $n$ nodes. The number of such ADMGs can be found as follows. We first generate all possible DAGs (not necessarily connected) with $n$ nodes. Then, for each pair of nodes in a DAG, there can be a bidirected edge or not. Thus, the number of ADMGs under consideration is $2^{\binom{n}{2}}$ multiplied by the number of all DAGs. Each time, we randomly pick an ADMG $G$ as the ground truth graph. Assuming that the size of the intervention targets is 2, we construct the twin augmented MAG $\mcM_1$ of $G$ following Definition~\ref{def: twin augmented mag} and the Augmented MAG $\mcM_2$ of $G$ following Definition 4 in~\cite{kocaoglu2019characterization}. We then enumerate all possible ADMGs, and for each candidate ADMG $G^\prime$, we construct its corresponding $\mcM_1^\prime$ and $\mcM_2^\prime$, and compare $\mcM_1, \mcM_1^\prime$ following Theorem~\ref{thm: I-MEC}, $\mcM_2, \mcM_2^\prime$ following Theorem 2 in~\cite{kocaoglu2019characterization}. For each $n$, we repeat the experiment 30 times and calculate the average size of $\mcI$-MEC size together with standard errors. We consider two types of sampling ADMGs, random and complete. 'Random' means the ADMG is constructed by randomly adding directed and bidirected edges with a probability of 0.5 between each pair of nodes while not creating any cycle. 'Complete' means we first construct a random complete DAG and then randomly add bidirected edges with a probability of 0.5 between each pair. For intervention targets, we assume they are either atomic or an empty set. We only show the results for small $n$, as the number of ADMGs grows super-exponentially with $n$. The numerical results are shown in Table~\ref{table: enumerate admg}. We notice that the number of ADMGs grows very fast with $n$. The results have a high standard error. This is due to the differences between the sampled graphs. Nevertheless, under hard interventions, the $\mcI$-MEC size tends to shrink the $\mcI$-MEC size more efficiently than with soft interventions on average, showing the power of hard interventions. Furthermore, as $n$ grows, the ratio of hard $\mcI$-MEC divided by the soft $\mcI$-MEC decreases, meaning that hard interventions extract more information on the causal graphs.

\begin{table}[h]
    \centering
    \begin{tabular}{l|c|c|c|c|r}  
        \hline
        $n$ & Mean under Hard & Mean under Soft & Graph & Ratio & Total Number of ADMGs \\
        \hline
        2   & $2.03 \pm 0.15$   & $2.93 \pm 0.29$ & Random & $0.69 \pm 0.05$  & 6 \\
        2   & $2.37 \pm 0.12$   & $3.67 \pm 0.22$ & Complete & $0.65 \pm 0.05$  & 6 \\
        3   & $19.50 \pm 3.41$   & $30.57 \pm 4.36$ & Random & $0.64 \pm 0.11  $& 200 \\
        3   & $14.03 \pm 2.69$   & $24.70 \pm 4.12$ & Complete & $0.57 \pm 0.05$  & 200 \\
        4  & $677.13 \pm 227.72$  & $ 1218.83 \pm 361.83$ & Random & $0.56 \pm 0.18$  & 34,752 \\
        4  & $721.37 \pm 276.36$  & $1529.57 \pm 368.68$ & Complete & $0.47 \pm 0.07$  & 34,752 \\
        \hline
    \end{tabular}
    \caption{Estimation of $\mcI$-MEC size by enumerating all ADMGs of the same size. We consider random ADMGs and ADMGs by adding bidirected edges to  random complete DAGs. For each setting we sample 30 ground truth ADMGs and calculate the mean and standard error of $\mcI$-MEC size and the ratio.}
    \label{table: enumerate admg}
\end{table}

\subsection{Sample ADMGs}
When $n = 5$, the total number of valid ADMGs is 29,983,744, which is intractable to enumerate. Instead, we can sample ADMGs to estimate the expectation of the probability of a randomly sampled ADMG to be $\mcI$-Markov equivalent to a given ADMG using Hoeffding's Inequality. To do this, given $n$, we randomly sample a DAG that is a complete graph using the uniform CPDAG sampler by~\cite{wienobst2021polynomial}. Then, for each sampled complete DAG, we add bidirected edges uniformly to each pair of nodes to construct a ground truth ADMG. To compare with the ADMG, we randomly pick two intervention targets that are either an empty set or atomic, and then randomly sample ADMGs following the same process and construct the augmented graphs. Suppose for each ground truth ADMG, we draw $M$ such random samples. For the $i$-th sample, $S_i=1$, if it is $\mcI$-Markov equivalent to the true graph and $S_i = 0$ otherwise. We define $\mathcal{S} = \sum_{i = 1}^M S_i$ which shows the number of $\mcI$-Markov equivalent ADMGs. We denote $\mathbb{E}_\mathcal{S}$ as the expectation we are approximating. Thus, according to Hoeffding's Inequality, we have:
\begin{equation}
        P(\left| \frac{\mathcal{S}} {M} - \mathbb{E}_\mathcal{S}\right| \geq \epsilon)\leq \exp(-2M\epsilon^2)
    \end{equation}
We choose $\epsilon = 0.01$ and $\exp(-2M\epsilon^2) = 0.01$ for $M$ with $M = 23025$. For each setting, we randomly sample 50 ground truth ADMGs and then take the average. The results are shown in Table~\ref{table: bound admg}, Table~\ref{table: bound admg fix latents}, and Table~\ref{table: bound admg fix n, change rho}. We can see that the estimated $\mathbb{E}^{hard}_{\mathcal{S}}$ is significantly lower than $\mathbb{E}^{soft}_{\mathcal{S}}$ meaning hard interventions can more efficiently learn the causal structure. Notice that as $n$ becomes larger, the number of ADMGs grows fast and thus the expectations get close to $0$. Consequently, a much smaller $\epsilon$ would be necessary to approximate the expectations, leading to a much larger number of samples $M$.

\begin{table}[h]
    \centering
    \begin{tabular}{l|c|c|c}  
        \hline
        $n$ & Estimated $\mathbb{E}^{hard}_{\mathcal{S}}$ & Estimated $\mathbb{E}^{soft}_{\mathcal{S}}$ & Ratio  \\
        \hline
        2   & $0.417 \pm 0.010$   & $0.584 \pm 0.010$ & $0.715 \pm 0.011$  \\
        3   & $0.143 \pm 0.012$   & $0.235 \pm 0.016$ & $0.607 \pm 0.022$ \\
        4   & $0.058 \pm 0.011$   & $0.112\pm 0.015$ & $0.514 \pm 0.024$  \\
        5  & $0.028 \pm 0.011 $  & $0.061 \pm 0.013 $ & $0.459 \pm 0.028$ \\
        6  & $0.015 \pm 0.010$  & $0.036 \pm 0.012$ &  $0.420 \pm 0.030$ \\
        
        \hline
    \end{tabular}
    \caption{Estimation of $\mcI$-MEC size on complete DAGs with different sizes and $0.5$ density of bidirected edges.}
    \label{table: bound admg}
\end{table}

\begin{table}[h]
    \centering
    \begin{tabular}{l|c|c|c}  
        \hline
        $n$ & Estimated $\mathbb{E}^{hard}_{\mathcal{S}}$ & Estimated $\mathbb{E}^{soft}_{\mathcal{S}}$ & Ratio  \\
        \hline
        3   & $0.151 \pm 0.011$   & $0.264 \pm 0.015$ & $0.571 \pm 0.017$  \\
        4   & $0.067 \pm 0.010$   & $0.149 \pm 0.013$ & $0.451 \pm 0.017$  \\
        5  & $0.034 \pm 0.010$  & $0.091 \pm 0.013$ &  $0.373 \pm 0.017$ \\
        6  & $0.019 \pm 0.010$  & $0.059 \pm 0.012$ &  $0.317 \pm 0.017$ \\
        
        \hline
    \end{tabular}
    \caption{Estimation of $\mcI$-MEC size with different $n$ and fixed $0.45n(n-1)$ bidirected edges (approximately 0.9 in density).}
    \label{table: bound admg fix latents}
\end{table}
\begin{table}[h]
    \centering
    \begin{tabular}{l|c|c|c}  
        \hline
        $\rho$ & Estimated $\mathbb{E}^{hard}_{\mathcal{S}}$ & Estimated $\mathbb{E}^{soft}_{\mathcal{S}}$ & Ratio  \\
        \hline
        0.1   & $0.020 \pm 0.011$   & $0.024 \pm 0.011$ & $0.804 \pm 0.018$  \\
        0.3   & $0.022 \pm 0.011$   & $0.040 \pm 0.012$ & $0.566 \pm 0.030$  \\
        0.5   & $0.028 \pm 0.011 $  & $0.061 \pm 0.013 $ & $0.459 \pm 0.028$  \\
        0.7  & $0.031 \pm 0.011$  & $0.081 \pm 0.013$ &  $0.395 \pm 0.021$ \\
        0.9  & $0.034 \pm 0.010$  & $0.093 \pm 0.011$ &  $0.365 \pm 0.015$ \\
        \hline
    \end{tabular}
    \caption{Estimation of $\mcI$-MEC size on completed DAGs with $5$ nodes and different densities of bidirected edges. It shows that when the density of bidirected edges goes up, hard interventions shrink the $\mcI$-MEC size more efficiently than soft interventions.}
    \label{table: bound admg fix n, change rho}
\end{table}
\section{Conclusion}
\label{sec: conclusion}
We address the challenge of learning the causal structure underlying a phenomenon of interest using a combination of several experimental data. Specifically, different from \cite{kocaoglu2019characterization}, we study the problem under hard interventions. The motivation comes from the observation that hard interventions are more informative than soft ones, and thus we can learn more about the original causal graph. Our approach builds on a generalization of the converse of Pearl’s do-calculus (Theorem~\ref{thm: do-calculus}), which introduces new tests that can be applied to data. These tests translate into structural constraints. We define an interventional equivalence class based on these criteria (Definition~\ref{def: I-Markov}) and provide a graphical characterization for the equivalence of two causal graphs (Theorem~\ref{thm: I-MEC}) using the proposed twin augmented MAG structure (Definition~\ref{def: twin augmented mag}). To construct a unified graphical representation that is closer to the ground truth ADMG, we combine the twin augmented MAGs into a $\mcI$-augmented MAG (Definition~\ref{def: I aug mag}) and show the equivalence of the two representations (Proposition~\ref{prop: I aug mag I markov equivalent}. Finally, we propose an algorithm (Algorithm~\ref{alg: I-MEC learning}) to learn the interventional equivalence class represented by the $\mcI$-augmented graphs (Definition~\ref{def: I aug network}) from data, incorporating novel orientation rules. We also prove the soundness of the proposed algorithm (Theorem~\ref{thm: soundness of algo}). The empirical results in Section~\ref{sec: Exp} show that hard interventions efficiently decrease the size of $\mcI$-Markov Equivalence Class.

\section*{Acknowledgement}
\label{sec: broader impact}
This research has been supported in part by NSF CAREER 2239375, IIS 2348717, Amazon Research Award, Adobe Research and Intuit.

\bibliography{references}
\bibliographystyle{icml2024}

\newpage
\appendix
\onecolumn
\addtocontents{toc}{\protect\setcounter{tocdepth}{3}}
\renewcommand{\contentsname}{Appendix Contents}

\tableofcontents

\newpage

\section{Detailed Related Works}
\textbf{Equivalence Class:} The learning process uses data constraints to infer the causal diagram. However, these constraints often cannot uniquely identify the complete diagram. As a result, the analysis typically focuses on an equivalence class (EC) of causal diagrams that captures the constraints implied by the underlying causal system. When only observational data is available, such EC is called MEC. It characterizes causal graphs with the same set of d-separation statements over observable variables~\citep{spirtes2001causation, verma1992algorithm, spirtes2013causal, meek2013causal}. Under causal sufficiency, 
~\citet{hauser2012characterization, yang2018characterizing} characterizes the $\mcI$-MEC. ~\citet{tian2001causal} first considers the equivalence class under local changes. When there are latents under soft intervention with known targets, \citet{kocaoglu2019characterization} characterize the $\mcI$-MEC while with unknown targets, it is called $\psi$-MEC~\citep{jaber2020causal}. When there is access to multiple domains, ~\citet{li2023causal} propose S-MEC.

\textbf{Learning from Combined Datasets:}
There are plenty of works in the literature on learning the causal structure from experiments (or across domains). However, most of them provide empirical approaches without theoretical analysis. Approaches like~\citet{perry2022causal, peters2016causal, ghassami2017learning, heinze2018invariant, huang2020causal} often assume Markovianity or specific functional models (e.g., linearity), or combinations of observational and interventional data with both known and unknown targets. In contrast, JCI combines all data into a single dataset and performs learning on the pooled data~\citep{mooij2020joint}. \citet{ke2019learning, brouillard2020differentiable} introduces neural-network-based frameworks that leverages observational and interventional data to identify causal structures, even when intervention targets are unknown, demonstrating superior performance in structure recovery tasks but there is no soundness analysis. \citet{acharya2018learning} demonstrates that with \(O(\log n)\) interventions and \(O(n/\epsilon^2)\) samples per intervention, one can effectively distinguish whether an unknown CBN matches a given model or differs by more than \(\epsilon\) in total variation distance. \citet{jiang2023learning} offers the first results to characterize conditions under which causal representations are identifiable without parametric assumptions, even in settings with unknown interventions and without assuming faithfulness. \citet{addanki2021intervention} aims to determine the directions of all causal or ancestral relations in $G$ using a minimum-cost set of interventions. GIES~\citep{hauser2012characterization} and IGSP~\citep{wang2017permutation} are score-based and aim at learning a single DAG under known targets; there is no soundness or equivalence class analysis either. \citet{lopez2022large} developed a differentiable causal discovery for large and high-dimensional data. \citet{tigas2022interventions} proposed an experiment design algorithm to adaptively choose the intervention targets. \citet{zhou2024sample, mascaro2023bayesian} proposes Bayesian causal discovery algorithms under causal sufficiency. 

\section{Proofs}
\label{app: proofs}

\subsection{Proof for Proposition~\ref{prop: do-constraint}}
In this section, we extend the do-calculus rules to enable their application across two arbitrary interventions. This extension is crucial for characterizing our equivalence class when arbitrary sets of interventional distributions are provided.

\textit{Proof.} 
\textbf{Rule 1:} The results follow from rule 1 of Theorem~\ref{thm: do-calculus}.

\textbf{Rule 2:} From $(\mY \indep \mK_\mJ|\mW, \mI)_{\mcD_{\overline{\mI}, \underline{\mK_\mJ}}}$, we can derive from rule 2 of Theorem~\ref{thm: do-calculus} that $P_\mI(\my| \mw, \mz) = P_{\mI, \mJ}( \my| \mw, \mz)$. Similarly, given $(\mY \indep \mK_\mI | \mW, \mJ )_{\mcD_{\overline{\mJ}, \underline{\mK_\mI}}}$, we have $P_\mJ( \my| \mw, \mz) = P_{\mI, \mJ}(\my | \mw, \mz)$. Together, we have $P_\mI( \my | \mw, \mz) = P_{\mJ}(\my | \mw, \mz)$. Here, we use $P_{\mI, \mJ}$ as the intermediate distribution to show the equality and use rule 2 in Theorem~\ref{thm: do-calculus} twice. This is necessary, as for any $\mS \subset \mI \cup \mJ$, $\mcD_{\overline{\mS}}$ would be a denser graph that contains fewer m-separation statements than $\mcD_{\overline{\mI, \mJ}}$. Next, we show that it is also sufficient. 
\begin{lemma}
    \label{lm: sufficiency rule 2}
    If $\exists \mS \subset \mW \setminus ( \mI \cup \mJ)$, s.t. $(\mY \indep \mK_J, \mS|\mW, \mI)_{\mcD_{\overline{\mI}, \underline{\mK_\mJ, \mS}}} \land 
    (\mY \indep \mK_\mI, \mS | \mW, \mJ)_{ \mcD_{\overline{\mJ}, \underline{\mK_\mI, \mS}}}$, then $(\mY \indep \mK_\mJ | \mW, \mI)_{\mcD_{\overline{\mI}, \underline{\mK_\mJ}}} \land (\mY \indep \mK_\mI|\mW, \mJ )_{\mcD_{\overline{\mJ}, \underline{\mK_\mI}}}$.  
\end{lemma}
\textit{Proof.} Suppose otherwise $(\mY \dep \mK_J|\mW, \mI)_{ \mcD_{\overline{\mI}, \underline{\mK_\mJ}}}$, then there is an m-connecting path $p$ from $Y \in \mY$ to $U\in \mK_\mJ$ in $\mcD_{\overline{\mI}, \underline{\mK_\mJ}}$. Given $(\mY \indep \mK_\mJ, \mS|\mW, \mI )_{\mcD_{\overline{\mI}, \underline{\mK_\mJ, \mS}}}$, we have $(\mY \indep \mK_J|\mW, \mI)_{\mcD_{\overline{\mI}, \underline{\mK_\mJ, \mS}}}$. Comparing $(\mY \dep \mK_J|\mW, \mI)_{\mcD_{\overline{\mI}, \underline{\mK_\mJ}}}$ and $(\mY \indep \mK_\mJ|\mW, \mI)_{ \mcD_{\overline{\mI}, \underline{\mK_\mJ, \mS}}}$, the only difference between the two statements is the edges outgoing from $\mS$. Removing the edges creates $p$. Consider the case that $S \in \mS$ is in $p$. For $p$ to be m-connecting, $S$ can only be a collider. However, such a m-connecting path indicates that there is also a m-connecting path from $Y$ to $S$ which contradicts the given conditions. Thus, $\mS$ cannot be in $p$. Then, there has to be a collider in $p$ activated by some $S\in \mS$ in $\mcD_{\overline{\mI}, \underline{\mK_\mJ}}$. Consider a collider $C$ that is closest to $Y$ in $p$. $C$ is an ancestor of $S$. Then the path created by concatenating the subpath from $Y$ to $C$ and $C$ to $S$ is an m-connecting path which is also in $\mcD_{\overline{\mI}, \underline{\mK_\mJ, \mS}}$. This is a contradiction. Following the same process we can also show that $(\mY \indep \mK_\mI | \mW, \mJ)_{ \mcD_{\overline{\mJ}, \underline{\mK_\mI}}}$. \qed

Lemma~\ref{lm: sufficiency rule 2} shows that the graphical conditions for $P_\mI( \my | \mw, \mz) = P_{\mI, \mJ}(\my | \mw, \mz) = P_\mJ( \my| \mw, \mz)$ are sufficient to use another interventional distribution $P_{\mI, \mJ, \mS}$ as the intermediate distribution to show the equality.

\textbf{Rule 3:} From $(\mY \indep \mK_\mJ| \mW, \mI)_{ \mcD_{\overline{\mI, \mK_\mJ(\mW)}}}$, we can tell using rule 3 of Theorem~\ref{thm: do-calculus} that $P_\mI(\my|\mw) = P_{\mI, \mJ}(\my|\mw) $. From $(\mY \indep \mK_\mI | \mW, \mJ)_{ \mcD_{\overline{\mJ, \mK_\mI(\mW)}}}$, we have $P_{\mI, \mJ}(\my|\mw) = P_\mJ(\my|\mw)$. Together, we have $P_\mI(\my|\mw) = P_\mJ(\my|\mw)$. Similarly, we apply the do-do rule twice and use $P_{\mI, \mJ}$ as an intermediate distribution to show the equality. Likewise, any distribution that corresponds to a denser graph indicates the conditions. Next, we show the sufficiency of the conditions.
\begin{lemma}
    \label{lm: sufficiency rule 3}
    If $\exists \mS \subset \mV\setminus (\mI \cup \mJ \cup \mW)$, such that $(\mY \indep \mK_\mJ, \mS | \mW, \mI )_{\mcD_{\overline{\mI, \mK_\mJ(\mW), \mS(\mW)}}} \land (\mY \indep \mK_\mI, \mS|\mW, \mJ)_{\mcD_{\overline{\mJ, \mK_\mI(\mW), \mS(\mW)}}}$, then $(\mY \indep \mK_\mJ | \mW, \mI )_{\mcD_{\overline{\mI, \mK_\mJ(\mW)}}} \land (\mY \indep \mK_\mI | \mW, \mJ )_{\mcD_{\overline{J, \mK_\mI(\mW)}}}$.
\end{lemma}
\textit{Proof.} Suppose otherwise $(\mY \dep \mK_\mJ | \mW, \mI )_{\mcD_{\overline{\mI, \mK_\mJ(\mW)}}}$, then there is an m-connecting path $p$ from $Y \in \mY$ to $U \in \mK_\mJ(\mW)$ in $\mcD_{\overline{\mI, \mK_\mJ(\mW)}}$. Given $(\mY \indep \mK_\mJ, \mS | \mW, \mI )_{\mcD_{\overline{\mI, \mK_\mJ(\mW), \mS(\mW)}}}$, we have $(\mY \indep \mK_\mJ | \mW, \mI )_{\mcD_{\overline{\mI, \mK_\mJ(\mW), \mS(\mW)}}}$. Comparing $(\mY \indep \mK_\mJ | \mW, \mI )_{\mcD_{\overline{\mI, \mK_\mJ(\mW), \mS(\mW)}}}$ and $(\mY \dep \mK_\mJ|\mW, \mI)_{\mcD_{\overline{\mI, \mK_\mJ(\mW)}}}$, the only difference in the graph is the edges into $\mS(\mW)$. Consider the case that there is some $S \in \mS(\mW)$ in $p$. Consider $S$ that is closest to $Y$. $S$ cannot be a collider on $p$ since it is not an ancestor of any $W\in \mW$, and thus $p$ will be blocked. Let us consider the 2 cases:

If $S$ has an outgoing edge towards $Y$ on $p$, then there is a directed path from $S$ to $Y$ because otherwise there has to be a collider $C\in \mW$ in between $Y$ and $S$, and $S$ will be an ancestor of $C$. However, the subpath from $Y$ to $S$ in $p$ would then be m-connecting in $\mcD_{\overline{\mI, \mK_\mJ(\mW), \mS(\mW)}}$, which is a contradiction.

If $S$ has an incoming edge from $Y$ on $p$, then there is a directed path $S$ to $U$ on $p$. Otherwise, there has to be a collider $C$ in the subpath between $S$ and $U$ on $p$. Such $C$ that is closest to $S$ makes $S$ an ancestor of $C\in \mW$, a contradiction. Nevertheless, if there is a directed path from $S$ to $U$, $S$ is an ancestor of $U$. If $U$ is an ancestor of any $W\in \mW$, $S$ will also be an ancestor of $W$, a contradiction. Thus $U$ cannot be an ancestor of any $W\in \mW$, and the edges into $U$ will be removed in $\mcD_{\overline{\mI, \mK_\mJ(\mW)}}$ which contradicts the assumption that $p$ is an m-connecting path.

Therefore, $S$ cannot be on $p$. While $S$ has to be an ancestor of some $W \in \mW$ to activate $p$ if it is not on $p$, this contradicts the assumption that $S\in \mS(\mW)$. To conclude, we show that $(\mY \indep \mK_\mJ | \mW, \mI )_{\mcD_{\overline{\mI, \mK_\mJ(\mW)}}}$. Following the same process, we can also show that $(\mY \indep \mK_\mI | \mW, \mJ )_{\mcD_{\overline{\mJ, \mK_\mI(\mW)}}}$.
\qed

Lemma~\ref{lm: sufficiency rule 3} shows that the graphical conditions for $P_\mI(\my | \mw) = P_{\mI, \mJ}(\my | \mw) = P_\mJ(\my | \mw)$ are sufficient to use another interventional distribution $P_{\mI, \mJ, \mS}$ as the intermediate distribution to show the equality by using rule 3 of Theorem~\ref{thm: do-calculus} twice.

\textbf{Rule 4:} We begin by introducing a useful lemma. 
\begin{lemma}
    \label{lm: inter rule 4}
    For disjoint $\mX, \mY, \mW \subseteq \mV, \mS \subseteq \mV \setminus (\mX \cup \mY), \mS \cap \mW \neq \emptyset, \mS \neq \mW$, then $P_\mx(\my | \mw ) = P_{\mx, \ms}(\my | \mw)$ if: $ (\mS_\mR \indep \mY |\mW, \mX)_{\mcD_{\overline{\mX, \mS_\mR(\mW)}}} \land (\mS_\mW \indep \mY | \mW \setminus \mS_\mW, \mX)_{\mcD_{\overline{\mX}, \underline{\mS_\mW} }}$ where $\mS_\mW = \mS \cap \mW, \mS_\mR = \mS \setminus \mW$.
\end{lemma}
\textit{Proof.} Denote the statement $(\mS_\mR \indep \mY |\mW, \mX)_{\mcD_{\overline{\mX, \mS_\mR(\mW)}}} \land (\mS_\mW \indep \mY | \mW \setminus \mS_\mW, \mX)_{\mcD_{\overline{\mX}, \underline{\mS_\mW} }}$ as $\mcC_1$. We first show that, by applying rule 3 and rule 2 in Theorem~\ref{thm: do-calculus} sequentially, we can transform from $P_\mx$ to $P_{\mx, \ms}$. Specifically, $P_\mx(\my | \mw) = P_{\mx, \ms_\mR}(\my | \mw)$ if $(\mS_\mR \indep \mY |\mW, \mX)_{\mcD_{\overline{\mX, \mS_\mR(\mW)}}}$, and 
$P_{\mx, \ms_\mR}(\my | \mw) = P_{\mx, \ms_\mR, \ms_\mW}(\my | \mw)$ if $(\mS_\mW \indep \mY | \mW \setminus \mS_\mW, \mX, \mS_\mR)_{\mcD_{\overline{\mX, \mS_\mR}, \underline{\mS_\mW} }}$. By definition, $P_{\mx, \ms_\mR, \ms_\mW}(\my | \mw) = P_{\mx, \ms}(\my | \mw)$. 

Denote the statement $ (\mS_\mR \indep \mY |\mW, \mX)_{\mcD_{\overline{\mX, \mS_\mR(\mW)}}} \land (\mS_\mW \indep \mY | \mW \setminus \mS_\mW, \mX, \mS_\mR)_{\mcD_{\overline{\mX, \mS_\mR}, \underline{\mS_\mW} }}$ as $\mcC_2$. Next, we show that $\mcC_1$ and $\mcC_2$ are equivalent. Notice that they share the same statement, and we just need to show that the other one holds true.

($\mcC_1 \Rightarrow \mcC_2$): Suppose otherwise, $(\mS_\mW \dep \mY | \mW \setminus \mS_\mW, \mX, \mS_\mR)_{\mcD_{\overline{\mX, \mS_\mR}, \underline{\mS_\mW} }}$, then there is an m-connecting path $p$ from $Y \in \mY$ to $U \in \mS_\mW$ in $\mcD_{\overline{\mX, \mS_\mR}, \underline{\mS_\mW} }$. Comparing $(\mS_\mW \dep \mY | \mW \setminus \mS_\mW, \mX, \mS_\mR )_{\mcD_{\overline{\mX, \mS_\mR}, \underline{\mS_\mW} }}$ and $(\mS_\mW \indep \mY | \mW \setminus \mS_\mW, \mX)_{\mcD_{\overline{\mX}, \underline{\mS_\mW} }}$, the difference is the $\mS_\mR$ in the conditioning set and the edges into $\mS_\mR$. Consider the case that there is some $S \in \mS_\mR$ in $p$. Since the edges into $S$ are removed, $S$ can only have outgoing edges, but conditioning on $S$ will then block $p$. Thus, $S$ cannot be in $p$. If $S$ is not in $p$, since the edges into $S$ are removed, conditioning on $S$ will not activate any path. Thus the supposition cannot hold.

($\mcC_2 \Rightarrow \mcC_1$): Suppose otherwise, $(\mS_\mW \dep \mY | \mW \setminus \mS_\mW, \mX)_{\mcD_{\overline{\mX}, \underline{\mS_\mW} }}$. Then there is an m-connecting path $p$ from $Y \in \mY$ to $U \in \mS_\mW$ in $\mcD_{\overline{\mX}, \underline{\mS_\mW} }$. Comparing $(\mS_\mW \indep \mY | \mW \setminus \mS_\mW, \mX, \mS_\mR )_{\mcD_{\overline{\mX, \mS_\mR}, \underline{\mS_\mW} }}$ and $(\mS_\mW \dep \mY | \mW \setminus \mS_\mW, \mX)_{\mcD_{\overline{\mX}, \underline{\mS_\mW} }}$, the difference is the $\mS_\mR$ in the conditioning set and the edges into $\mS_\mR$. Consider the case that there is some $S\in \mS_\mR$ that is closest to $Y$ in $p$. 

If $S$ is a colliser on $p$, then $S$ is an ancestor of some $W \in \mW$. There is a m-connecting path from $S$ to $Y$ in $\mcD_{\overline{\mX, \mS_\mR(\mW)}}$. A contradiction.

If $S$ has an outgoing edge towards $Y$ in $p$, there has to be a collider $C \notin \mW$ between $S$ and $Y$ in $p$. Otherwise, the subpath between $S$ and $Y$ will be m-connecting in $\mcD_{\overline{\mX, \mS_\mR(\mW)}}$ which contradicts the supposition. Since $p$ is m-connecting, $C$ has to be an ancestor of some $W \in \mW$ in $\mcD_{\overline{\mX}, \underline{\mS_\mW}}$. To block $p$ in $\mcD_{\overline{\mX, \mS_\mR}, \underline{\mS_\mW}}$, there has to be some $S^\prime \in \mS$ that is in between $C$ and $W$. This will create an m-connecting path from $S^\prime$ to $Y$ in $\mcD_{\overline{\mX, \mS_\mR(\mW)}}$, which is a contradiction.

If $S$ has an outgoing edge towards $S_W$ in $p$, then $S$ has to be an ancestor of some $W\in \mW$ in $\mcD_{\overline{\mX, \mS_\mR(\mW)}}$, which is a contradiction.

Therefore, $S$ is not in $p$. There has to be a collider $C \notin \mW$ in $p$. Consider such $C$ closest to $Y$. There is a directed path from $C$ to $S$ and a directed path from $S$ to some $W\in \mW$ in $\mcD_{\overline{\mX}, \underline{\mS_\mW}}$ for $p$ to be m-connecting. However, this makes $S$ an ancestor of $W$ in $\mcD_{\overline{\mX, \mS_\mR(\mW)}}$, and the path created by concatenating the directed path from $C$ to $S$ and the subpath from $C$ to $Y$ in $p$ would be d-connecting in $\mcD_{\overline{\mX, \mS_\mR(\mW)}}$. A contradiction. The supposition does not hold.

This concludes the proof of this lemma. \qed

By applying Lemma~\ref{lm: inter rule 4} twice, we can derive the graphical condition for $P_\mI( \my | \mw) =P_{\mI, \mJ}(\my | \mw) = P_\mJ(\my | \mw)$ for two arbitrary interventions $\mI, \mJ$. 
The following lemma shows that this is a sufficient condition.
\begin{lemma}
    \label{lemma: mix do see sufficiency}
    If $\exists \mS \subset \mV\setminus (\mI \cup \mJ), \mS_\mR = \mS \setminus \mW, \mS_\mW = \mS \cap \mW$, such that $(\mY \indep \mR_\mJ, \mS_\mR | \mW, \mI )_{ \mcD_{\overline{\mI, \mK_\mJ(\mW), \mS_\mR(\mW)}}} \land ( \mY \indep \mW_J, \mS_\mW | \mW \setminus (\mS_\mW \cup \mW_\mJ), \mI)_{\mcD_{\overline{\mI} \underline{ \mW_\mJ, \mS_\mW}}}$, then $(\mY \indep \mR_\mJ | \mW, \mI )_{ \mcD_{\overline{\mI, \mR_\mJ(\mW)}}} \land (\mY \indep \mW_\mJ | \mW \setminus \mW_\mJ, \mI )_{\mcD_{\overline{\mI}, \underline{\mW_\mJ}}}$.
\end{lemma}
\textbf{Proof.} 
We first consider $(\mY \indep \mR_\mJ | \mW, \mI )_{ \mcD_{\overline{\mI, \mR_\mJ(\mW)}}}$. Suppose otherwise $(\mY \dep \mR_\mJ | \mW, \mI )_{ \mcD_{\overline{\mI, \mR_\mJ(\mW)}}}$, then there is an m-connecting path $p$ from $Y \in \mY$ to $U\in \mR_\mJ$ in $\mcD_{\overline{\mI, \mR_\mJ(\mW)}}$. From the given condition, we know that $(\mY \indep \mR_\mJ | \mW, \mI )_{ \mcD_{\overline{\mI, \mK_\mJ(\mW), \mS_\mR(\mW)}}}$. The only difference is the edges into $\mS_\mR(\mW)$ in $\mcD_{\overline{\mI}}$. If there is any $S\in \mS_\mR$ in $p$, then it has only outgoing edges. Since $p$ is m-connecting, the subpath from $S$ to $Y$ on $p$ is also m-connecting in $\mcD_{\overline{\mI, \mR_\mJ(\mW), \mS_\mR(\mW)}}$, which is a contradiction. Thus $\mS_\mR$ cannot be on $p$. To block $p$ in $\mcD_{\overline{\mI, \mR_\mJ(\mW), \mS_\mR(\mW)}}$, there is a collider $C$ in $p$ that is an ancestor of some $W\in \mW$. $C$ is activated in $p$ in $\mcD_{\overline{\mI, \mR_\mJ(\mW)}}$ but not activated in $\mcD_{\overline{\mI, \mR_\mJ(\mW), \mS_\mR(\mW)}}$. This requires some $S\in \mS_\mR(\mW)$ to be an ancestor of $W$, which is impossible. Therefore, the supposition does not hold.

Next, we consider $(\mY \indep \mW_\mJ | \mW \setminus \mW_\mJ, \mI )_{\mcD_{\overline{\mI}, \underline{\mW_\mJ}}}$. Suppose otherwise, $(\mY \dep \mW_\mJ | \mW \setminus \mW_J, \mI )_{\mcD_{\overline{\mI}, \underline{\mW_\mJ}}}$, then there is an m-connecting path $p$ from $Y\in \mY$ to $U\in \mW_\mJ$ in $\mcD_{\overline{\mI}, \underline{\mW_\mJ}}$. $p$ is blocked in $\mcD_{\overline{\mI}, \underline{\mW_\mJ, \mS_\mW}}$. Comparing $(\mY \indep \mW_\mJ | \mW \setminus \mW_\mJ, \mI )_{\mcD_{\overline{\mI}, \underline{\mW_\mJ}}}$ and $( \mY \indep \mW_\mJ | \mW \setminus (\mS_\mW \cup \mW_\mJ), \mI )_{ \mcD_{\overline{\mI} \underline{ \mW_\mJ, \mS_\mW}}}$, the only difference is the edges outgoing from $\mS_\mW$. If there is some $S \in \mS_\mW$ in $p$, then $S$ has to be a collider. Consider such $S$ closest to $Y$. Since $p$ is m-connecting, then the subpath from $S$ to $Y$ is also m-connecting in $ \mcD_{\overline{\mI} \underline{ \mW_J, \mS_W}}$, which is a contradiction. Thus, $S$ cannot be in $p$. For $S$ to block $p$ in $ \mcD_{\overline{\mI} \underline{ \mW_\mJ, \mS_\mW}}$ while not in $p$, it has to be an ancestor of some $W\in \mW$ and a descendant of some collider in $p$. However, since $S\in \mW$, it can still activate the collider in $p$ in $ \mcD_{\overline{\mI} \underline{ \mW_\mJ, \mS_\mW}}$. Therefore, the supposition does not hold.
\qed

By applying Lemma~\ref{lemma: mix do see sufficiency} twice, we can show that it is sufficient to transfer through $P_{\mI, \mJ}$. This concludes the proof of this theorem. \qed

\subsection{Proof for Proposition~\ref{prop: F node}}
We show the graphical conditions on the augmented pair graphs are equivalent to those given in the generalized causal calculus rules.
\begin{proposition}
    \label{prop: generalized F-node}
    Consider a CBN ($\mcD = (\mV \cup \mL, \mE), P$) with latent variables $\mL$ and its augmented pair graph $\Aug_{(\mI, \mJ)}(\mcD) = (\mV^{(\mI)}\cup \mV^{(\mJ)}\cup \{F\}, \mE^{(\mI)}\cup \mE^{(\mJ)} \cup \mcE)$ with respect to a pair of interventions $\mI, \mJ \in \mcI$. Let $\mS = \mI \Delta \mJ$ be a set of nodes, $F$ is adjacent to $\mS^{(\mI)}, \mS^{(\mJ)}$. We have the following equivalence relations:
    
    Suppose disjoint $\mY, \mZ, \mW \subseteq \mV$. We have
    \begin{equation}
        \label{eq: F-node r1}
        (\mY \indep \mZ |\mW, \mI )_{\mcD_{\overline{\mI}}} \iff 
        (\mY^{(\mI)} \indep \mZ^{(\mI)} |\mW^{(\mI)}, \mI^{(\mI)}, F)_{ \Aug_{(\mI, \mJ)}(\mcD)}
    \end{equation}

    Suppose $\mY, \mW$ are disjoint subsets of $\mV \setminus \mS$. We have
    \begin{equation}
        \label{eq: F-node r2}
        \begin{split}
        \left
        \{
        \begin{array}{ll}
          (\mY \indep \mW_\mJ | \mW \setminus \mW_\mJ, \mI )_{\mcD_{\overline{\mI}, \underline{\mW_\mJ}}}\\
          (\mY \indep \mW_\mI | \mW \setminus \mW_\mI, \mJ )_{\mcD_{\overline{\mJ}, \underline{\mW_\mI}}}
        \end{array}
        \right. 
        \iff 
        \left
            \{
        \begin{array}{cc}
            (F \indep \mY^{(\mI)}|\mW^{(\mI)}, \mI^{(\mI)})_{\Aug_{(\mI, \mJ)}(\mcD)} \\
            (F \indep \mY^{(\mJ)}|\mW^{(\mJ)}, \mJ^{(\mJ)})_{\Aug_{(\mI, \mJ)}(\mcD)}
        \end{array}
        \right.
        \end{split}
    \end{equation}

    \begin{equation}
        \label{eq: F-node r3}
        \begin{split}
        \left
        \{
        \begin{array}{ll}
          (\mY \indep \mR_\mJ|\mW, \mI)_{\mcD_{\overline{\mI, \mR_\mJ(\mW)}}}\\
          (\mY \indep \mR_\mI | \mW, \mJ )_{\mcD_{\overline{\mJ, \mR_\mI(\mW)}}} 
        \end{array}
        \right. 
        \iff 
        \left
            \{
        \begin{array}{cc}
            (F \indep \mY^{(\mI)}|\mI^{(\mI)}, \mW^{(\mI)})_{\Aug_{(\mI, \mJ)}(\mcD)} \\
            (F \indep \mY^{(\mJ)}|\mJ^{(\mJ)}, \mW^{(\mJ)})_{\Aug_{(\mI, \mJ)}(\mcD)}
        \end{array}
        \right.
        \end{split}
    \end{equation}

    For disjoint $\mY, \mZ, \mW \subseteq \mV$, where $\mK_\mI = \mK \setminus \mJ, \mK_\mJ = \mK \setminus \mI, \mW_\mI = \mK_\mI \cap \mW, \mW_\mJ = \mK_\mJ \cap \mW, \mR = \mK \setminus \mW, \mR_\mI = \mR \cap \mK_\mI, \mR_\mJ = \mR \cap \mK_\mJ$
    \begin{equation}
        \label{eq: F-node r4}
        \begin{split}
        \left
        \{
        \begin{array}{ll}
          (\mY \indep \mR_\mJ | \mW, \mI )_{\mcD_{\overline{\mI, \mR_\mJ(\mW)}}}\\
          (\mY \indep \mW_\mJ | \mW \setminus \mW_\mJ, \mI )_{\mcD_{\overline{\mI}, \underline{\mW_\mJ}}}\\
          (\mY \indep \mR_\mI|\mW, \mJ )_{\mcD_{\overline{J, \mR_\mI(\mW)}}} \\
          (\mY \indep \mW_\mI | \mW \setminus \mW_\mI, \mJ)_{\mcD_{\overline{\mJ}, \underline{\mW_\mI}}}
        \end{array}
        \right. 
        \iff 
        \left
            \{
        \begin{array}{cc}
            (F \indep \mY^{(\mI)}|\mI^{(\mI)}, \mW^{(\mI)})_{\Aug_{(\mI, \mJ)}(\mcD)} \\
            (F \indep \mY^{(\mJ)}|\mJ^{(\mJ)}, \mW^{(\mJ)})_{\Aug_{(\mI, \mJ)}(\mcD)}
        \end{array}
        \right.
        \end{split}
    \end{equation}
    
\end{proposition}
\textit{Proof.} Consider Equation~\ref{eq: F-node r1}. For the right statement, conditioning on $F$ is equivalent to removing it and the subgraph induced by $\mV^{(\mJ)}$ from $Aug(\mcD)$. Then the statements on the two sides are equivalent.

Consider Equation~\ref{eq: F-node r2}. We need to show that $(\mY \indep \mW_\mJ | \mW \setminus \mW_\mJ, \mI)_{\mcD_{\overline{\mI}, \underline{\mW_\mJ}}} \Leftrightarrow (F \indep \mY^{(\mI)} | \mW^{(\mI)}, \mI^{(\mI)})_{\Aug_{(\mI, \mJ)}(\mcD)}$.

\textbf{($\Rightarrow$)} Suppose otherwise $(F \dep \mY^{(\mI)}|\mW^{(\mI)}, \mI^{(\mI)})_{\Aug_{(\mI, \mJ)}(\mcD)}$, then there is an m-connecting path $p$ from $F$ to $Y^{(\mI)} \in \mY^{(\mI)}$ in $\Aug_{(\mI, \mJ)}(\mcD)$. Since $\mW^{(\mI)}$ is conditioned on, $p$ cannot be a frontdoor path. Also, edges into $\mI^{(\mI)}$ are removed, thus $p$ cannot be a backdoor path through $\mW_\mI$. However, given $(\mY\indep \mW_\mJ|\mW \setminus \mW_\mJ, \mI)_{\mcD_{\overline{\mI}, \underline{\mW_\mJ}}}$, there is no backdoor path through $\mW_\mJ$. The supposition does not hold.

\textbf{($\Leftarrow$)} Suppose otherwise $(\mY \dep \mW_\mJ | \mW \setminus \mW_\mJ, \mI )_{\mcD_{\overline{\mI}, \underline{\mW_\mJ}}}$, then there is an m-connecting path $p$ from $Y^{(\mI)} \in \mY^{(\mI)}$ to $U \in \mW_\mJ$. It has to be a backdoor path from $U$. However, $(F \indep \mY^{(\mI)} | \mW^{(\mI)} , \mI^{(\mI)})_{\Aug_{(\mI, \mJ)}(\mcD)}$ will not hold, if there is a backdoor path from $U$ to $Y$. Thus the supposition does not hold.

For the same reason, we can show that $(\mY \indep \mW_I | \mW \setminus \mW_\mI, \mJ )_{\mcD_{\overline{\mJ}, \underline{\mW_\mI}}} \Leftrightarrow (F \indep \mY^{(\mJ)}|\mW^{(\mJ)}, \mJ^{(\mJ)})_{\Aug_{(\mI, \mJ)}(\mcD)}$.

Consider Equation~\ref{eq: F-node r3}. We need to show that $(\mY \indep \mR_\mJ | \mW, \mI)_{\mcD_{\overline{ \mI, \mR_\mJ(\mW)}}} \Leftrightarrow (F \indep \mY^{(\mI)}|\mI^{(\mI)}, \mW^{(\mI)})_{\Aug_{(\mI, \mJ)}(\mcD)}$.

\textbf{($\Rightarrow$)} Suppose otherwise $(F \dep \mY^{(\mI)}| \mI^{(\mI)}, \mW^{(\mI)} )_{\Aug_{(\mI, \mJ)}(\mcD)}$, then there is an m-connecting path $p$ from $F$ to $Y^{(\mI)} \in \mY^{(\mI)}$ in $\Aug_{(\mI, \mJ)}(\mcD)$. Since $\mI^{(\mI)}$ is conditioned on and the edges into $\mI^{(\mI)}$ are removed, $p$ cannot be through $\mI^{(\mI)}$. If $F$ has a frontdoor path to $Y^{(\mI)}$ through $\mR_\mJ^{(\mI)}$, then $Y$ to $\mR_\mJ$ will also be m-connecting in $\mcD_{\overline{\mI, \mR_\mJ(\mW)}}$. Else if it is a backdoor path, it can only go through some $U^{(\mI)} \in \mR_\mJ^{(\mI)}(\mW)$; it contradicts $(\mY \indep \mR_\mJ | \mW, \mI )_{\mcD_{\overline{\mI, \mR_\mJ(\mW)}}}$. Thus the supposition does not hold.

\textbf{($\Leftarrow$)} Suppose otherwise $(\mY \dep \mR_\mJ|\mW, \mI )_{\mcD_{\overline{\mI, \mR_\mJ(\mW)}}}$, then there is an m-connecting path $p$ from $Y \in \mY$ to $U \in \mR_\mJ$ in $\mcD_{\overline{\mI, \mR_\mJ(\mW)}}$. If $p$ has an outgoing from $U$, then $F$ would have a frontdoor path through $U^{(\mI)}$ to $Y^{(\mI)}$ that is m-connecting. This contradicts $(F \indep \mY^{(\mI)}| \mI^{(\mI)}, \mW^{(\mI)})_{\Aug_{(\mI, \mJ)}(\mcD)}$. Thus $p$ can only be a backdoor path from $U$. If $U$ is not an ancestor of any $W \in \mW$ in $\mcD_{\overline{\mI}}$, the edges into $U$ are removed and $p$ do not exist in this case. Else if $U$ is an ancestor of some $W \in \mW$, the same backdoor path would also be activated in $\Aug(\mcD)$. Therefore, the supposition does not hold.

For the same reason, we can show that $(\mY \indep \mR_\mI|\mW, \mJ)_{ \mcD_{\overline{\mJ, \mR_\mI(\mW)}}} \Leftrightarrow (F \indep \mY^{(\mJ)}|\mJ^{(\mJ)}, \mW^{(\mJ)})_{\Aug_{(\mI, \mJ)}(\mcD)}$.

Consider Equation~\ref{eq: F-node r4}. We need to show that $(\mY \indep \mR_\mJ | \mW, \mI )_{ \mcD_{\overline{\mI, \mR_\mJ(\mW)}}} \land (\mY \indep \mW_\mJ | \mW \setminus \mW_\mJ, \mI )_{\mcD_{\overline{\mI}, \underline{\mW_\mJ}}} \Leftrightarrow (F \indep \mY^{(\mI)}|\mI^{(\mI)}, \mW^{(\mI)})_{\Aug_{(\mI, \mJ)}(\mcD)}$.

\textbf{($\Rightarrow$)} Suppose otherwise $(F \dep \mY^{(\mI)}| \mI^{(\mI)}, \mW^{(\mI)})_{\Aug_{(\mI, \mJ)}(\mcD)}$, then there is an m-connecting path $p$ from $F$ to $Y^{(\mI)}\in \mY^{(\mI)}$, in $\Aug_{(\mI, \mJ)}(\mcD)$. If $p$ is a frontdoor path through $\mS^{(\mI)}$, since $\mI^{(\mI)}, \mW^{(\mI)}$ are conditioned on, only $\mR_\mJ^{(\mI)}$ could be in $p$. However, the path from $\mR_\mJ$ to $Y$ would be m-connecting in $\mcD_{\overline{\mI, \mR_\mJ(\mW)}}$ which is a contradiction. If $p$ is a backdoor path from $F$ to $U \in \mS^{(\mI)}$ to $Y^{(\mI)} \in \mY^{(\mI)}$, since the edges into $\mI^{(\mI)}$ are removed, $U$ can only be from $\mR_J^{(\mI)}$ or $\mW_J^{(\mI)}$. If $U \in \mW_\mJ^{(\mI)}$, then it contradicts $(\mY\indep \mW_\mJ | \mW \setminus \mW_\mJ, \mI )_{ \mcD_{\overline{\mI}, \underline{\mW_\mJ}}}$. If $U \in \mR_\mJ^{(\mI)}$, $U$ could be either an ancestor of some $W^{(\mI)} \in \mW^{(\mI)}$ or a non-ancestor of any $W^{(\mI)} \in \mW^{(\mI)}$.  For the case that $U$ is not an ancestor of any $W^{(\mI)} \in \mW^{(\mI)}$, $p$ is blocked by $U$ as an inactivated collider. If $U$ is an ancestor of some $W^{(\mI)} \in \mW^{(\mI)}$, then the subpath from $U$ to $\mY^{(\mI)}$ would indicate an m-connecting path from $\mR_\mJ$ to $\mY$ in $\mcD_{\overline{\mI}, \underline{\mW_\mJ}}$ which contradicts $(\mY \indep \mR_\mJ | \mW, \mI )_{\mcD_{\overline{\mI, \mR_\mJ(\mW)}}}$. Thus, the supposition does not hold.

\textbf{($\Leftarrow$)} Suppose otherwise, there are two cases to consider, either $(\mY \dep \mR_\mJ | \mW, \mI )_{\mcD_{\overline{\mI, \mR_\mJ(\mW)}}}$ or $(\mY \dep \mW_\mJ | \mW \setminus \mW_\mJ, \mI )_{\mcD_{\overline{\mI}, \underline{\mW_\mJ}}}$. 

If $(\mY\dep \mW_\mJ | \mW \setminus \mW_\mJ, \mI )_{\mcD_{\overline{\mI}, \underline{\mW_\mJ}}}$, there is an m-connecting path $p$ from $Y\in \mY$ to $U\in \mW_\mJ$ in $\mcD_{\overline{\mI}, \underline{\mW_\mJ}}$. $p$ cannot be a frontdoor path at $U$, since all edges outgoing from $\mW_\mJ$ are removed. While a valid backdoor path at $U$ indicates that there is also a valid path from $F$ to $Y^{(\mI)}$ through $U^{(\mI)}$ in $\Aug_{(\mI, \mJ)}(\mcD)$. Thus, this case is impossible.

If $(\mY \dep \mW_J | \mW \setminus \mW_J, \mI )_{\mcD_{\overline{\mI}, \underline{\mW_\mJ}}}$, there is an m-connecting path $p$ from $Y \in \mY$ to $U \in \mR_\mJ$ in $\mcD_{\overline{\mI, \mR_\mJ(\mW)}}$. $p$ cannot be a frontdoor path at $U$, because otherwise the path constructed by adding $F\rightarrow U^{(\mI)}$ to $p$ in $\Aug_{(\mI, \mJ)}(\mcD)$ will be m-connecting. Thus, $p$ can only be a backdoor path at $U$. If $U$ is not an ancestor of any $W \in \mW$ in $\mcD_{\overline{\mI}}$, the edges into $U$ are removed and such $p$ do not exist. Else if $U$ is an ancestor of some $W \in \mW$ in $\mcD_{\overline{\mI}}$, $W^{(\mI)}$ will activate the path constructed by adding $F \rightarrow U^{(\mI)}$ to $p$ in $\Aug_{(\mI, \mJ)}(\mcD)$, which contradicts $(F \indep \mY^{(\mI)} | \mI^{(\mI)}, \mW^{(\mI)})_{\Aug_{(\mI, \mJ)}(\mcD)}$. Therefore, the supposition does not hold.

For the same reason, we can show that $ (\mY \indep \mR_\mI|\mW, \mJ )_{\mcD_{\overline{\mJ, \mR_\mI(\mW)}}} \land (\mY \indep \mW_\mI | \mW \setminus \mW_\mI, \mJ)_{\mcD_{\overline{\mJ}, \underline{\mW_\mI}}} \Leftrightarrow (F \indep \mY^{(\mJ)}|\mJ^{(\mJ)}, \mW^{(\mJ)})_{\Aug_{(\mI, \mJ)}(\mcD)}$.
\qed 

\textbf{Proof of Proposition~\ref{prop: F node}:} The follows from Proposition~\ref{prop: generalized F-node}. \qed

\subsection{Proof for Theorem~\ref{thm: I-MEC}}
\textit{Proof.} \textbf{(If)} Suppose that the twin augmented MAGs $\Twin_{(\mI, \mJ)}(\mcD_1), \Twin_{(\mI, \mJ)}(\mcD_2)$ for all $\mI, \mJ\in \mcI$ satisfy the 3 conditions. Then they induce the same m-separations and vice versa. Then by Proposition~\ref{prop: F node} that $\mcD_1$ and $\mcD_2$ impose the same constraints over the distribution tuples. Thus $\mcP_{\mcI}(\mcD_1, \mV) = \mcP_{\mcI}( \mcD_2, \mV)$.

\textbf{(Only if)} Suppose for a pair of interventions $\mI, \mJ \in \mcI$, $\mcM_1 = \Twin_{(\mI, \mJ)}(\mcD_1), \mcM_2 = \Twin_{(\mI, \mJ)}(\mcD_2)$ do not fully satisfy the 3 conditions. Then they must induce at least one different m-separation statement. We need to show that all the differences in m-separation statements induced by different $\mcM$ structures can be captured by some m-separation statements that are testable by the distribution tuples, and therefore, the difference in m-separation would be inducing different constraints on $\mcP_{\mcI}(\mcD_1, \mV)$ and $\mcP_{\mcI}(\mcD_2, \mV)$. Thus the condition that $\mcP_{\mcI}(\mcD_1, \mV) = \mcP_{\mcI}(\mcD_2, \mV)$ will no longer hold, which is a contradiction.

We start by showing all testable m-separation statements. For an arbitrary twin augmented MAG $\mcM = (\mV^{(\mI)}\cup \mV^{(\mJ)} \cup \{F\}, \mE^{(\mI)}\cup \mE^{(\mJ)}\cup \mcE)$, the testable m-separation statements are as follows:
\begin{align*}
    \mathcal{T} = 
    &\{( \mX^{(\mI)}\indep \mY^{(\mI)}| \mZ^{(\mI)}, \mI^{(\mI)}, F )_{\mcM}:\mX, \mY \subseteq \mV, \mZ \subseteq \mV\setminus (\mX \cup \mY)\} \cup \\
    &\{( \mX^{(\mJ)}\indep \mY^{(\mJ)}| \mZ^{(\mJ)}, \mJ^{(\mJ)}, F )_{\mcM}:\mX, \mY \subseteq \mV, \mZ \subseteq \mV\setminus (\mX \cup \mY)\} \cup \\
    &\{(F\indep \mY^{(\mI)} | \mI^{(\mI)}, \mZ^{(\mI)})_{\mcM} \land  (F\indep \mY^{(\mJ)} | \mJ^{(\mJ)}, \mZ^{(\mJ)})_{\mcM}: \mY \subseteq \mV, \mZ\subseteq \mV \setminus \mY \}    
\end{align*}
Next, we show that $\mcM_1$ and $\mcM_2$ should have the same skeleton. 

First, we show that they have the same skeleton on $\mV^{(\mI)}$. Suppose otherwise, in $\mcM_1$, $X^{(\mI)}$ and $Y^{(\mI)}$ are adjacent but they are non-adjacent in $\mcM_2$. Then it implies that $(X^{(\mI)} \dep Y^{(\mI)}|\mZ^{(\mI)}, F)_{\mcM_1} \land (X^{(\mI)} \indep Y^{(\mI)}|\mZ^{(\mI)}, F)_{\mcM_2}$ for some $\mZ \subseteq \mV$. Then we can further condition on $I^{(\mI)}$ while preserving the m-separations since edges into $\mI^{(\mI)}$ are removed and thus conditioning on it will not activate any extra path. Therefore, we have $(X^{(\mI)} \dep Y^{(\mI)}|\mZ^{(\mI)}, \mI^{(\mI)}, F)_{\mcM_1} \land (X^{(\mI)} \indep Y^{(\mI)}|\mZ^{(\mI)}, \mI^{(\mI)}, F )_{\mcM_2}$ which is a pair of different testable statements in $\mathcal{T}$. Similarly, $\mcM_1$ and $\mcM_2$ also have the same skeleton in $\mV^{(\mJ)}$. What remains to be demonstrated is that the \(F\) nodes share the same adjacencies in both graphs.

By the construction of twin augmented MAGs, $F$ is adjacent to $\mK^{(\mI)}, \mK^{(\mJ)}$ in both $\mcM_1$ and $\mcM_2$. We need to show that $F$ has the same adjacencies to $X\notin \mK$ in both graphs. Suppose otherwise, $F$ is adjacent to $X^{(\mI)}$ in $\mcM_1$, but non-adjacent to $X^{(\mI)}$ in $\mcM_2$. According to our construction of $\mcM_1, \mcM_2$, $F$ is adjacent to $X^{(\mI)}, X^{(\mJ)}$ in $\mcM_1$ but non-adjacent to them in $\mcM_2$. We introduce the following lemma which shows that in this case, we can still find a pair of different testable m-separation statements which reflect this structural difference.
\begin{lemma}
    \label{lm: F node separation}
    Consider a causal graph $\mcD = (\mV \cup \mL, \mE)$ given a set of intervention targets $\mcI \subseteq 2^V$. Construct its twin augmented MAG $\mcM = (\mV^{(\mI)}\cup \mV^{(\mJ)}\cup \{F\}, \mE^{(\mI)}\cup \mE^{(\mJ)}\cup \mcE)$, for $\mI, \mJ \in \mcI$. If there exists minimal $\mW_1, \mW_2 \subseteq \mV$, such that $(F\indep X^{(\mI)} | \mI^{(\mI)}, \mW_1^{(\mI)})_{\mcM} \land (F\indep X^{(\mJ)} | \mJ^{(\mJ)}, \mW_2^{(\mJ)})_{\mcM}$, then $(F\indep X^{(\mI)}|\mI^{(\mI)}, \mW^{(\mI)})_{\mcM} \land (F\indep X^{(\mJ)} | \mJ^{(\mJ)}, \mW^{(\mJ)})_{\mcM}$, where $\mW = \mW_1 \cup \mW_2$.
\end{lemma}
\textit{Proof.} Suppose otherwise $(F\dep X^{(\mI)} | \mI^{(\mI)}, \mW^{(\mI)})_{\mcM}$, meaning that conditioning on $\mW^{(\mI)}$ activates extra paths from $F$ to $X^{(\mI)}$ in $\mcM$. Obviously, $\mI^{(\mI)}$ can neither be on the paths nor activate a collider on the paths. Consider a m-connecting path $p^{(\mI)}$ from $F$ to $X^{(\mI)}$ given $\mW^{(\mI)}$ in $\mcM$. Then there is a collider $C_1^{(\mI)}$ on $p^{(\mI)}$ which is activated by some $W_2^{(\mI)} \in \mW_2^{(\mI)}$. Let the subgraph of $\mcM$ induced by $V^{(\mJ)}\cup F$ be $G_J$. $F$ and $X^{(\mJ)}$ are blocked by $\mW_2$ meaning that the corresponding path $p^{(\mJ)}$ in $G_J$ is either blocked by inactivated colliders or does not exist due to removed edges into $\mJ$. In both cases, $p^{(\mJ)}$ is blocked by $\emptyset$. However, due to the minimality of $\mW_2$, it has to block some path other than $p^{(\mJ)}$. Consider $W_x \in \mW_2$ which is closest to $X$ and is a descendant of a collider $C_1^{(\mJ)}$ on $p^{(\mJ)}$. Denote the path created by concatenating direct paths from $C_1^{(\mJ)}$ to $W_x^{\mJ)}$ and $C_1^{(\mJ)}$ to $X^{(\mJ)}$ as $p_x^{(\mJ)}$. Suppose there is a path $p_1^{(\mJ)}$ from $F$ to $X^{(\mJ)}$ that is blocked by $W_x^{(\mJ)}$. Since $p_1^{(\mJ)}$ is blocked by $W_x^{(\mJ)}$,  $W_x^{(\mJ)}$ cannot be a collider on $p_1^{(\mJ)}$. If there is an edge into $W_x^{(\mJ)}$ from $F$'s side, then the subpath of $p_x^{(\mJ)}$ will be m-connecting. The subpath from $F$ to $W_x^{(\mJ)}$ has to be blocked by inactivated colliders or conditioning on a non-collider. In either case, $W_x$ is not necessary for blocking $p_1^{(\mJ)}$. Conversely, there has to be an outgoing edge from $W_x^{(J)}$ towards $F$. Consequently, on $p_1^{(\mJ)}$, $F$ has an outgoing edge towards $W_x^{(\mJ)}$, while $W_x^{(\mJ)}$ has an outgoing edge towards $F$. Thus there has to be a collider in between $F$ and  $W_x^{(\mJ)}$. Consider such a collider $C_2^{(\mJ)}\in \mW_2^{(\mJ)}$ that is closest to $F$. Nevertheless, according to the minimality of $\mW_2$, $C_2^{(\mJ)}$ has to block another path. If we repeat this process for $n = |\mV|$ times, we show that $|\mW_2|>n$, which is impossible. Thus the supposition that $(F\dep X^{(\mI)} | \mI^{(\mI)}, \mW^{(\mI)})_{\mcM}$ does not hold. Similarly, we can show that $(F\indep X^{(\mJ)}|\mJ^{(\mJ)}, \mW^{(\mJ)})_{\mcM}$ which concludes the proof.
\qed

Therefore, according to Lemma~\ref{lm: F node separation}, the structural difference implies that we can find some $\mW \in \mW \setminus \{X\}$ such that 
$(F\dep X^{(\mI)} | \mI^{(\mI)}, \mW^{(\mI)})_{\mcM_1} \lor  (F \dep X^{(\mJ)} | \mJ^{(\mJ)}, \mW^{(\mJ)})_{\mcM_1}$ while $(F\indep X^{(\mI)} | \mI^{(\mI)}, \mW^{(\mI)})_{\mcM_2} \land  (F \indep X^{(\mJ)} | \mJ^{(\mJ)}, \mW^{(\mJ)})_{\mcM_2}$, which is a pair of different testable m-separation statements. To conclude, $\mcM_1, \mcM_2$ have the same skeleton.

Next, we show that $\mcM_1$ and $\mcM_2$ have the same unshielded colliders. We start by showing that they have the same unshielded colliders in the vertex induced subgraph on $\mV^{(\mI)}$ and $\mV^{(\mJ)}$. Suppose otherwise, $\langle X^{(\mI)}, Y^{(\mI)}, Z^{(\mI)} \rangle$ is an unshielded collider in $\mcM_1$ but not in $\mcM_2$. Since $Y^{(\mI)}$ is not a collider in $\mcM_2$, it has to be conditioned on to make $X^{(\mI)}$ and $Z^{(\mI)}$ m-separable. Then we have $(X^{(\mI)}\indep Z^{(\mI)} | \mI^{(\mI)}, \mW^{(\mI)})_{\mcM_1} \land  (X^{(\mI)} \dep Z^{(\mI)} | \mI^{(\mI)}, \mW^{(\mI)}, Y^{(\mI)})_{\mcM_1}$ and $(X^{(\mI)} \indep Z^{(\mI)} | \mI^{(\mI)}, \mW^{(\mI)})_{\mcM_2} \land  (X^{(\mI)} \indep Z^{(\mI)} | \mI^{(\mI)}, \mW^{(\mI)}, Y^{(\mI)})_{\mcM_2}$ for some $\mW \subseteq \mV\setminus \{Y\}$, which contains a pair of different testable statements.

Then we need to show that unshielded colliders which include $F$ nodes are also the same in both graphs. Due to the construction, $F$ can only have outgoing edges, thus it can only be an end node in the collider. Suppose $\langle F, Y^{(\mI)}, Z^{(\mI)}\rangle$ is an unshielded collider in $\mcM_!$ but not in $\mcM_2$. More specifically, $F\rightarrow Y^{(\mI)} \sleftarrow Z^{(\mI)}$ in $\mcM_1$ and $F \rightarrow Y^{(\mI)} \rightarrow Z^{(\mI)}$ in $\mcM_2$. There are 2 cases: $Y \in \mK$ or $Y \notin \mK$.

First, consider $Y \in \mK$.  $Y$ has to be in $\mJ \setminus \mI$, then in the induced subgraph of $\mV^{(\mJ)}$, we cannot have $Y^{(\mJ)} \sleftarrow Z^{(\mJ)}$ in $\mcM_1$ or 
$\mcM_2$. Thus we can find some $\mW \subseteq \mV \setminus \{Y\}$, such that $(F \dep Z^{(\mI)}|\mI^{(\mI)}, \mW^{(\mI)}, Y^{(\mI)} )_{\mcM_1} \lor (F \dep Z^{(\mJ)} | \mJ^{(\mJ)}, \mW^{(\mJ)}, Y^{(\mJ)} )_{\mcM_1}$, while $(F \indep Z^{(\mI)}|\mI^{(\mI)}, \mW^{(\mI)}, Y^{(\mI)} )_{\mcM_2} \land (F \dep Z^{(\mJ)} | \mJ^{(\mJ)}, \mW^{(\mJ)}, Y^{(\mJ)} )_{\mcM_2}$, which is a pair of different testable m-separation statements.

Second, consider the case that $Y\notin \mK$, but is adjacent to $F$ in both $\mcM_1$ and $\mcM_2$. Then there is an inducing path from $F$ to $Y^{(\mI)}$ or $Y^{(\mJ)}$ in the augmented pair graphs. According to our construction of twin augmented MAGs, $F$ is adjacent to both $Y^{(\mI)}$ and $Y^{(\mJ)}$ in both $\mcM_1$ and $\mcM_2$. Since $F$ is not adjacent to $Z^{(\mI)}, Z^{(\mJ)}$ in $\mcM_2$, we know according to Lemma~\ref{lm: F node separation} that there exists $\mW \subseteq \mV$ such that $(F \indep Z^{(\mI)} | \mI^{(\mI)}, \mW^{(\mI)})_{\mcM_2} \land (F \indep Z^{(\mJ)} | \mJ^{(\mJ)}, \mW^{(\mJ)})_{\mcM_2}$. $\mW$ has to contain $Y$, otherwise $F$ is m-connecting to $Z^{(\mI)}$ or $Z^{(\mJ)}$ in $\mcM_2$ through $Y^{(\mI)}$ or $Y^{(\mJ)}$. We need to further show that there cannot be $Y^{(\mJ)} \sleftarrow Z^{(\mJ)}$ in $\mcM_2$. Suppose otherwise, if $Y^{(\mJ)} \leftarrow Z^{(\mJ)}$ in $\mcM_2$, then there is a cycle in $\mcD_2$. Else if $Y^{(\mJ)} \leftrightarrow Z^{(\mJ)}$ in $\mcM_2$, then there is a latent common ancestor of $Y, Z$ in $\mcD_2$. This will create an inducing path $\langle F, Y^{(\mI)}, Z^{(\mI)}\rangle$ in $\mcM_2$ which contradicts the condition that $F$ and $Z^{(\mI)}$ are non-adjacent.
In $\mcM_1$, since there is $F \rightarrow Y \sleftarrow Z$, we have $(F\dep Z^{(\mI)} | \mI^{(\mI)}, \mW^{(\mI)} )_{\mcM_1} \land (F\dep X^{(\mJ)} | \mJ^{(\mJ)}, \mW^{(\mJ)} )_{\mcM_1}$, which contains a pair of different testable statements in $\mathcal{T}$. Therefore, $\mcM_1$ and $\mcM_2$ have the same unshielded colliders. 

Suppose $\langle U, W_1, W_2, ..., W_k, Y, Z\rangle$ is a discriminating path in both $\mcM_1$ and $\mcM_2$, $Y$ is a collider in $\mcM_1$ but a non-collider in $\mcM_2$. By our construction, $F$ can only be $U$ or $Y$ in the path. There are 3 cases.

First, if $F$ is not in the path. Then all the nodes in the path are in $\mV^{(\mI)}$ or $\mV^{(\mJ)}$. Suppose the path is in $\mV^{(\mI)}$. Then in $\mcM_2$, there exists some $\mW \subseteq \mV$, such that $(U^{(\mI)} \indep Z^{(\mI)}|\mI^{(\mI)}, F, \mW^{(\mI)} )_{\mcM_2}, W_i, Y \in \mW, i \in [k]$. While in $\mcM_1$, we have $(U^{(\mI)} \dep Z^{(\mI)}| \mI^{(\mI)}, F, \mW^{(\mI)})_{\mcM_1}$, since conditioning on $Y^{(\mI)}$ would activate the path. Thus there is a pair of testable m-separation statements.

Next, if $F$ is $Y$ in the path, this case is not valid, since $F$ can only have outgoing edges. Therefore, if $F$ is $Y$ in the path, it should have the same collider status in $\mcM_1$ and $\mcM_2$.

Lastly, $F$ is $U$ in the path. All of the other nodes in the path have to be all in either $\mV^{(\mI)}$ or $\mV^{(\mJ)}$. Suppose that all other nodes are in $\mV^{(\mI)}$. Since $F$ is non-adjacent to $Z^{(\mI)}$ in $\mcM_1$ and $\mcM_2$, we know that $F$ is also non-adjacent to $Z^{(\mJ)}$ according to Lemma~\ref{lm: F node separation}. Thus in $\mcM_2$, there exists $\mW \subseteq \mV$, such that $(F\indep Z^{(\mI)} | \mI^{(\mI)}, \mW^{(\mI)})_{\mcM_2} \land (F\indep Z^{(\mJ)} | \mJ^{(\mJ)}, \mW^{(\mJ)})_{\mcM_2}, W_i, Y \in \mW, i \in [k]$. $W_i, Y$ are in $\mW$ because otherwise $F$ is m-connecting with $Z^{(\mI)}$ through the path. However, in $\mcM_1$ we have $(F\indep Z^{(\mI)} | \mI^{(\mI)}, \mW^{(\mI)})_{\mcM_1} \land (F\indep Z^{(\mJ)} | \mJ^{(\mJ)}, \mW^{(\mJ)})_{\mcM_1}$ because conditioning on $W_i^{(\mI)}, Y^{(\mI)}$ will activate every collider in the path from $F$ to $Z^{(\mI)}$. Therefore, there is a pair of different testable statements in $\mathcal{T}$. To conclude, for each discriminating path in $\mcM_1$ and $\mcM_2$, $F$ is a collider in $\mcM_1$ if and only if it is a collider in $\mcM_2$. Up to now, we showed that if $\mcM_1$ and $\mcM_2$ do not satisfy the three graphical conditions in the theorem, then there exists a testable m-separation statement that holds in one graph but not the other.

 When there exists a pair of intervention targets \( \mI, \mJ \in \mathcal{I} \) such that 
\(
\mathcal{M}_1 = \text{Twin}_{(\mI, \mJ)}(\mathcal{D}_1) \quad \text{and} \quad 
\mathcal{M}_2 = \text{Twin}_{(\mI, \mJ)}(\mathcal{D}_2)
\) do not satisfy either of the three conditions mentioned in the theorem statement, this implies that \(
\mcD_1 \quad \text{and} \quad \mcD_2 \) are not $\mcI$-Markov equivalent. This is because there is a m-separation statement that appears as a condition in the definition of \(\mathcal{I}\)-Markov equivalence that is different in the two graphs $\mcM_1$ and $\mcM_2$. There is a m-separating path in $\mcM_1$ that is m-connecting in $\mcM_2$. We now show that $\mathcal{P}_{\mathcal{I}}(\mathcal{D}_2,\mathbf{V})$ contains tuples of distributions
that are not in $\mathcal{P}_{\mathcal{I}}(\mathcal{D}_1,\mathbf{V}) $. We leverage a key result from Meek, demonstrating that the set of unfaithful distributions has Lebesgue measure zero. Building on this, we construct a jointly Gaussian structural causal model that incorporates latent variables.

\begin{lemma} \cite{meek2013strong}
 Consider a causal DAG $D=(\mV, \mE)$, where $(A \not \indep B \mid C)_D$. Let $D_s=\left( \mV_s, \mE_s\right)$ be the subgraph that contains all the nodes in the m-connecting path that induces $(A \not \indep B \mid C)_D$. Then any distribution $P$ over $\mV_s$ where every adjacent pair of variables is dependent satisfies $(A \not \indep  B \mid C)_p$.
 \label{lemma_meek}
\end{lemma}

The proof of Lemma \ref{lemma_meek} uses weak transitivity and an inductive argument and can be found in \citet{meek2013strong}. Suppose that $\mX, \mY, \mZ \subseteq \mV$ such that $(\mX \indep \mY \mid \mZ, F)_{\operatorname{Aug}_{(\mI, \mJ)}\left(\mcD_1\right)}$ and $(\mX \not\indep \mY \mid \mZ, F)_{\operatorname{Aug}_{(\mI, \mJ)}\left(\mcD_2\right)}$. Suppose that both $\mX, \mY$ are observed variables. In this case, any tuple of interventional distribution obtained from an observational distribution that is faithful to the causal graph with latent variables constitutes a valid example. Suppose $\mX=F$ for some and $\mY \in \mathbf{V}$. Therefore, an $F$ node is m-connected to an observed node in $\operatorname{Aug}_{(\mI, \mJ) }\left(\mcD_2\right)$ but not in $\operatorname{Aug}_{(\mI, \mJ)}\left(\mcD_1\right)$.

Consider the causal graph \( \mcD_2 = (\mathbf{V} \cup \mathbf{L}, \mathbf{E}) \) with latent variables. Focus on the subgraph of \( \mcD_2 \) that includes all variables contributing to the m-connecting path of \( (\mX \not \indep \mY \mid \mZ, F)_{\operatorname{Aug}_{(\mI, \mJ)}(\mcD_2)} \). An example can be found in \citet{meek2013strong}. Let us call this subgraph \( \mcD_{\text{path}} = (V_{\text{path}}, E_{\text{path}}) \).  Consider a jointly Gaussian distribution on \( V_{\text{path}} \) that is faithful to \( \mcD_{\text{path}} \). Such a distribution exists by construction in Meek (Theorem 7, \cite{meek2013strong}). Denote this distribution by \( P_{\text{path}} \). We will focus on \( P_{\text{path}} \) and later expand it by adding the remaining variables in \( \mcD_{\text{suff}} \) as jointly independent and independent of the variables in \( D_{\text{path}} \).  Now, consider two intervention targets \( \mI \) and \( \mJ \) on the CBN \( (\mcD_{\text{path}}, P_{\text{path}}) \), where \( \mI \Delta \mJ = K \). This implies that the distributions \( P_\mI \) and \( P_\mJ \) account for the graphical separation of \( F_I \). For this proof, we treat \( F_I \) as a regime variable indicating when we switch between \( P_\mI \) and \( P_\mJ \). This treatment is valid because we add only this single \( F \) node, without introducing others. Define the distribution \( P^* \) as follows:  
\[
P^*(\cdot \mid F_I = 0) = P_\mI(\cdot), \quad 
P^*(\cdot \mid F_I = 1) = P_\mJ(\cdot),
\]
and let \( F_I \) follow a uniform distribution. We need to show that the invariances implied by the graph separation in the generalized causal calculus rules fail for \( P_\mI \) and \( P_\mJ \). This is equivalent to demonstrating that \( F_I \) is dependent on \( \mY \) given \( \mZ \) in the distribution \( P^* \). To construct the interventional distributions, we use a SCM that implies the given CBN. Let \( \mathbf{x} \) represent the vector of all variables in the graph, including latent ones. Consider the SCM given as follows:
\[
\mathbf{x} = \mathbf{A} \mathbf{x} + \mathbf{e},
\]  
where \( \mathbf{A} \) is a lower triangular matrix describing the graph structure and parent-child relationships in \( \mathcal{D}_{\text{path}} \), and \( \mathbf{e} \) is the vector of exogenous Gaussian noise terms, $|\mx| = n$.

Let \( P_\mI \) represent the distribution obtained by introducing the noise vector \( \mathbf{e}_I \) into the system. The vector \( \mathbf{e}_I \) is non-zero only in the rows \( i \) where \( x_i \in \mI \). Furthermore, the matrix \( \mathbf{A} \) is modified to a new matrix \( \mathbf{A}_1 \), which is identical to \( \mathbf{A} \) except that all entries in the rows corresponding to \( i \) ($\forall$\( x_i \in \mI \)) are set to zero. This modification effectively removes the influence of the parents of the variables in the intervention set \( \mI \), as required by the definition of a hard intervention. Consequently, \( P_\mI \) qualifies as a valid hard interventional distribution. Similarly, let \( \mathbf{e}_J \) denote the noise vector introduced for an intervention on \( \mJ \). The matrix \( \mathbf{A} \) is similarly replaced with \( \mathbf{A}_2 \), where all entries in the rows corresponding to \( j \) (\(\forall x_j \in \mJ \)) are set to zero, achieving the same effect for the variables in \( \mJ \). We show that in the combined distribution $p*$ using these $P_\mI, P_\mJ$ every adjacent variable is dependent. Clearly, when $\mathbf{e}_I$ and $\mathbf{e}_J$ are different, $F$ variable is dependent with the variables in $\mK:=\mI \Delta \mJ$, since $P^*(\mK \mid F=0) \neq P^*(\mK \mid F=1 \text { ), which implies ( } \mK \not \indep F \mid \emptyset)_{P^*}$. Therefore, we focus on establishing that every pair of variables that are adjacent are correlated except for the $F$ variable. The correlation of the variables in $\mcD_{\text {path }}$ is calculated as follows:
$$
\begin{aligned}
& \mathbf{x}=\mathbf{A_1} \mathbf{x}+\mathbf{e}+\mathbf{e}_I \Rightarrow(\mI_n - \mathbf{A_1}) \mathbf{x}=\mathbf{e}+ \mathbf{e}_I \Rightarrow \mathbf{x}=(\mI_n-\mathbf{A_1})^{-1} (\mathbf{e}+ \mathbf{e}_I) \\
& \mathbf{x}=\mathbf{A_2} \mathbf{x}+\mathbf{e}+\mathbf{e}_J \Rightarrow(\mI_n-\mathbf{A_2}) \mathbf{x}=\mathbf{e}+ \mathbf{e}_J \Rightarrow \mathbf{x}=(\mI_n-\mathbf{A_2})^{-1} (\mathbf{e}+ \mathbf{e}_J)
\end{aligned}
$$
Where $\mI_n$ is the identity matrix of size $n$ by $n$. Let \( \mathbf{e}_1 = \mathbf{e} + \mathbf{e}_I \) and \( \mathbf{e}_2 = \mathbf{e} + \mathbf{e}_J \). The correlation matrix between the observed variables, with respect to the distribution \( P^* \) after marginalizing out the binary regime variable, is computed as follows:
$$
\begin{aligned}
E\left[\mathbf{x x}^T\right] & =0.5(\mI_n -\mathbf{A_1})^{-1} E\left[\mathbf{e}_1 \mathbf{e}_1^T\right](\mI_n -\mathbf{A_1})^{-1^T}+0.5(\mI_n -\mathbf{A_2})^{-1} E\left[\mathbf{e}_2 \mathbf{e}_2^T\right](\mI_n-\mathbf{A_2})^{-1^T} 
\end{aligned}
$$

Let \( \mathbf{D}_1 = E\left[\mathbf{e}_1 \mathbf{e}_1^T\right] \) and \( \mathbf{D}_2 = E\left[\mathbf{e}_2 \mathbf{e}_2^T\right] \) represent the diagonal covariance matrices for the noise introduced by the hard interventions. Additionally, assume that all the noise variables, including \( \mathbf{e} \), \( \mathbf{e}_I \), and \( \mathbf{e}_J \), follow zero-mean Gaussian distributions. Consider two adjacent variables, \( x_i \) and \( x_j \), within \( \mathcal{D}_{\text{path}} \). We observe that the matrices \( \mI_n - \mathbf{A}_1 \) and \( \mI_n - \mathbf{A}_2 \) are full-rank, as \( \mathbf{A} \) is a strictly lower triangular matrix, and the same holds for \( \mathbf{A}_1 \) and \( \mathbf{A}_2 \). As a result, the matrix inverses in these equations exist and are unique.

We now treat \( \mathbf{D}_1 \) and \( \mathbf{D}_2 \) as variables in this system. When performing a hard intervention, we can freely select the variance of each added noise term. Our objective is to demonstrate that there always exist hard interventions, represented by \( \mathbf{D}_1 \) and \( \mathbf{D}_2 \), such that \( x_i \) and \( x_j \) become dependent. Since both \( x_i \) and \( x_j \) are jointly Gaussian, they are dependent if and only if they are correlated. Therefore, we need to show that \( E\left[x_i x_j\right] \neq 0 \) for any adjacent pair \( x_i, x_j \). If we set \( \mathbf{D}_1 = \mathbf{D}_2 = \mathbf{0} \), we return to the observational system. By the assumption that the original distribution is faithful to the graph \( \mathcal{D}_{\text{path}} \), any adjacent variables are dependent. This implies that the corresponding system of linear equations is not trivially zero. Hence, by randomly choosing the variances of the noise terms, we can ensure that, with probability 1, any adjacent pair of variables will be dependent (by applying a union bound).

Thus, we have shown that, in the graph \( \mathcal{D}_{\text{path}} \) along with the \( F \) variable, every pair of adjacent variables is dependent. Next, we can extend this distribution to include the variables outside \( \mathcal{D}_{\text{path}} \). To do this, we select the remaining variables to be jointly independent and independent from those in \( \mathcal{D}_{\text{path}} \). The interventional distributions can then be constructed by applying a similar hard intervention, where extra noise terms are added to the variables being intervened upon and replacing the matrix $\mathbf{A}$. The resulting set of interventional distributions will belong to \( \mathcal{P}_{\mathcal{I}}\left(\mathcal{D}_2, \mathbf{V}\right) \), but not to \( \mathcal{P}_{\mathcal{I}}\left(\mathcal{D}_1, \mathbf{V}\right) \). This is because m-separation should imply invariance across the interventional distributions, but we have constructed them in such a way that this condition does not hold. This concludes the proof.
\qed

\subsection{Proof for Proposition~\ref{prop: I aug mag I markov equivalent}}
\textit{Proof.} We have shown in Theorem~\ref{thm: I-MEC} that two causal graphs are $\mcI$-Markov equivalent if and only if their twin augmented MAGs satisfy the 3 conditions. Here we just need to show that the twin augmented MAGs follow the 3 conditions if and only if the $\mcI$-augmented MAGs follow the 3 conditions.

\textbf{(If:)} Notice that each twin augmented MAG $\Twin_{\mI, \mJ}(\mcD)$ can be constructed by taking the graph union of $\Aum_\mI(\mcD, \mcI)$ and $\Aum_\mJ(\mcD, \mcI)$ and removing irrelevant $F$ nodes. Therefore, if $\Aum_\mI(\mcD_1, \mcI)$ and $\Aum_\mJ(\mcD_2, \mcI)$ have the same skeleton for any $\mI \in \mcI$, $\Twin_{\mI, \mJ}(\mcD_1)$ and $\Twin_{\mI, \mJ}(\mcD_2)$ will also have the same skeleton. Furthermore, $\Twin_{\mI, \mJ}(\mcD)$ and $\Aum_\mI(\mcD, \mcI)$ have the same unshielded colliders within $\mV^{(\mI)} \cup \{ F^{(\mI, \mJ)} \}$ because they have the same subgraph on $\mV^{(\mI)} \cup \{ F^{(\mI, \mJ)} \}$. Since twin augmented MAGs do not have other $F$ nodes, $\Twin_{\mI, \mJ}(\mcD_1)$ and $\Twin_{\mI, \mJ}(\mcD_2)$ have the same unshielded colliders. Similarly, any discriminating path in $\Aum_\mI(\mcD, \mcI)$ will also be preserved in $\Twin_{\mI, \mJ}(\mcD)$ while there cannot be discriminating paths that pass through the $F$ node in $\Twin_{\mI, \mJ}(\mcD)$. Therefore, $\Twin_{\mI, \mJ}(\mcD_1)$ and $\Twin_{\mI, \mJ}(\mcD_2)$ have the same collider status on discriminating paths.

\textbf{(Only if:)} Now we show that if $\Twin_{\mI, \mJ}(\mcD_1)$ and $\Twin_{\mI, \mJ}(\mcD_2)$ satisfy the 3 conditions for any $\mI, \mJ \in \mcI$, then $\Aum_\mI(\mcD_1, \mcI)$ and $\Aum_\mI(\mcD_2, \mcI)$ also satisfy the 3 conditions for any $\mI \in \mcI$. Since $\Aum_\mI(\mcD_1, \mcI)$ is the graph union of $\Twin_{\mI, \mJ}(\mcD_1)$ on $\mV^{(\mI)} \cup \{ F^{(\mI, \mJ)} \}$, all the adjacencies are kept and the same for $\Aum_\mI(\mcD_2, \mcI)$. Thus, they have the same adjacencies. For unshielded triples, if both ends are $F$ nodes, then they have to be unshielded colliders in both $\Aum_\mI(\mcD_1, \mcI)$ and $\Aum_\mI(\mcD_2, \mcI)$ given $F$ nodes have only outgoing edges by construction. If at most one endpoint of the unshielded triple is an $F$ node, then the same structure can be retrieved from the relevant twin augmented MAG. Therefore, $\Aum_\mI(\mcD_1, \mcI)$ and $\Aum_\mI(\mcD_2, \mcI)$ have the same unshielded colliders. Finally, we need to show that if $p = \langle U, W_1, W_2, ..., W_k, Y, Z\rangle$ is a discriminating path in both $\Aum_\mI(\mcD_1, \mcI)$ and $\Aum_\mI(\mcD_2, \mcI)$, then $p$ has the same collider status. If $p$ is in $\mV^{(\mI)}$, then $p$ is also in $\Twin_{\mI, \mJ}(\mcD_1)$ and $\Twin_{\mI, \mJ}(\mcD_2)$ for any $\mJ \in \mcI$ and thus it shows the same collider status in both $\mcI$-augmented MAGs. If $F^{(\mI, \mJ)}$ is an endpoint of $p$, then the path is also in $\Twin_{\mI, \mJ}(\mcD_1)$ and $\Twin_{\mI, \mJ}(\mcD_2)$ with the same collider status. By the construction of $\mcI$-augmented MAGs, an $F$ node can only be $Y$ in $p$ if it is not the starting node. Since $F$ nodes only have outgoing edges, $p$ will have the same collider status in both $\Aum_\mI(\mcD_1, \mcI)$ and $\Aum_\mI(\mcD_2, \mcI)$,
\qed

\subsection{Proof for Theorem~\ref{thm: soundness of algo}}
The algorithm is learning the causal graph through finding the separating sets between each pair of nodes using the distributional invariance tests. The invariance tests are tied to the m-separation statements in the causal graph according to the h-faithfulness assumption, and the properties in Definition~\ref{def: I-Markov} are mapped to the m-separation statements in the twin augmented MAGs by Proposition~\ref{prop: F node}. We show that twin augmented MAGs are combined to construct $\mcI$-augmented MAGs which preserve the m-separation statements in Proposition~\ref{prop: I aug mag I markov equivalent}. Ideally, our algorithm would learn a structure that is close to the $\mcI$-augmented MAGs. Therefore, to show the soundness of Algorithm~\ref{alg: I-MEC learning}, we need to first define the $\mcI$-essential graph as follows:
\begin{definition}[$\mcI$-essential graph]
    \label{def: I-essential graph}
    Given a causal graph $\mcD = (\mV \cup \mL, \mE)$ and a set of intervention targets $\mcI \subseteq 2^V$, the $\mcI$-essential graph of $\mcD$ related to intervention targets $\mcI$ and a target $\mI \in \mcI$, denoted as $\mcE_{I}(\mcD, \mcI)$ is the union graph of the $\mcI$-augmented MAGs of $\mcD^\prime$ for all $\mcD^\prime$ that are ADMGs $\mcI$-Markov equivalent to $\mcD$. 
\end{definition}

\begin{definition}[Union Graph of \( k \) ADMGs]
\label{def: union graph}
Let \( \mathcal{G}_1, \mathcal{G}_2, \dots, \mathcal{G}_k \) be \( k \) ADMGs, where each graph \( \mathcal{G}_i = (\mV_i, \mE_i) \) 

The \textbf{union graph} of \( k \) ADMGs, denoted as \( \mathcal{G}_\cup = (\mV_\cup, \mE_\cup) \), where
$\mV_\cup = \bigcup_{i=1}^{k} \mV_i$,
for each pair of vertices \( (X, Y) \), the edge set of \( \mathcal{G}_\cup \) is determined by:

\[
\mathcal{E}_\cup = \left\{
\begin{array}{ll}
X \to Y, & \text{if } (X \to Y) \text{ appears in all } \mathcal{G}_i, X, Y \in \mV_i. \\
X \leftrightarrow Y, & \text{if } (X \leftrightarrow Y) \text{ appears in all } \mathcal{G}_i, X, Y\in \mV_i. \\
X \crightarrow Y, & \text{if } (X \to Y) \;\text{appears in some } \mathcal{G}_i, (X\leftrightarrow Y) \;\text{appears in some } \mcG_j, X, Y\in \mV_i, \mV_j. \\
X \circlecircle Y, & \text{if } (X \leftarrow Y)\; \text{ appears in some 
 } \mathcal{G}_i, (X\rightarrow Y) \text{appears in some } \mcG_j, X, Y\in \mV_i, \mV_j.
\end{array}
\right.
\]

\end{definition}

The $\mcI$-essential graphs denote the structure that is ultimately learnable by any causal discovery algorithm. Based on this definition, to demonstrate the soundness of our algorithm, it suffices to address the following three questions:
\begin{enumerate}[label=(\alph*)]
    \item Are the separating sets identified by Algorithm~\ref{alg: find sep set} valid?
    \item Do all the $\mcI$-augmented graphs returned by Algorithm~\ref{alg: I-MEC learning} have the same adjacencies as the $\mcI$-essential graphs?
    \item Are the orientation rules sound, i.e. any arrowhead/arrowtail learned by the algorithm is also present in the $\mcI$-essential graph?
\end{enumerate}

We first address (a). For a pair of nodes with at least one non-$F$ node in $\mcG_\mI$, if they are separable in the graph, there exists some $\mW$ that preserves the invariance in the statistics test. $\mW$ may contain some $F$ nodes. In Algorithm~\ref{alg: find sep set}, we put all $F$ nodes in $\mW$ since $F$ nodes are all designed to be source nodes and thus adding them does not activate any d-connecting path. If both nodes are $F$ nodes, we set the separating set to be $\emptyset$ since by the design of $\mcG_\mI$, there is no edge between $F$ nodes. Also notice that in Proposition~\ref{prop: generalized F-node}, the separating sets include the interventional target of that domain, and thus in Algorithm~\ref{alg: find sep set}, we also put the domain targets into the $SepSet$. While this is correct according to the hard do-calculus rules, we claim that removing the domain targets from the separating sets will not lead to wrong edge orientations. Concretely, although there might be multiple separating sets for a pair of nodes, we will show that our choice of the separating sets does not lead to wrong orientations of the edges. To verify this, we just need to check the orientation rules that require using separating sets.

\textbf{Orienting Unshielded Colliders:} Consider an unshielded triple $\langle X, Y, Z\rangle$ in $\mcG_\mI$. If it is a collider, conditioning on $F$ nodes would not open any extra path between $X, Z$ since there are no edges into $F$ nodes. Similarly, the intervention target $\mI$ does not have any incoming edges thus cannot a collider or a descendant of any collider. We do not need to condition on the target to identify any collider. If the triple is not a collider, the target would either be in the identified separating set if it is a confounder of $X, Z$ or it is not in the set when $X, Z$ are already separated by $SepSet$.

\textbf{Discriminating Paths:} By the definition of discriminating path and construction of $\mcG_\mI$, there cannot be discriminating paths between pairs of $F$ nodes. Consider the path $\langle X, ..., W, U, Y\rangle$. If $X\in \mcF$ and $Y \in \mV$, notice that no $F$ nodes can be in between $X, U$ since they cannot be colliders. If $U$ is not an $F$ node, then adding extra $F$ nodes into $SepSet$ does not affect the rule. If $U$ is an $F$ node, then it will be included in the $SepSet$ according to Algorithm~\ref{alg: find sep set}. The rule will then orient $U\rightarrow Y$ which aligns with Rule 8. Similarly, if $U$ is in the target set, it has to be a confounder to its neighbours on the path and will be included by $SepSet$. 

Then, we address (b). Consider an $\mcI$-augmented graph $\mcG_I$, the edges either contain a $F$ node or not. There are no edges between any two $F$ nodes by our construction of $\mcG_\mI$ and $\Aum_\mI(\mcD, \mcI)$. Consider the edge $(F^{(\mI, \mJ)}, Y^{(\mI)})$ in $\mcG_\mI$. It is recovered because there does not exist $\mW \subseteq \mV$, such that $P_\mI(y|\mw) = P_\mJ(y|\mw)$ under h-faithfulness. According to Definition~\ref{def: I-Markov} and Proposition~\ref{prop: F node}, it implies that $F^{(\mI, \mJ)}$ and $Y^{(\mI)}, Y^{(\mJ)}$ are also adjacent in $\Twin_{(\mI, \mJ)}(\mcD^\prime)$ for any $\mcD^\prime$ that is $\mcI$-Markov equivalent to $\mcD$. Thus $F^{(\mI, \mJ)}$ and $Y^{(\mI)}$ are also adjacent in $\Aum_\mI(\mcD^\prime, \mcI)$. Consequently, they are also adjacent in the $\mcI$-essential graph. For the same reason, if $F^{(\mI, \mJ)}$ and $Y^{(\mI)}$ are not adjacent in $\mcG_\mI$, we can derive that they are also not adjacent in $\mcE_I(\mcD, \mcI)$. 

Next, consider the edges that do not contain $F$ nodes. There are no edges between two nodes in different domains in both $\mcG_\mI$ and $\Aum_\mI(\mcD, \mcI)$ by our construction. We call the vertex induced subgraph on $\mV^{(\mJ)}, \mJ \in \mcI$ a domain. We just need to show that they have the same adjacencies within each domain. According to our algorithm, within each domain, all the edges are connected by applying the FCI algorithm, which is proved to be sound by~\cite{zhang2008completeness}. Thus, $\mcG_\mI[\mV^{(\mI)}]$ will have the same adjacencies as $\Aum_\mI(\mcD^\prime, \mcI)[\mV^{(\mI)}], \mI \in \mcI$ for all $\mcD^\prime$ that are $\mcI$-Markov equivalent to $\mcD$. 

Finally, we address the soundness of orientation rules. In phase I of the learning algorithm, we use FCI rules within each domain to learn the skeletons. \citet{zhang2008completeness} showed that FCI is sound and complete. Thus, the orientations in the skeleton learned in this phase are shared across all $\Aum_\mI(\mcD^\prime)$, for all $\mcD^\prime$ that is $\mcI$-Markov equivalent to $\mcD$. 
In phase II, $F$ nodes are introduced and adjacent to the symmetric difference of the targets, and the nodes that are not separable from $F$. All the edges induced to $F$ nodes are oriented outgoing from $F$ nodes by Rule 8. This is also true in $\Aum_\mI(\mcD^\prime, \mcI)$ for all $\mcD^\prime$ that is $\mcI$-Markov equivalent to $\mcD$. Thus, so far, all the orientations learned are sound.
What remains to be shown is the soundness of the extra orientation rules. 

\textbf{Soundness of Rule $0$:} If both end nodes are in $\mV$, then the soundness is guaranteed by the soundness of FCI. Suppose $\langle F, X^{(\mI)}, Y^{(\mI)}\rangle$ is an unshielded collider identified by the learning algorithm, i.e., there is some $\mW \subseteq \mV \setminus \mI$, such that, $F \indep Y^{(\mI)}| \mI^{(\mI)}, \mW^{(\mI)}$, while $X \notin \mW \cup \mI$. Suppose otherwise, $\langle F, X^{(\mI)}, Y^{(\mI)}\rangle$ is not an unshielded collider, then it can only be $F\rightarrow X^{(\mI)} \rightarrow Y^{(\mI)}$. Since $F, Y^{(\mI)}$ are non-adjacent, the soundness of the skeleton indicates that there is a set of nodes that separates $F$ from $Y^{(\mI)}$. However, in this case, to separate $F$ from $Y^{(\mI)}$, $X^{(\mI)}$ has to be in the condition set; otherwise, the path $F\rightarrow X^{(\mI)} \rightarrow Y^{(\mI)}$ would be d-connecting. 

\textbf{Soundness of Rule 9:} 
To address the soundness of Rule 9, we need to show that if $X^{(\mI)}, Y^{(\mI)}$ are adjacent, and $X\in \mI$, then $Y$ can only be a descendant of $X$ in the interventional causal graph $\mcD_{\overline{\mI}}$. Suppose otherwise, then $Y$ is an ancestor of $X$ or they have at least one common ancestor in $\mcD_{\overline{\mI}}$. However, since $X$ is intervened, there will be no ancestor of $X$ in $\mcD_{\overline{\mI}}$, which is a contradiction.

\textbf{Soundness of Rule 10:} We need to show that if we recover $X\rightarrow Y$ in the domain $G^{(\mI)}$, there cannot be $X^{(\mJ)}\leftarrow Y^{(\mJ)}$ in another domain of $\mJ$. Suppose otherwise, it indicates that there is a directed path from $Y$ to $X$ in $\mcD_{\overline{\mJ}}$. However, there is also a directed path from $X$ to $Y$ in $\mcD_{\overline{\mI}}$. $\mcD_{\overline{\mJ}}$ and $\mcD_{\overline{\mI}}$ are subgraphs of $\mcD$, thus there is at least one cycle in $\mcD$ which is a contradiction.

\textbf{Soundness of Rule 11:}
$F^{(\mI, \mJ)}$ is adjacent to $Y^{(\mI)}, Y^{(\mJ)}$ while $Y \notin \mI$ shows that there does not exist any $\mW \subseteq \mV$ such that $P_\mI(y|\mw) = P_\mJ(y|\mw)$. This means that there is no separating set that separates both $F^{(\mI, \mJ)}, Y^{(\mI)}$ and $F^{(\mI, \mJ)}, Y^{(\mJ)}$. This indicates that there is an inducing path from $F^{(\mI, \mJ)}$ to $ Y^{(\mI)}$
or from $F^{(\mI, \mJ)}$ to $ Y^{(\mJ)}$ relative to the latent variables in the augmented pair graph. Due to the definition of the inducing path, it has to go through $X^{(\mI)}$ or $X^{(\mJ)}$ as $\mK = \{ X\}$. However, in $G^{(\mI)}$, since $X^{(\mI)}$ is intervened, $X^{(\mI)}$ cannot serve as a collider in the path. Consequently, we can infer that the inducing path has to go through $X^{(\mJ)}$ to $Y^{(\mJ)}$. Furthermore, $X^{(\mJ)}$ has to be a parent of one of the end nodes of the path. Given that $F$ nodes are source nodes, we can orient $X^{(\mJ)}\rightarrow Y^{(\mJ)}$. 

\qed

\newpage

\section{An Example of the Learning Process}
\label{app: ex learning}
\begin{figure}[H]
    \centering
    \begin{subfigure}{0.24\columnwidth}
        \centering
        \begin{tikzpicture}
            \node (1) {{$X^{(1)}$}};
            \node (2) [right of = 1,xshift = 0.6 cm] {{$Z^{(1)}$}};
              \node (3) [right of = 2,xshift = 0.6 cm] { {$Y^{(1)}$}}; 
              \node (4) [below of = 2] {$F$};

        \draw[-] (1) -- (2)
            node[pos=0.19, left] {\(\circ\)} 
            node[pos=0.82, right] {\(\circ\)}; 
        \draw[-, bend left=45] (1) to 
            node[pos= 0.08, yshift = -12 pt] {\(\circ\)} 
            node[pos= 0.92, yshift = -11 pt] {\(\circ\)}
            (3);
        \draw[-] (2) -- (3)
            node[pos= 0.19, left] {\(\circ\)} 
            node[pos= 0.82, right] {\(\circ\)}; 
        \draw[-] (4) -- (1)
            node[pos= 0.3, xshift = 17] {\(\circ\)} 
            node[yshift = -7, xshift = 11] {\(\circ\)}; 
        \draw[-] (4) -- (2)
            node[xshift = 0, yshift = -6] {\(\circ\)} 
            node[xshift = 0, yshift = -23] {\(\circ\)}; 
        \draw[-] (4) -- (3)
            node[yshift = -24, xshift = -40] {\(\circ\)} 
            node[yshift = -7, xshift = -12] {\(\circ\)}; 
        
        \end{tikzpicture}
        \caption{Initialize $\mcG_\emptyset(\mcD_{1})$}
        \label{fig: D_1 obs pag}
    \end{subfigure}%
    \begin{subfigure}{0.24\columnwidth}
        \centering
        \begin{tikzpicture}
            \node (1) {{$X^{(1)}$}};
            \node (2) [right of = 1,xshift = 0.6 cm] {{$Z^{(1)}$}};
              \node (3) [right of = 2,xshift = 0.6 cm] { {$Y^{(1)}$}}; 
              \node (4) [below of = 2] {$F$};

        \draw[-] (1) -- (2)
            node[pos=0.19, left] {\(\circ\)} 
            node[pos=0.82, right] {\(\circ\)}; 
        \draw[-, bend left=45] (1) to 
            node[pos= 0.08, yshift = -12 pt] {\(\circ\)} 
            node[pos= 0.92, yshift = -11 pt] {\(\circ\)}
            (3);
        \draw[-] (2) -- (3)
            node[pos= 0.19, left] {\(\circ\)} 
            node[pos= 0.82, right] {\(\circ\)}; 
        \draw[-] (4) -- (1)
            node[pos= 0.3, xshift = 17] {\(\circ\)} 
            node[yshift = -7, xshift = 11] {\(\circ\)}; 
        \draw[-] (4) -- (2)
            node[xshift = 0, yshift = -6] {\(\circ\)} 
            node[xshift = 0, yshift = -23] {\(\circ\)}; 
        \draw[-] (4) -- (3)
            node[yshift = -24, xshift = -40] {\(\circ\)} 
            node[yshift = -7, xshift = -12] {\(\circ\)}; 
        \end{tikzpicture}
        \caption{Initialize $\mcG_\emptyset(\mcD_{2})$}
        \label{fig: D_2 obs pag}
    \end{subfigure}
\begin{subfigure}{0.24\columnwidth}
        \centering
        \begin{tikzpicture}
            \node (1) {{$X^{(2)}$}};
            \node (2) [right of = 1,xshift = 0.6 cm] {{$Z^{(2)}$}};
              \node (3) [right of = 2,xshift = 0.6 cm] { {$Y^{(2)}$}}; 
              \node (4) [below of = 2] {$F$};

        \draw[-] (1) -- (2)
            node[pos=0.19, left] {\(\circ\)} 
            node[pos=0.82, right] {\(\circ\)}; 
        \draw[-, bend left=45] (1) to 
            node[pos= 0.08, yshift = -12 pt] {\(\circ\)} 
            node[pos= 0.92, yshift = -11 pt] {\(\circ\)}
            (3);
        \draw[-] (2) -- (3)
            node[pos= 0.19, left] {\(\circ\)} 
            node[pos= 0.82, right] {\(\circ\)}; 
        \draw[-] (4) -- (1)
            node[pos= 0.3, xshift = 17] {\(\circ\)} 
            node[yshift = -7, xshift = 11] {\(\circ\)}; 
        \draw[-] (4) -- (2)
            node[xshift = 0, yshift = -6] {\(\circ\)} 
            node[xshift = 0, yshift = -23] {\(\circ\)}; 
        \draw[-] (4) -- (3)
            node[yshift = -24, xshift = -40] {\(\circ\)} 
            node[yshift = -7, xshift = -12] {\(\circ\)}; 
        \end{tikzpicture}
        \caption{Initialize $\mcG_{ \{Z\}}(\mcD_1)$}
        \label{fig: D_1 int Z}
    \end{subfigure}%
    \begin{subfigure}{0.24\columnwidth}
        \centering
        \begin{tikzpicture}
            \node (1) {{$X^{(2)}$}};
            \node (2) [right of = 1,xshift = 0.6 cm] {{$Z^{(2)}$}};
              \node (3) [right of = 2,xshift = 0.6 cm] { {$Y^{(2)}$}}; 
              \node (4) [below of = 2] {$F$};

        \draw[-] (1) -- (2)
            node[pos=0.19, left] {\(\circ\)} 
            node[pos=0.82, right] {\(\circ\)}; 
        \draw[-, bend left=45] (1) to 
            node[pos= 0.08, yshift = -12 pt] {\(\circ\)} 
            node[pos= 0.92, yshift = -11 pt] {\(\circ\)}
            (3);
        \draw[-] (2) -- (3)
            node[pos= 0.19, left] {\(\circ\)} 
            node[pos= 0.82, right] {\(\circ\)}; 
        \draw[-] (4) -- (1)
            node[pos= 0.3, xshift = 17] {\(\circ\)} 
            node[yshift = -7, xshift = 11] {\(\circ\)}; 
        \draw[-] (4) -- (2)
            node[xshift = 0, yshift = -6] {\(\circ\)} 
            node[xshift = 0, yshift = -23] {\(\circ\)}; 
        \draw[-] (4) -- (3)
            node[yshift = -24, xshift = -40] {\(\circ\)} 
            node[yshift = -7, xshift = -12] {\(\circ\)}; 
        \end{tikzpicture}
        \caption{Initialize $\mcG_{ \{Z\} }(\mcD_2)$}
        \label{fig: D_2 int Z}
    \end{subfigure}

    \begin{subfigure}{0.24\columnwidth}
        \centering
        \begin{tikzpicture}
            \node (1) {{$X^{(1)}$}};
            \node (2) [right of = 1,xshift = 0.6 cm] {{$Z^{(1)}$}};
              \node (3) [right of = 2,xshift = 0.6 cm] { {$Y^{(1)}$}};    
              \node (4) [below of = 2, yshift = -0.1 cm] {$F$};
              \node (8) [below of = 4, yshift = 0 cm] {$F$};
              \node (5) [below of = 8, yshift= -0.1 cm] {$Z^{(2)}$};
              \node (6) [left of = 5, xshift= -0.6 cm] {$X^{(2)}$};
              \node (7) [right of = 5, xshift= 0.6 cm] {$Y^{(2)}$};

        \draw[-] (1) -- (2) 
            node[pos=0.25, left] {\(\circ\)} 
            node[pos=0.75, right] {\(\circ\)}; 
        \draw[-] (2) -- (3) 
            node[pos=0.25, left] {\(\circ\)} 
            node[pos=0.75, right] {\(\circ\)}; 
        \draw[-, bend left=45] (1) to 
            node[pos=0.08, yshift = -12 pt] {\(\circ\)} 
            node[pos=0.92, yshift = -11 pt] {\(\circ\)}
            (3);
        \draw[-] (4) -- (2)
            node[pos=-0.1, xshift = 6 pt] {\(\circ\)} 
            node[pos=1.1, xshift = 6 pt] {\(\circ\)}; 
        \draw[-] (8) -- (5)
            node[pos=-0.1, xshift = -6 pt] {\(\circ\)} 
            node[pos=1.1, xshift = -6 pt] {\(\circ\)}; 
        \draw[-] (8) -- (7)
            node[pos=0, xshift = -7 pt, yshift = -5 pt] {\(\circ\)} 
            node[pos=1.05, xshift = -5 pt, yshift = -6 pt] {\(\circ\)}; 
        \draw[-] (4) -- (3)
            node[pos=0.15, yshift = -10 pt] {\(\circ\)} 
            node[pos=1.05, xshift = 7 pt, yshift = -6 pt] {\(\circ\)}; 
        \draw[-] (5) -- (7)
            node[pos=0.2, left] {\(\circ\)} 
            node[pos=0.8, right] {\(\circ\)}; 
        \end{tikzpicture}
        \caption{$\mcG_{\emptyset}(\mcD_1)$ and $\mcG_{\{Z\}}(\mcD_1)$ \\After Phase II}
        \label{fig: PAG D_1 phase II}
    \end{subfigure}%
    \begin{subfigure}{0.24\columnwidth}
        \centering
        \begin{tikzpicture}
            \node (1) {{$X^{(1)}$}};
            \node (2) [right of = 1,xshift = 0.6 cm] {{$Z^{(1)}$}};
              \node (3) [right of = 2,xshift = 0.6 cm] { {$Y^{(1)}$}};    
              \node (4) [below of = 2, yshift = -0.1 cm] {$F$};
              \node (8) [below of = 4, yshift = 0 cm] {$F$};
              \node (5) [below of = 8, yshift= -0.1 cm] {$Z^{(2)}$};
              \node (6) [left of = 5, xshift= -0.6 cm] {$X^{(2)}$};
              \node (7) [right of = 5, xshift= 0.6 cm] {$Y^{(2)}$};

        \draw[-] (1) -- (2)
            node[pos=0.25, left] {\(\circ\)} 
            node[pos=0.75, right] {\(\circ\)}; 
        \draw[-] (2) -- (3) 
            node[pos=0.25, left] {\(\circ\)} 
            node[pos=0.75, right] {\(\circ\)}; 
        \draw[-, bend left=45] (1) to 
            node[pos=0.08, yshift = -12 pt] {\(\circ\)} 
            node[pos=0.92, yshift = -11 pt] {\(\circ\)}
            (3);
        \draw[-] (4) -- (2)
            node[pos=-0.1, xshift = 6 pt] {\(\circ\)} 
            node[pos=1.1, xshift = 6 pt] {\(\circ\)}; 
        \draw[-] (8) -- (5)
            node[pos=-0.1, xshift = -6 pt] {\(\circ\)} 
            node[pos=1.1, xshift = -6 pt] {\(\circ\)}; 
        \draw[-] (8) -- (7)
            node[pos=0, xshift = -7 pt, yshift = -5 pt] {\(\circ\)} 
            node[pos=1.05, xshift = -5 pt, yshift = -6 pt] {\(\circ\)}; 
        \draw[-] (4) -- (3)
            node[pos=0.18, yshift = -10 pt] {\(\circ\)} 
            node[pos=1.05, xshift = 7 pt, yshift = -6 pt] {\(\circ\)}; 
        \draw[-] (5) to 
            node[pos=0.25, left] {\(\circ\)} 
            node[xshift = 12 pt, yshift = -6 pt] {\(\circ\)}
            (7); 
        \draw[-, bend right=45] (6) to
            node[pos=0, xshift = -6 pt, yshift = -4 pt] {\(\circ\)} 
            node[xshift = 38 pt, yshift = 11 pt] {\(\circ\)}
            (7);
        \end{tikzpicture}
        \caption{$\mcG_{\emptyset}(\mcD_2)$ and $\mcG_{\{Z\}}(\mcD_2)$ \\After Phase II}
        \label{fig: PAG D_2 phase II}
    \end{subfigure}
    \begin{subfigure}{0.24\columnwidth}
        \centering
        \begin{tikzpicture}
            \node (1) {{$X^{(1)}$}};
            \node (2) [right of = 1,xshift = 0.6 cm] {{$Z^{(1)}$}};
              \node (3) [right of = 2,xshift = 0.6 cm] { {$Y^{(1)}$}};    
              \node (4) [below of = 2, yshift = -0.1 cm] {$F$};
              \node (8) [below of = 4, yshift = 0 cm] {$F$};
              \node (5) [below of = 8, yshift= -0.1 cm] {$Z^{(2)}$};
              \node (6) [left of = 5, xshift= -0.6 cm] {$X^{(2)}$};
              \node (7) [right of = 5, xshift= 0.6 cm] {$Y^{(2)}$};

        \draw[->] (1) -- (2) 
            node[pos=0.25, left] {\(\circ\)}; 
        \draw[-] (2) -- (3) 
            node[pos=0.25, left] {\(\circ\)} 
            node[pos=0.75, right] {\(\circ\)}; 
        \draw[->, bend left=45] (1) to 
            node[pos=0.08, yshift = -12 pt] {\(\circ\)} 
            (3);
        \draw[->] (4) -- (2)
            node[pos=-0.1, xshift = 6 pt] {\(\circ\)} 
            ; 
        \draw[-] (8) -- (5)
            node[pos=-0.1, xshift = -6 pt] {\(\circ\)} 
            node[pos=1.1, xshift = -6 pt] {\(\circ\)}; 
        \draw[-] (8) -- (7)
            node[pos=0, xshift = -7 pt, yshift = -5 pt] {\(\circ\)} 
            node[pos=1.05, xshift = -5 pt, yshift = -6 pt] {\(\circ\)}; 
        \draw[->] (4) -- (3)
            node[pos=0.15, yshift = -10 pt] {\(\circ\)} 
            ; 
        \draw[-] (5) -- (7)
            node[pos=0.2, left] {\(\circ\)} 
            node[pos=0.8, right] {\(\circ\)}; 
        \end{tikzpicture}
        \caption{$\mcG_{\emptyset}(\mcD_1)$ and $\mcG_{\{Z\}}(\mcD_1)$\\ After Rule 0}
        \label{fig: After Rule 8 D_1}
    \end{subfigure}%
    \begin{subfigure}{0.24\columnwidth}
        \centering
        \begin{tikzpicture}
            \node (1) {{$X^{(1)}$}};
            \node (2) [right of = 1,xshift = 0.6 cm] {{$Z^{(1)}$}};
              \node (3) [right of = 2,xshift = 0.6 cm] { {$Y^{(1)}$}};    
              \node (4) [below of = 2, yshift = -0.1 cm] {$F$};
              \node (8) [below of = 4, yshift = 0 cm] {$F$};
              \node (5) [below of = 8, yshift= -0.1 cm] {$Z^{(2)}$};
              \node (6) [left of = 5, xshift= -0.6 cm] {$X^{(2)}$};
              \node (7) [right of = 5, xshift= 0.6 cm] {$Y^{(2)}$};

        \draw[->] (1) -- (2)
            node[pos=0.25, left] {\(\circ\)} 
            ; 
        \draw[-] (2) -- (3) 
            node[pos=0.25, left] {\(\circ\)} 
            node[pos=0.75, right] {\(\circ\)}; 
        \draw[->, bend left=45] (1) to 
            node[pos=0.08, yshift = -12 pt] {\(\circ\)} 
            (3);
        \draw[->] (4) -- (2)
            node[pos=-0.1, xshift = 6 pt] {\(\circ\)} 
            ; 
        \draw[-] (8) -- (5)
            node[pos=-0.1, xshift = -6 pt] {\(\circ\)} 
            node[pos=1.1, xshift = -6 pt] {\(\circ\)}; 
        \draw[->] (8) -- (7)
            node[pos=0, xshift = -7 pt, yshift = -5 pt] {\(\circ\)} 
            ; 
        \draw[->] (4) -- (3)
            node[pos=0.18, yshift = -10 pt] {\(\circ\)} 
            ; 
        \draw[->] (5) to 
            node[pos=0.25, left] {\(\circ\)} 
            (7); 
        \draw[->, bend right=45] (6) to
            node[pos=0, xshift = -6 pt, yshift = -4 pt] {\(\circ\)} 
            (7);
        \end{tikzpicture}
        \caption{$\mcG_{\emptyset}(\mcD_2)$ and $\mcG_{\{Z\}}(\mcD_2)$\\ After Rule 0}
        \label{fig: After Rule 8 D_2}
    \end{subfigure}

    \begin{subfigure}{0.24\columnwidth}
        \centering
        \begin{tikzpicture}
            \node (1) {{$X^{(1)}$}};
            \node (2) [right of = 1,xshift = 0.6 cm] {{$Z^{(1)}$}};
              \node (3) [right of = 2,xshift = 0.6 cm] { {$Y^{(1)}$}};    
              \node (4) [below of = 2, yshift = -0.1 cm] {$F$};
              \node (8) [below of = 4, yshift = 0 cm] {$F$};
              \node (5) [below of = 8, yshift= -0.1 cm] {$Z^{(2)}$};
              \node (6) [left of = 5, xshift= -0.6 cm] {$X^{(2)}$};
              \node (7) [right of = 5, xshift= 0.6 cm] {$Y^{(2)}$};

        \draw[->] (1) -- (2)
             node[pos=0.25, left] {\(\circ\)} ;
        \draw[-] (2) -- (3) 
            node[pos=0.25, left] {\(\circ\)} 
            node[pos=0.75, right] {\(\circ\)}; 
        \draw[->, bend left=45] (1) to 
            node[pos=0.08, yshift = -12 pt] {\(\circ\)} 
            (3);
        \draw[->] (4) -- (2) ;
        \draw[->] (8) -- (5) ;
        \draw[->] (4) -- (3) ;
        \draw[->] (8) -- (7) ;
        \draw[-] (5) -- (7) 
            node[pos=0.25, left] {\(\circ\)} 
            node[pos=0.75, right] {\(\circ\)}; 
        \end{tikzpicture}
        \caption{$\mcG_{\emptyset}(\mcD_1)$ and $\mcG_{\{Z\}}(\mcD_1)$\\ After Rule 8}
        \label{fig: After Rule 0' D_1}
    \end{subfigure}%
    \begin{subfigure}{0.24\columnwidth}
        \centering
        \begin{tikzpicture}
            \node (1) {{$X^{(1)}$}};
            \node (2) [right of = 1,xshift = 0.6 cm] {{$Z^{(1)}$}};
              \node (3) [right of = 2,xshift = 0.6 cm] { {$Y^{(1)}$}};    
              \node (4) [below of = 2, yshift = -0.1 cm] {$F$};
              \node (8) [below of = 4, yshift = 0 cm] {$F$};
              \node (5) [below of = 8, yshift= -0.1 cm] {$Z^{(2)}$};
              \node (6) [left of = 5, xshift= -0.6 cm] {$X^{(2)}$};
              \node (7) [right of = 5, xshift= 0.6 cm] {$Y^{(2)}$};

        \draw[->] (1) -- (2)
             node[pos=0.25, left] {\(\circ\)} ;
        \draw[-] (2) -- (3) 
            node[pos=0.25, left] {\(\circ\)} 
            node[pos=0.75, right] {\(\circ\)}; 
        \draw[->, bend left=45] (1) to 
            node[pos=0.08, yshift = -12 pt] {\(\circ\)} 
            (3);
        \draw[->] (4) -- (2) ;
        \draw[->] (8) -- (5) ;
        \draw[->] (4) -- (3) ;
        \draw[->] (8) -- (7) ;
        \draw[->] (5) -- (7)
            node[pos=0.25, left] {\(\circ\)} 
            ; 
        \draw[->, bend right=45] (6) to 
            node[pos=0, xshift = -6 pt, yshift = -4 pt] {\(\circ\)} 
            (7);
        \end{tikzpicture}
        \caption{$\mcG_{\emptyset}(\mcD_2)$ and $\mcG_{\{Z\}}(\mcD_2)$\\ After Rule 8}
        \label{fig: After Rule 0' D_2}
    \end{subfigure}
    \begin{subfigure}{0.24\columnwidth}
        \centering
        \begin{tikzpicture}
            \node (1) {{$X^{(1)}$}};
            \node (2) [right of = 1,xshift = 0.6 cm] {{$Z^{(1)}$}};
              \node (3) [right of = 2,xshift = 0.6 cm] { {$Y^{(1)}$}};    
              \node (4) [below of = 2, yshift = -0.1 cm] {$F$};
              \node (8) [below of = 4, yshift = 0 cm] {$F$};
              \node (5) [below of = 8, yshift= -0.1 cm] {$Z^{(2)}$};
              \node (6) [left of = 5, xshift= -0.6 cm] {$X^{(2)}$};
              \node (7) [right of = 5, xshift= 0.6 cm] {$Y^{(2)}$};

        \draw[->] (1) -- (2)
             node[pos=0.25, left] {\(\circ\)} ;
        \draw[->] (2) -- (3) ;
        \draw[->, bend left=45] (1) to 
            node[pos=0.08, yshift = -12 pt] {\(\circ\)} 
            (3);
        \draw[->] (4) -- (2) ;
        \draw[->] (8) -- (5) ;
        \draw[->] (4) -- (3) ;
        \draw[->] (8) -- (7) ;
        \draw[->] (5) -- (7) ;
        \end{tikzpicture}
        \caption{$\mcG_{\emptyset}(\mcD_1)$ and $\mcG_{\{Z\}}(\mcD_1)$}
        \label{fig:  twin pag D_1}
    \end{subfigure}%
    \begin{subfigure}{0.24\columnwidth}
        \centering
        \begin{tikzpicture}
            \node (1) {{$X^{(1)}$}};
            \node (2) [right of = 1,xshift = 0.6 cm] {{$Z^{(1)}$}};
              \node (3) [right of = 2,xshift = 0.6 cm] { {$Y^{(1)}$}};    
              \node (4) [below of = 2, yshift = -0.1 cm] {$F$};
              \node (8) [below of = 4, yshift = 0 cm] {$F$};
              \node (5) [below of = 8, yshift= -0.1 cm] {$Z^{(2)}$};
              \node (6) [left of = 5, xshift= -0.6 cm] {$X^{(2)}$};
              \node (7) [right of = 5, xshift= 0.6 cm] {$Y^{(2)}$};

        \draw[->] (1) -- (2)
             node[pos=0.25, left] {\(\circ\)} ;
        \draw[->] (2) -- (3) ;
        \draw[->, bend left=45] (1) to 
            node[pos=0.08, yshift = -12 pt] {\(\circ\)} 
            (3);
        \draw[->] (4) -- (2) ;
        \draw[->] (8) -- (5) ;
        \draw[->] (4) -- (3) ;
        \draw[->] (8) -- (7) ;
        \draw[->] (5) -- (7) ;
        \draw[->, bend right=45] (6) to 
            node[pos=0, xshift = -6 pt, yshift = -4 pt] {\(\circ\)} 
            (7);
        \end{tikzpicture}
        \caption{$\mcG_{\emptyset}(\mcD_2)$ and $\mcG_{\{Z\}}(\mcD_2)$}
        \label{fig: twin pag D_2}
    \end{subfigure}
    
    \caption{An example of the learning process of Algorithm~\ref{alg: I-MEC learning}.}
    \label{fig: ex of learning}
\end{figure}
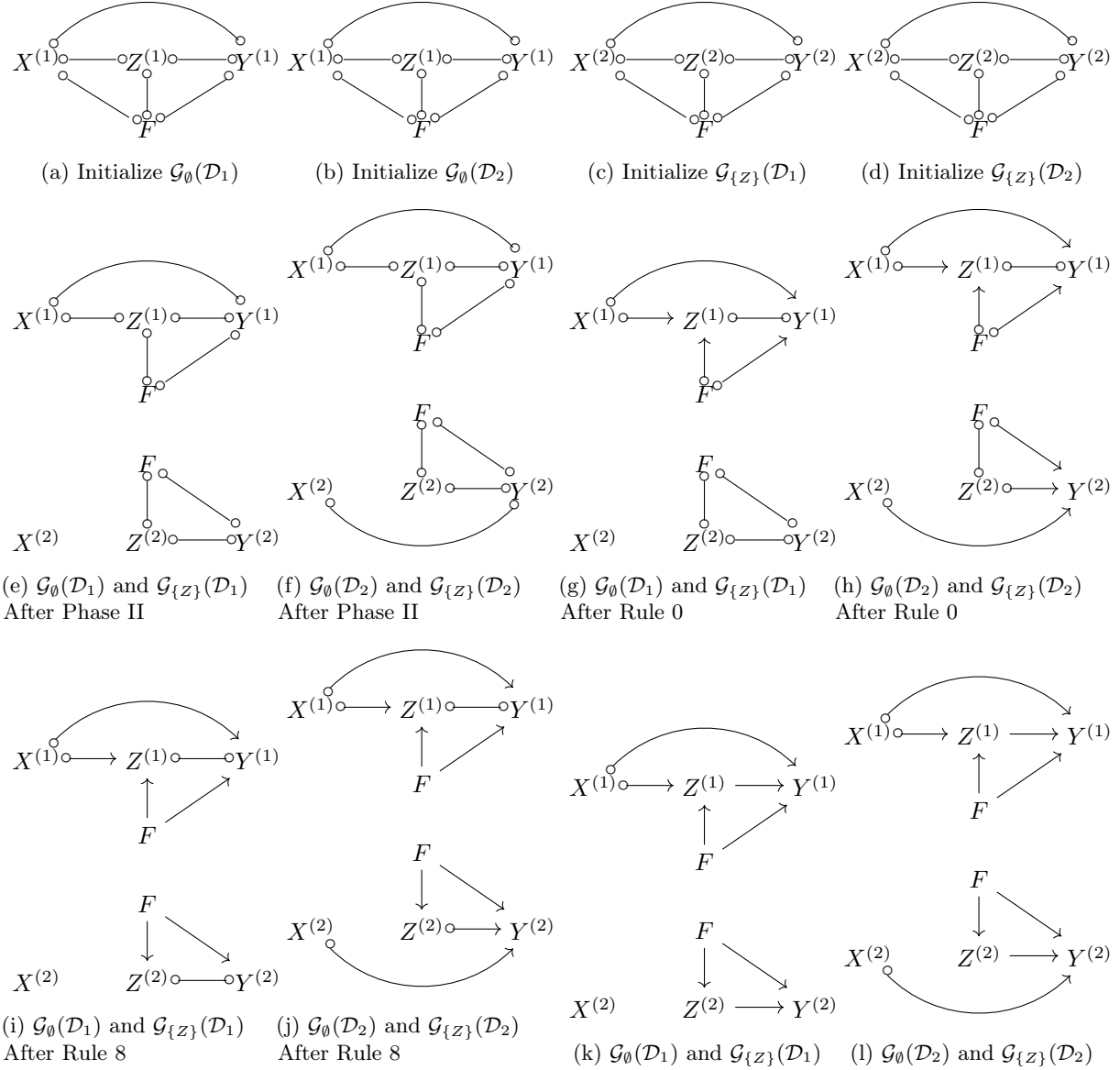

Here we show an example of the learning process of Algorithm~\ref{alg: I-MEC learning} in Figure~\ref{fig: ex of learning}. Consider the two causal graphs $\mcD_1, \mcD_2$ as shown in Figure~\ref{fig: D_1} and Figure~\ref{fig: D_2} respectively. Assume that we have access to $(P_{obs}, P_Z)$, i.e. $\mcI = \{\mI_1 = \emptyset, \mI_2 = \{Z\} \}$. In Phase I, we initialize the $\mcI$-augmented graphs under each interventional target for both graphs by constructing complete graphs with only circle edges within the observable nodes. After that, we create the $F$ nodes between each pair of intervention targets using Algorithm~\ref{alg: create F nodes} and connect each $F$ node to all observable nodes. The initialized $\mcI$-augmented graphs are shown in Figure~\ref{fig: D_1 obs pag}, Figure~\ref{fig: D_2 obs pag} Figure~\ref{fig: D_1 int Z}, and Figure~\ref{fig: D_2 int Z}. Here we omit the superscript for $F$ nodes since there are only two domains. Then in Phase II, we learn skeletons by using Algorithm~\ref{alg: find sep set} to check the invariance statements. Specifically, for each pair of $X, Y \in \mV$, if there exists some $\mW \subseteq \mV$ that separates $X, Y$ in any domain, we remove the circle edge between $X, Y$ accordingly. Similarly, if $F$ is separable from a vertex in $\mV$ given a condition set $\mW \subseteq \mV$, i.e. $P_\emptyset(y|\mw) = P_{\{ Z \}}(y|\mw)$, then we remove the circle edge between $F, Y$ in both domains. Thus we construct the skeletons of $\mcI$-augmented graphs for $\mcD_1$ and $\mcD_2$ as shown in Figure~\ref{fig: PAG D_1 phase II} and Figure~\ref{fig: PAG D_2 phase II} respectively. The upper graphs are for the observational domain while the lower graphs are for the domain of $P_Z$. Next, in Phase III, we apply the orientation rules. We start by finding all the unshielded colliders using Rule 0. In $\mcG_{\emptyset}(\mcD_1)$, we notice that conditioning on $Z^{(1)}$, $F$ is dependent on $X^{(1)}$. Similarly, conditioning on $Y^{(1)}$ , $F$ is dependent on $X^{(1)}$. We can then orient $X^{(1)} \cleftrightarrow Z^{(1)}$, $X^{(1)} \cleftrightarrow Y^{(1)}$, $F \cleftrightarrow Z^{(1)}$, and $F \cleftrightarrow Y^{(1)}$ as $X^{(1)} \crightarrow Y^{(1)}$, $X^{(1)} \crightarrow Z^{(1)}$, $F \crightarrow Z^{(1)}$, and $F \crightarrow Y^{(1)}$ respectively. The same structure appears in $\mcG_{\emptyset}(\mcD_2)$. However, in $\mcG_{\emptyset}(\mcD_2)$, we can identify the unshielded triplets $\langle X^{(2)}, Y^{(2)}, Z^{(2)} \rangle$ and $\langle X^{(2)}, Y^{(2)}, F \rangle$ which help us orient the colliders accordingly. The resulting graphs are plotted in Figure~\ref{fig: After Rule 8 D_1} and Figure~\ref{fig: After Rule 8 D_2}.
Using Rule 8, we orient all edges incident to $F$ out of $F$. The $\mcI$-augmented graphs after Rule 8 are presented in Figure~\ref{fig: After Rule 0' D_1} and Figure~\ref{fig: After Rule 0' D_2}.
Since only $Z$ is intervened but not $Y$, we can orient the edge $Z^{(2)}\rightarrow Y^{(2)}$ in both $\mcG_{\{ Z\}}(\mcD_1)$ and $\mcG_{\{ Z \}}(\mcD_2)$ using Rule 9. While Rule 10 helps us to orient $Z^{(1)}\crightarrow Y^{(1)}$, Rule 11 implies that we can orient $Z^{(1)}\rightarrow Y^{(1)}$ since $F$ points to $Y^{(1)}, Y^{(2)}$ in both $\mcG_{\emptyset}(\mcD_1)$ and $\mcG_{\emptyset}(\mcD_2)$ as $Y$ is not intervened. Thus, there has to be an inducing path from $F$ to $Y^{(1)}, Y^{(2)}$ that goes through $Z^{(1)}$ or $Z^{(2)}$ in both augmented pair graphs for $\mcD_1$ and $\mcD_2$. Notice that these edges cannot be oriented by the FCI rules.
At this stage, none of the rules apply anymore; therefore, $\mcG_{\emptyset}(\mcD_1)$, $\mcG_{\{Z\}}(\mcD_1)$ and $\mcG_{\emptyset}(\mcD_2)$, $\mcG_{ \{Z\} }(\mcD_2)$ are returned as the learned $\mcI$-augmented graphs of Algorithm~\ref{alg: I-MEC learning}. The final results are shown in Figure~\ref{fig:  twin pag D_1} and Figure~\ref{fig: twin pag D_2}. Notice that the learned $\mcI$-augmented graphs of the two causal graphs have different adjacencies and unshielded colliders. Therefore, they can be distinguished by the conditions listed in Theorem~\ref{thm: I-MEC}. 

\section{Comparison of Learning Objectives with Other Interventional Causal Discovery Algorithms}
\label{app: comparison with baselines}
Most existing interventional causal discovery algorithms aim to learn a single causal graph. However, the goal of our learning algorithm is to learn a tuple of $\mcI$-augmented graphs that entails (a superset of) $\mcI$-Markov equivalent ADMGs. None of the previous approaches outputs the same object. Concretely, $\mcI$-FCI~\citep{kocaoglu2019characterization}\footnote{The algorithm was proposed in~\citet{kocaoglu2019characterization} and later named as $\mcI$-FCI by~\citet{li2023causal}.} (known targets)/$\psi$-FCI~\citep{jaber2020causal} (unknown targets) assume soft interventions and cannot exploit the additional invariance constraints incurred by hard interventions. They use a single augmented MAG as the learning objective. JCI~\citep{mooij2020joint} claims it works for any interventions but it is overly simplistic and the learned PAG is less informative (in~\citet{jaber2020causal} App. D shows that $\psi$-FCI can learn more than JCI). GIES~\citep{hauser2012characterization} and IGSP~\citep{wang2017permutation} are score-based and aim at learning a single DAG under known targets. They neither guarantee inclusion-minimality nor provide equivalence-class certificates. Nevertheless, score-based methods usually impose parametric assumptions like linear function or additive Gaussian noise to the causal models. Such methods are not robust when the underlying mechanism deviates from those assumptions. Additionally, GIES and IGSP assume no latents and therefore their outputs are less informative. Because the outputs are of different types, a direct performance metric is unavailable to measure the performance of the methods. The previous sections have shown that empirically how much smaller the $\mcI$-MEC is vs. the soft $\mcI$-MEC produced by $\mcI$-FCI/$\psi$-FCI. Here we compare with GIES and IGSP with a simple example.

Consider the graph $[ X\rightarrow, Z \rightarrow Y, Z \leftarrow L \rightarrow Y ]$ where $L$ is a latent variable. We generate a Bayesian network with binary variables according to the graph. We take 100k samples from the observational distribution and 100k samples from $P_Z$. We choose 100k as it is more than enough to return correct CI tests. GIES returns $[Y \rightarrow Z \rightarrow X]$ which has the wrong causal order and does not belong to the same observational MEC. IGSP returns $[X \rightarrow Z \rightarrow Y, X \rightarrow Y]$ which is the observational MAG of the true graph, but it cannot tell anything to judge if there is a latent confounder between $Z, Y$. Under the same setting, when the simulated dataset is from a linear Gaussian structural causal model, both GIES and IGSP output $[X \rightarrow Z \rightarrow Y]$ which catches the right causal order but does not belong to the observational MEC. In both settings, our algorithm is able to learn the graphs illustrated in Figure~\ref{fig: twin pag D_1} which preserve information from both domains. We use the Python implementation of GIES by Olga Kolotuhina and Juan L. Gamella~\citep{gamella-gies}. The IGSP implementation is from the \texttt{causaldag} package~\citep{squires2018causaldag}.

\section{Further Discussion}
\subsection{Assumptions}
\subsubsection{Positivity}
In theory, Algorithm~\ref{alg: I-MEC learning} does not require a strict global positivity assumption. All results rely solely on h-faithfulness. “Strictly positive distribution” is added in some works on discovery as a convenience. This is because within the space of strictly positive parameterizations, the subset that violates faithfulness sits on algebraic varieties of lower dimension, hence has Lebesgue measure 0 (see ~\citet{robins2003uniform} and many follow-ups). In practice, with finite data, a zero or near-zero cell makes test statistics undefined or unstable, so implementations often impose minimum-count rules even when the theory would allow the zero.

The positivity assumption is required for identifiability (for example, \citet{kivva2022revisiting, kandasamy2019minimum}). This is because that the manipulation of numerical expression explicitly require the denominator to be strictly positive. By contrast, structural discovery uses only qualitative independence relations. As long as we can evaluate a test where, no global support assumption is required.

\subsubsection{Faithfulness}
There are weaker faithfulness assumptions. However, the soundness of Algorithm~\ref{alg: I-MEC learning} requires h-faithfulness. The $\mcI$-MEC we defined is based on the generalized do-calculus rules and so as the learning algorithm and hence h-faithfulness is minimal for the algorithm to be sound.

Like classic faithfulness, the set of parameterizations that violate h-faithfulness has Lebesgue measure zero for standard SEM families with continuous noise if we consider linear models; i.e., it holds generically (it also holds for many other models, see follow-ups of~\citet{robins2003uniform}). It is therefore no stronger in a measure–theoretic sense than ordinary faithfulness, but it is stronger logically: it rules out a few additional, finely-tuned parameter combinations that would create the extra cancellations specific to our Rules 1-4—those cancellations correspond exactly to our new orientation rules, so dropping h-faithfulness would break thesoundness. From the high level, imagine a causal graph under faithfulness assumption. It could be the case that for certain conditioning values $Z = z$, two variables $X$
and $Y$ that are d-connected are independent. While faithfulness covers the case that $X$  is dependent with $Y$
given $Z$ (when the dependency metric is averaged over all $z$), h-faithfulness is similar to saying $X, Y$ are dependent given every $Z = z$. This analogy is exact if the intervened node is a source node, but we believe it gives some insight on the leap from observational faithfulness to h-faithfulness.

\subsection{Algorithm Complexity Analysis}
Since our algorithm is also constraint‐based—like FCI and PC—the bulk of the computational cost comes from invariance tests, which can be seen as conditional independence tests involving the $F$ node. In the worst case, each pair of nodes may require an exponential number of tests to identify a separating set. However, in practice, many implementations (including FCI and PC) limit the size of the conditioning set (e.g., to a constant such as 3), which reduces the worst-case complexity to a polynomial bound of \(O(n^5)\) (see, e.g., \citet{spirtes2001causation}).

For the new orientation rules we introduce:
Rules 8, 9, and 11 operate in \(O(1)\) time since they only check the adjacencies of a fixed number of nodes.
Rules 10 and 11 have a complexity of \(O(n^2)\) as they potentially examine all pairs of nodes.
In comparison, the FCI orientation rules can take up to \(O(n^5)\) in the worst-case.

Thus, for each \(\mathcal{I}\)-augmented MAG, our learning algorithm has a similar worst-case complexity as the FCI algorithm, namely \(O(n^5)\). If there are \(k\) targets, the total complexity becomes \(O(kn^5)\).

\subsection{Comparison between Hard Interventions and Soft Interventions}
\begin{figure*}[ht]
    \centering
 \begin{subfigure}{0.32\columnwidth}
        \centering
        \begin{tikzpicture}
            \node (1) {{$X_1$}};
            \node (2) [right of = 1,xshift = 0.5 cm] {{$Y_1$}};
              \node (3) [right of = 2,xshift = 0.5 cm] { {$X_2$}};    
              \node (4) [right of = 3, xshift = 0.5 cm] {$Y_2$};

        \draw[<->] (1) -- (2) ;
        \draw[<->] (2) -- (3) ;
        \draw[<->] (3) -- (4) ;
        \end{tikzpicture}
        \caption{Example graph $\mcD$}
        \label{fig: ex D in app}
    \end{subfigure}%
\begin{subfigure}{0.32\columnwidth}
    \centering
    \begin{tikzpicture}
        \node (1) {{$X_1^{(1)}$}};
        \node (2) [right of = 1,xshift = 0.5 cm] {{$Y_1^{(1)}$}};
          \node (3) [right of = 2,xshift = 0.5 cm] { {$X_2^{(1)}$}};    
          \node (4) [right of = 3, xshift = 0.5 cm] {$Y_2^{(1)}$};
          \node (5) [above of = 2, xshift= 0.5 cm] {$F$};

        \draw[->] (5) -- (1) ;
        \draw[->] (5) -- (2) ;
        \draw[->] (5) -- (3) ;
        \draw[->] (5) -- (4) ;
    \end{tikzpicture}
    \caption{$\Aum_{\mI_1} (\mcD, \mcI)$}
    \label{fig: I aug mag D I1 app ex}
\end{subfigure}%
\begin{subfigure}{0.32\columnwidth}
    \centering
    \begin{tikzpicture}
        \node (1) {{$X_1^{(2)}$}};
        \node (2) [right of = 1,xshift = 0.5 cm] {{$Y_1^{(2)}$}};
          \node (3) [right of = 2,xshift = 0.5 cm] { {$X_2^{(2)}$}};    
          \node (4) [right of = 3, xshift = 0.5 cm] {$Y_2^{(2)}$};
          \node (5) [above of = 2, xshift= 0.5 cm] {$F$};

        \draw[->] (5) -- (1) ;
        \draw[->] (5) -- (2) ;
        \draw[->] (5) -- (3) ;
        \draw[->] (5) -- (4) ;
    \end{tikzpicture}
    \caption{$\Aum_{\mI_2} (\mcD, \mcI)$}
    \label{fig: I aug mag D I2 app ex}
\end{subfigure}%

    \begin{subfigure}{0.32\columnwidth}
    \centering
    \begin{tikzpicture}
        \node (1) {{$X_1$}};
        \node (2) [right of = 1,xshift = 0.5 cm] {{$Y_1$}};
        \node (3) [right of = 2,xshift = 0.5 cm] { {$X_2$}};    
        \node (4) [right of = 3, xshift = 0.5 cm] {$Y_2$};
        \node (5) [above of = 2, xshift= 0.5 cm] {$F$};

        \draw[<->] (1) -- (2) ;
        \draw[<->] (2) -- (3) ;
        \draw[<->] (3) -- (4) ;
        \draw[->] (5) -- (1) ;
        \draw[->] (5) -- (2) ;
        \draw[->] (5) -- (3) ;
        \draw[->] (5) -- (4) ;
    \end{tikzpicture}
    \caption{Augmented MAG of $\mcD$ under soft interventions}
    \label{fig: aug mag D under soft app}
\end{subfigure}%
 \begin{subfigure}{0.32\columnwidth}
        \centering
        \begin{tikzpicture}
            \node (1) {{$X_1$}};
            \node (2) [right of = 1,xshift = 0.5 cm] {{$Y_1$}};
              \node (3) [right of = 2,xshift = 0.5 cm] { {$X_2$}};    
              \node (4) [right of = 3, xshift = 0.5 cm] {$Y_2$};

        \draw[<->] (1) -- (2) ;
        \draw[<->] (2) -- (3) ;
        \draw[<->] (3) -- (4) ;
        \draw[<->, bend left = 45] (1) to (4) ;
        \end{tikzpicture}
        \caption{Example graph $\mcD_1$}
        \label{fig: ex D_1 in app soft better}
    \end{subfigure}%
     \begin{subfigure}{0.32\columnwidth}
        \centering
        \begin{tikzpicture}
            \node (1) {{$X_1$}};
            \node (2) [right of = 1,xshift = 0.5 cm] {{$Y_1$}};
              \node (3) [right of = 2,xshift = 0.5 cm] { {$X_2$}};    
              \node (4) [right of = 3, xshift = 0.5 cm] {$Y_2$};

        \draw[<->] (1) -- (2) ;
        \draw[<->] (3) -- (4) ;
        \draw[<->, bend left = 45] (1) to (4) ;
        \end{tikzpicture}
        \caption{Example graph $\mcD_2$}
        \label{fig: ex D_2 in app soft better}
    \end{subfigure}%
    
    \caption{An example that soft interventions lead to a smaller $\mcI$-Markov equivalence class than hard interventions. (a) is the ground truth causal graph $\mcD$ with intervention targets $\mcI = \{ \{X_1, X_2\}, \{Y_1, Y_2\} \}$; (b) and (c) are the two $\mcI$-augmented MAGs under hard interventions; (d) shows the augmented MAG under soft interventions; (e) and (f) are two examples graphs that are $\mcI$-Markov equivalent to $\mcD$ when $\mcI$ is hard but not $\mcI$-Markov equivalent to $\mcD$ when $\mcI$ is soft.}
    \label{fig: app example soft better}
\end{figure*}
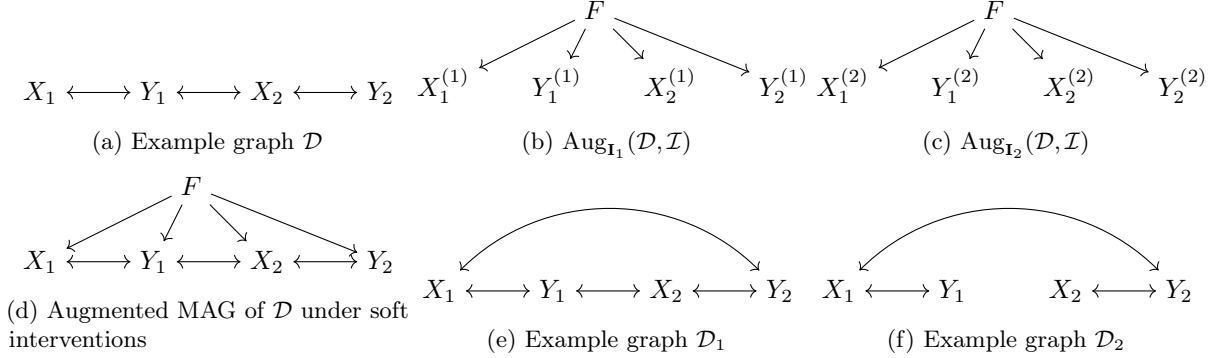

One may expect that hard interventions can always extract more information about the causal graph than soft interventions when there are latents. Here, we show an example in Figure~\ref{fig: app example soft better} where this is not true.  Consider the ground truth ADMG $\mcD$ in Figure~\ref{fig: ex D in app} with intervention targets $\mcI = \{ \mI_1 = \{X_1, X_2\}, \mI_2 = \{Y_1, Y_2\} \}$. The $\mcI$-augmented MAGs under hard interventions are shown in Figure~\ref{fig: I aug mag D I1 app ex} and Figure~\ref{fig: I aug mag D I2 app ex} respectively. The augmented MAG under soft interventions is shown in Figure~\ref{fig: aug mag D under soft app}. We suppress the superscript of $F$ nodes since they are the same. We notice that under hard interventions, the skeletons in both domains are empty graphs because the hard interventions remove all the bidirected edges in $\mcD$. However, under soft interventions, the skeleton of $\mcD$ is preserved since soft interventions do not modify the graphical structures. As a result, any graph that has only bidirected edges between $X_i$ and $Y_j, i, j \in \{1, 2\}$ is $\mcI$-Markov to $\mcD$ given $\mI_1, \mI_2$ as hard interventions while this is not true for soft interventions with the same targets. For example, consider the two graphs $\mcD_1, \mcD_2$ plotted in Figure~\ref{fig: ex D_1 in app soft better} and Figure~\ref{fig: ex D_2 in app soft better}. Given $\mcI$, their domain-specific skeletons will all be empty graphs as $\mcD$ when $\mcI$ is hard, thus they are $\mcI$-Markov equivalent to $\mcD$. However, with soft interventions, their skeletons can be preserved and thus not $\mcI$-Markov equivalent to $\mcD$ when $\mcI$ is soft. Therefore, for this kind of graphs and intervention targets, soft interventions may end up with a smaller $\mcI$-Markov equivalence class than hard interventions.

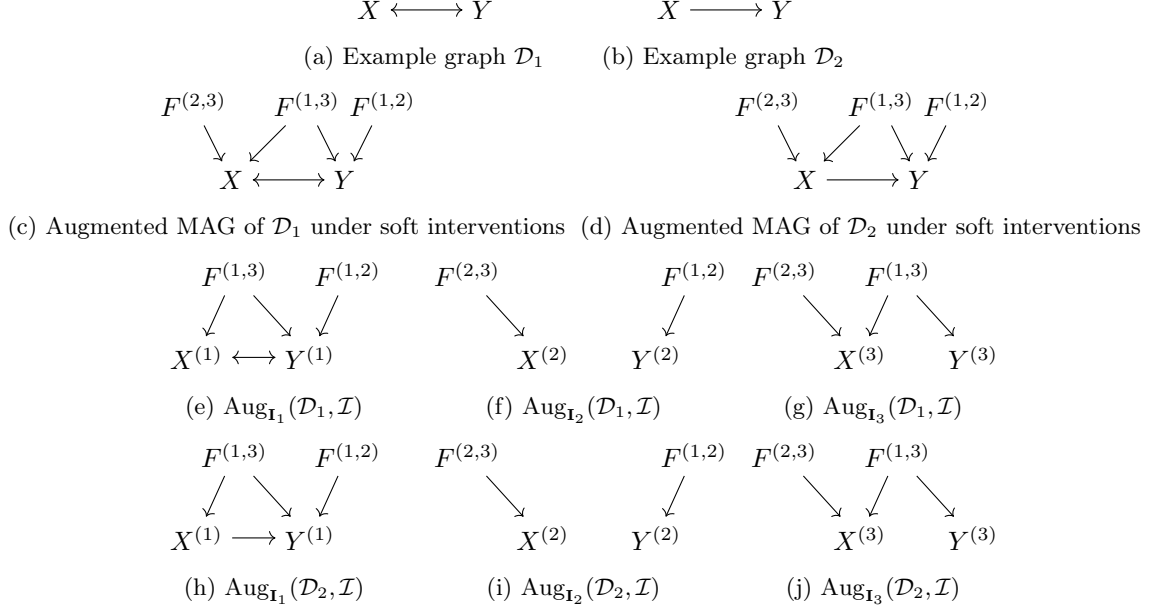
\begin{figure*}[ht]
    \centering
 \begin{subfigure}{0.24\columnwidth}
        \centering
        \begin{tikzpicture}
            \node (1) {{$X$}};
            \node (2) [right of = 1,xshift = 0.5 cm] {{$Y$}};

        \draw[<->] (1) -- (2) ;
        \end{tikzpicture}
        \caption{Example graph $\mcD_1$}
        \label{fig: ex D1 in app with obs}
    \end{subfigure}%
\begin{subfigure}{0.24\columnwidth}
    \centering
    \begin{tikzpicture}
        \node (1) {{$X$}};
        \node (2) [right of = 1,xshift = 0.5 cm] {{$Y$}};

        \draw[->] (1) -- (2) ;
    \end{tikzpicture}
    \caption{Example graph $\mcD_2$}
    \label{fig: ex D2 in app with obs}
\end{subfigure}%

    \begin{subfigure}{0.46\columnwidth}
    \centering
    \begin{tikzpicture}
        \node (1) {{$X$}};
        \node (2) [right of = 1,xshift = 0.5 cm] {{$Y$}};
        \node (3) [above of = 2, xshift = 0.5 cm] { {$F^{(1, 2)}$}};    
        \node (4) [above of = 2, xshift = -0.5 cm] {$F^{(1, 3)}$};
        \node (5) [above of = 1, xshift= -0.5 cm] {$F^{(2, 3)}$};

        \draw[<->] (1) -- (2) ;
        \draw[->] (3) -- (2) ;
        \draw[->] (4) -- (1) ;
        \draw[->] (4) -- (2) ;
        \draw[->] (5) -- (1) ;
    \end{tikzpicture}
    \caption{Augmented MAG of $\mcD_1$ under soft interventions}
    \label{fig: soft aug mag D1 with obs}
\end{subfigure}%
    \begin{subfigure}{0.46\columnwidth}
    \centering
    \begin{tikzpicture}
       \node (1) {{$X$}};
        \node (2) [right of = 1,xshift = 0.5 cm] {{$Y$}};
        \node (3) [above of = 2, xshift = 0.5 cm] { {$F^{(1, 2)}$}};    
        \node (4) [above of = 2, xshift = -0.5 cm] {$F^{(1, 3)}$};
        \node (5) [above of = 1, xshift= -0.5 cm] {$F^{(2, 3)}$};

        \draw[->] (1) -- (2) ;
        \draw[->] (3) -- (2) ;
        \draw[->] (4) -- (1) ;
        \draw[->] (4) -- (2) ;
        \draw[->] (5) -- (1) ;
    \end{tikzpicture}
    \caption{Augmented MAG of $\mcD_2$ under soft interventions}
    \label{fig: soft aug mag D2 with obs}
\end{subfigure}%

    \begin{subfigure}{0.24\columnwidth}
    \centering
    \begin{tikzpicture}
       \node (1) {{$X^{(1)}$}};
        \node (2) [right of = 1,xshift = 0.5 cm] {{$Y^{(1)}$}};
        \node (3) [above of = 2, yshift = 0.1 cm, xshift = 0.5 cm] { {$F^{(1, 2)}$}}; 
        \node (4) [above of = 1, yshift = 0.1 cm, xshift = 0.5 cm] {$F^{(1, 3)}$};

        \draw[<->] (1) -- (2) ;
        \draw[->] (3) -- (2) ;
        \draw[->] (4) -- (1) ;
        \draw[->] (4) -- (2) ;
    \end{tikzpicture}
    \caption{$\Aum_{\mI_1}(\mcD_1, \mcI)$}
    \label{fig: I1 aug mag D1 hard with obs}
\end{subfigure}%
    \begin{subfigure}{0.24\columnwidth}
    \centering
    \begin{tikzpicture}
       \node (1) {{$X^{(2)}$}};
        \node (2) [right of = 1,xshift = 0.5 cm] {{$Y^{(2)}$}};
        \node (3) [above of = 2, yshift = 0.1 cm, xshift = 0.5 cm] { {$F^{(1, 2)}$}}; 
        \node (4) [above of = 1, yshift = 0.1 cm, xshift = -1 cm] {$F^{(2, 3)}$};

        \draw[->] (3) -- (2) ;
        \draw[->] (4) -- (1) ;
    \end{tikzpicture}
    \caption{$\Aum_{\mI_2}(\mcD_1, \mcI)$}
    \label{fig: I2 aug mag D1 hard with obs}
\end{subfigure}%
    \begin{subfigure}{0.24\columnwidth}
    \centering
    \begin{tikzpicture}
       \node (1) {{$X^{(3)}$}};
        \node (2) [right of = 1,xshift = 0.5 cm] {{$Y^{(3)}$}};
        \node (3) [above of = 1, yshift = 0.1 cm, xshift = -1 cm] { {$F^{(2, 3)}$}}; 
        \node (4) [above of = 1, yshift = 0.1 cm, xshift = 0.5 cm] {$F^{(1, 3)}$};

        \draw[->] (3) -- (1) ;
        \draw[->] (4) -- (1) ;
        \draw[->] (4) -- (2) ;
    \end{tikzpicture}
    \caption{$\Aum_{\mI_3}(\mcD_1, \mcI)$}
    \label{fig: I3 aug mag D1 hard with obs}
\end{subfigure}%

        \begin{subfigure}{0.24\columnwidth}
    \centering
    \begin{tikzpicture}
       \node (1) {{$X^{(1)}$}};
        \node (2) [right of = 1,xshift = 0.5 cm] {{$Y^{(1)}$}};
        \node (3) [above of = 2, yshift = 0.1 cm, xshift = 0.5 cm] { {$F^{(1, 2)}$}}; 
        \node (4) [above of = 1, yshift = 0.1 cm, xshift = 0.5 cm] {$F^{(1, 3)}$};

        \draw[->] (1) -- (2) ;
        \draw[->] (3) -- (2) ;
        \draw[->] (4) -- (1) ;
        \draw[->] (4) -- (2) ;
    \end{tikzpicture}
    \caption{$\Aum_{\mI_1}(\mcD_2, \mcI)$}
    \label{fig: I1 aug mag D2 hard with obs}
\end{subfigure}%
    \begin{subfigure}{0.24\columnwidth}
    \centering
    \begin{tikzpicture}
       \node (1) {{$X^{(2)}$}};
        \node (2) [right of = 1,xshift = 0.5 cm] {{$Y^{(2)}$}};
        \node (3) [above of = 2, yshift = 0.1 cm, xshift = 0.5 cm] { {$F^{(1, 2)}$}}; 
        \node (4) [above of = 1, yshift = 0.1 cm, xshift = -1 cm] {$F^{(2, 3)}$};

        \draw[->] (3) -- (2) ;
        \draw[->] (4) -- (1) ;
    \end{tikzpicture}
    \caption{$\Aum_{\mI_2}(\mcD_2, \mcI)$}
    \label{fig: I2 aug mag D2 hard with obs}
\end{subfigure}%
    \begin{subfigure}{0.24\columnwidth}
    \centering
    \begin{tikzpicture}
       \node (1) {{$X^{(3)}$}};
        \node (2) [right of = 1,xshift = 0.5 cm] {{$Y^{(3)}$}};
        \node (3) [above of = 1, yshift = 0.1 cm, xshift = -1 cm] { {$F^{(2, 3)}$}}; 
        \node (4) [above of = 1, yshift = 0.1 cm, xshift = 0.5 cm] {$F^{(1, 3)}$};

        \draw[->] (3) -- (1) ;
        \draw[->] (4) -- (1) ;
        \draw[->] (4) -- (2) ;
    \end{tikzpicture}
    \caption{$\Aum_{\mI_3}(\mcD_2, \mcI)$}
    \label{fig: I3 aug mag D2 hard with obs}
\end{subfigure}%
    \caption{An example that given the observational distribution, soft interventions can distinguish the given two causal graphs while hard interventions cannot. (a), (b) are the ground truth causal graph $\mcD_1, \mcD_2$ respectively with intervention targets $\mcI = \{ \emptyset, \{Y\}, \{X, Y\} \}$; (c), (d) are the augmented MAGs under soft interventions for $\mcD_1$ and $\mcD_2$ respectively; (e), (f), (g) are the $\mcI$-augmented MAGs under hard interventions for $\mcD_1$; (h), (i), (j) are the $\mcI$-augmented MAGs under hard interventions for $\mcD_2$. Notice that $\mcD_1$ and $\mcD_2$ are not $\mcI$-Markov equivalent when $\mcI$ is soft because the triple $\langle F, X, Y\rangle$ has different unshielded collider status in the corresponding augmented MAGs. However, they are $\mcI$-Markov equivalent when $\mcI$ is hard because their $\mcI$-augmented MAGs corresponding to the same domains all satisfy the 3 conditions.}
    \label{fig: app example soft better with obs}
\end{figure*}

If the observational distribution is provided, the skeleton is preserved for both hard and soft interventions. However, the following example in Figure~\ref{fig: app example soft better with obs} shows that even given the observational distribution, there still exist causal graphs which can be distinguished by soft interventions but not so with hard interventions. Consider the causal graphs $\mcD_1$ and $\mcD_2$ as shown in Figure~\ref{fig: ex D1 in app with obs} and Figure~\ref{fig: ex D2 in app with obs} with intervention targets $\mI_1 = \emptyset, \mI_2 = \{Y\}, \text{and}\; \mI_3 = \{X, Y\}$. Here we use the index of the targets for superscripts and subscripts for simplicity. The augmented MAGs of $\mcD_1$ and $\mcD_2$ under soft interventions are shown in Figure~\ref{fig: soft aug mag D1 with obs} and Figure~\ref{fig: soft aug mag D2 with obs} respectively. We can see that the triple $\langle F^{(2, 3)}, X, Y\rangle$ forms an unshielded collider in the augmented MAG of $\mcD_1$ but it is a non-collider in the augmented MAG of $\mcD_2$. Therefore, they do not satisfy the 3 conditions,  meaning that they are not $\mcI$-Markov equivalent when $\mcI$ is soft. The $\mcI$-augmented MAGs for $\mcD_1$ and $\mcD_2$ are shown in Figure~\ref{fig: I1 aug mag D1 hard with obs}, ~\ref{fig: I2 aug mag D1 hard with obs}, ~\ref{fig: I3 aug mag D1 hard with obs} and Figure~\ref{fig: I1 aug mag D2 hard with obs}, ~\ref{fig: I2 aug mag D2 hard with obs}, ~\ref{fig: I3 aug mag D2 hard with obs} respectively. We can see that all the 3 pairs of corresponding $\mcI$-augmented MAGs satisfy the 3 conditions. Therefore, $\mcD_1$ and $\mcD_2$ are $\mcI$-Markov equivalent under hard interventions although the observational distribution is given. If an additional intervention target $\mI_4 = \{ X\}$ is given, $\mcD_1, \mcD_2$ will be non-$\mcI$-Markov equivalent with a different collider status of the triple $\langle F^{(1, 4)}, X^{(4)}, Y^{(4)}\rangle$ in $\Aum_{\mI_4}(\mcD_1, \mcI)$ and $\Aum_{\mI_4}(\mcD_2, \mcI)$. We conclude that although hard interventions are stronger than soft interventions, they do not always lead to a smaller $\mcI$-Markov equivalence class than soft interventions. It may also imply that using a mixture of hard and soft interventions may be more efficient in causal discovery than using either of them alone.

\subsection{Incompleteness of the Learning Algorithm}
\label{app: imcompleteness}
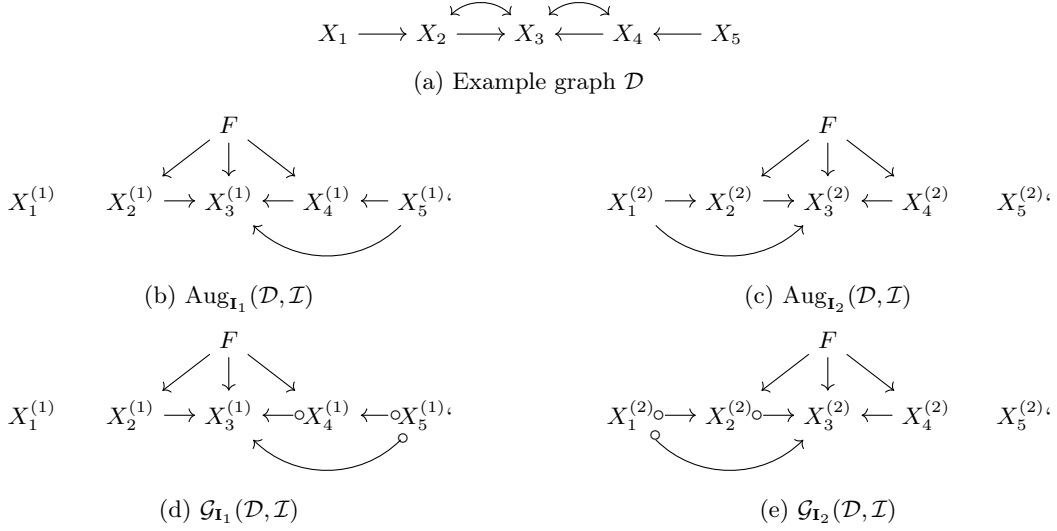
\begin{figure*}[ht]
    \centering
 \begin{subfigure}{0.48\columnwidth}
        \centering
        \begin{tikzpicture}[font=\sffamily\small]
            \node (1) {{$X_1$}};
            \node (2) [right of = 1,xshift = 0.3 cm] {{$X_2$}};
              \node (3) [right of = 2,xshift = 0.3 cm] { {$X_3$}};    
              \node (4) [right of = 3, xshift = 0.3 cm] {$X_4$};
              \node (5) [right of = 4, xshift = 0.3 cm] {$X_5$};

        \draw[->] (1) -- (2) ;
        \draw[->] (2) -- (3) ;
        \draw[->] (4) -- (3) ;
        \draw[->] (5) -- (4) ;
        \draw[<->, bend left=45] (2) to (3) ;
        \draw[<->, bend right=45] (4) to (3) ;
        
        \end{tikzpicture}
        \caption{Example graph $\mcD$}
        \label{fig: ex D in app learning incomplete}
    \end{subfigure}%
    
\begin{subfigure}{0.48\columnwidth}
    \centering
    \begin{tikzpicture}[font=\sffamily\small]
        \node (1) {{$X_1^{(1)}$}};
            \node (2) [right of = 1,xshift = 0.3 cm] {{$X_2^{(1)}$}};
              \node (3) [right of = 2,xshift = 0.3 cm] { {$X_3^{(1)}$}};    
              \node (4) [right of = 3, xshift = 0.3 cm] {$X_4^{(1)}$};
              \node (5) [right of = 4, xshift = 0.3 cm] {$X_5^{(1)}`$};
              \node (6) [above of = 3] {$F$};

        \draw[->] (2) -- (3) ;
        \draw[->] (4) -- (3) ;
        \draw[->] (5) -- (4) ;
        \draw[->, bend left=45] (5) to (3) ;
        \draw[->] (6) -- (2) ;
        \draw[->] (6) -- (3) ;
        \draw[->] (6) -- (4) ;
    \end{tikzpicture}
    \caption{$\Aum_{\mI_1}(\mcD, \mcI)$}
    \label{fig: I aug mag1 app learning imcomplete}
\end{subfigure}%
\begin{subfigure}{0.48\columnwidth}
    \centering
    \begin{tikzpicture}[font=\sffamily\small]
        \node (1) {{$X_1^{(2)}$}};
            \node (2) [right of = 1,xshift = 0.3 cm] {{$X_2^{(2)}$}};
              \node (3) [right of = 2,xshift = 0.3 cm] { {$X_3^{(2)}$}};    
              \node (4) [right of = 3, xshift = 0.3 cm] {$X_4^{(2)}$};
              \node (5) [right of = 4, xshift = 0.3 cm] {$X_5^{(2)}`$};
              \node (6) [above of = 3] {$F$};

        \draw[->] (2) -- (3) ;
        \draw[->] (4) -- (3) ;
        \draw[->] (1) -- (2) ;
        \draw[->, bend right=45] (1) to (3) ;
        \draw[->] (6) -- (2) ;
        \draw[->] (6) -- (3) ;
        \draw[->] (6) -- (4) ;
    \end{tikzpicture}
    \caption{$\Aum_{\mI_2}(\mcD, \mcI)$}
    \label{fig: I aug mag2 app learning imcomplete}
\end{subfigure}%
    
\begin{subfigure}{0.48\columnwidth}
    \centering
    \begin{tikzpicture}[font=\sffamily\small]
        \node (1) {{$X_1^{(1)}$}};
            \node (2) [right of = 1,xshift = 0.3 cm] {{$X_2^{(1)}$}};
              \node (3) [right of = 2,xshift = 0.3 cm] { {$X_3^{(1)}$}};    
              \node (4) [right of = 3, xshift = 0.3 cm] {$X_4^{(1)}$};
              \node (5) [right of = 4, xshift = 0.3 cm] {$X_5^{(1)}`$};
              \node (6) [above of = 3] {$F$};

        \draw[->] (2) -- (3) ;
        \draw[->] (4) -- (3) node[pos=0.3, right] {\(\circ\)} ;
        \draw[->] (5) -- (4) node[pos=0.3, right]{\(\circ\)};
        \draw[->, bend left=45] (5) to node[pos=0, xshift = -5, yshift = 6]{\(\circ\)} (3) ;
        \draw[->] (6) -- (2) ;
        \draw[->] (6) -- (3) ;
        \draw[->] (6) -- (4) ;
    \end{tikzpicture}
    \caption{$\mcG_{\mI_1}(\mcD, \mcI)$}
    \label{fig: I aug graph D_1 app learning imcomplete}
\end{subfigure}%
    \begin{subfigure}{0.48\columnwidth}
    \centering
    \begin{tikzpicture}[font=\sffamily\small]
        \node (1) {{$X_1^{(2)}$}};
            \node (2) [right of = 1,xshift = 0.3 cm] {{$X_2^{(2)}$}};
              \node (3) [right of = 2,xshift = 0.3 cm] { {$X_3^{(2)}$}};    
              \node (4) [right of = 3, xshift = 0.3 cm] {$X_4^{(2)}$};
              \node (5) [right of = 4, xshift = 0.3 cm] {$X_5^{(2)}`$};
              \node (6) [above of = 3] {$F$};

        \draw[->] (2) -- (3) node[pos=0.3, left]{\(\circ\)};
        \draw[->] (4) -- (3) ;
        \draw[->] (1) -- (2) node[pos=0.3, left]{\(\circ\)};
        \draw[->, bend right=45] (1) to node[pos=0, xshift = -6, yshift = -4]{\(\circ\)} (3) ;
        \draw[->] (6) -- (2) ;
        \draw[->] (6) -- (3) ;
        \draw[->] (6) -- (4) ;
    \end{tikzpicture}
    \caption{$\mcG_{\mI_2}(\mcD, \mcI)$}
    \label{fig: I aug graph 2 app learning incomplete}
\end{subfigure}%
    
    \caption{An example to show that Algorithm~\ref{alg: I-MEC learning} is not complete. (a) is the ground truth causal graph $\mcD$ with intervention targets $\mcI = \{ \{X_2\}, \{X_4\} \}$; (b) and (c) are the $\mcI$-augmented MAG under $\mI_1$ and $\mI_2$ respectively; (d) and (e) are the domain-specific $\mcI$-augmented graphs learned by Algorithm~\ref{alg: I-MEC learning} under $\mI_1$ and $\mI_2$ respectively. The proposed Algorithm~\ref{alg: I-MEC learning} cannot recover the circle marks at $X_2$ and $X_4$, but there does not exist a causal graph that has an arrowhead at the same places in their $\mcI$-augmented MAGs.}
    \label{fig: app example incomplete}
\end{figure*}

In this section, we will present an example in Figure~\ref{fig: app example incomplete} to show that Algorithm~\ref{alg: I-MEC learning} is not complete, i.e., for some causal graphs and intervention targets $\mcI$, there exist circle marks in the $\mcI$-augmented graph returned by Algorithm~\ref{alg: I-MEC learning} that can be an arrowhead in the $\mcI$-augmented MAG of a causal graph that is $\mcI$-Markov equivalent, and an arrowtail in another $\mcI$-augmented MAG. Let us consider the causal graph $\mcD$ as shown in Figure~\ref{fig: ex D in app learning incomplete} with intervention targets $\mcI = \{ \mI_1 = \{ X_2 \}, \mI_2 = \{ X_4 \} \}$. Here we use the domain index as the superscripts for simplicity. The $\mcI$-augmented MAGs $\Aum_{\mI_1}(\mcD, \mcI)$ and $\Aum_{\mI_2}(\mcD, \mcI)$ are shown in Figure~\ref{fig: I aug mag1 app learning imcomplete} and Figure~\ref{fig: I aug mag2 app learning imcomplete} respectively. The $\mcI$-augmented graphs $\mcG_{\mI_1}(\mcD, \mcI)$ and $\mcG_{\mI_2}(\mcD, \mcI)$ learned by Algorithm~\ref{alg: I-MEC learning} are shown in Figure~\ref{fig: I aug graph D_1 app learning imcomplete} and Figure~\ref{fig: I aug graph 2 app learning incomplete} respectively. $X_2^{(1)} \rightarrow X_3^{(1)}$ and $X_4^{(2)} \rightarrow X_3^{(2)}$ are oriented by the inducing path rule. However, our algorithm cannot orient $X_4^{(1)} \rightarrow X_3^{(1)}$ and $X_2^{(2)} \rightarrow X_3^{(2)}$. Without loss of generality, let's consider $X_4^{(1)} \crightarrow X_3^{(1)}$ in $\mcG_{\mI_1}(\mcD)$. Suppose that there exists a graph $\mcD^\prime$ such that there is $X_4^{(1)} \leftrightarrow X_3^{(1)}$ in $\mcG_{\mI_1}(\mcD^\prime, \mcI)$. Since we have $X_4^{(2)} \rightarrow X_3^{(2)}$ in $\mcG_{\mI_2}(\mcD^\prime, \mcI)$, this indicates that there is an inducing path and a directed path from $X_4$ to $X_3$ in $\mcD_{\overline{X_4}}^\prime$ and thus in $\mcD^\prime$ as $\mcD_{\overline{X_4}}^\prime$ is  a subgraph of $\mcD^\prime$. The bidirected edge between $X_4^{(1)}$ and $X_3^{(1)}$ in $\mcG_{\mI_1}(\mcD^\prime, \mcI)$ indicates that the directed path from $X_4$ to $X_3$ is broken in $\mcD_{\overline{X_2}}^\prime$. This can only happen when $X_2$ is in the directed path between $X_4$ and $X_3$ in $\mcD^\prime$ and thus $\mcD_{\overline{X_4}}^\prime$ as removing the edges into $X_4$ does not affect its descendants. Therefore, $X_4$ has a child other than $X_3$ and is an ancestor of $X_2$. Nevertheless, $X_4^{(2)}$ has only one possible child $X_3^{(2)}$ in $\mcG_{\mI_2}(\mcD, \mcI)$. Thus, the supposition does not hold. For the same reason, we can orient the circle mark in $X_2^{(2)} \crightarrow X_3^{(2)}$ in $\mcG_{\mI_2}(\mcD, \mcI)$. These circle marks can only be oriented by comparing the graphical structures across domains. We need extra orientation rules to catch them.

\section{Limitations, Future Work, and Broader Impact}
\label{app: broader impact}
\paragraph{Limitations:} While our work provides a general characterization of 
$\mcI$-Markov equivalence classes, the resulting representation is a tuple of augmented graphs, rather than a single unified graph. This makes the characterization less straightforward to interpret. In addition, although our proposed algorithm is sound, it is not complete. We just illustrate this limitation with a concrete example provided in Appendix~\ref{app: imcompleteness}.
\paragraph{Future Work:} This work opens several promising directions for future research. First, our empirical results suggest that hard interventions tend to be more informative than soft interventions. It would be valuable to formally analyze this observation and establish theoretical conditions under which this holds. Additionally, while we have proven the soundness of our learning algorithm, extending it with additional orientation rules to achieve completeness remains an open challenge. Finally, our current framework assumes full access to interventional distributions. Developing methods that can learn causal graphs from limited interventional samples is an important direction for making these approaches more practical in real-world settings.

\paragraph{Broader Impact:} This work contributes to advancing the theoretical and algorithmic foundations of causal discovery, with the potential to improve decision-making/causal inference in critical domains such as healthcare, economics, genomics, and social sciences~\citep{sharma2022lerna, zhou2022kratos, hernan2016using, sachs2005causal, taubman2014medicaid, lee2008randomized, belyaeva2021dci}. By combining observational and interventional data, the proposed methods can reduce reliance on large-scale experimentation, making causal analysis more accessible in data-scarce or resource-limited settings. This could empower practitioners to build more reliable models for understanding complex systems, ultimately benefiting scientific discovery and evidence-based policy-making.

However, the use of causal discovery methods also carries risks. Incorrect or biased causal inferences—whether due to data limitations, modeling assumptions, or algorithmic shortcomings—could lead to misguided conclusions, especially in sensitive applications like healthcare or policy-making. Furthermore, there is a risk of misuse if these methods are deployed without adequate validation or oversight, potentially reinforcing harmful biases or supporting flawed decision-making processes.

To mitigate these risks, it is crucial to promote transparency, reproducibility, and the inclusion of domain expertise when applying these methods. Future work should also explore techniques to quantify uncertainty in causal conclusions, improve model interpretability, and develop guidelines or safeguards for responsible deployment.


\end{document}